\documentclass[a4paper,fleqn]{cas-sc}

\usepackage{tcolorbox}
\tcbuselibrary{listingsutf8}
\newtcolorbox[auto counter]{promptbox}[2][]{%
  colback=gray!5!white, 
  colframe=blue!60!teal, 
  fonttitle=\bfseries,
  fontupper=\footnotesize,
  title=Prompt~\thetcbcounter: #2, 
  label={#1}
}
\definecolor{RowGray}{gray}{0.95}
\usepackage{subcaption}
\usepackage{longtable}
\usepackage[table]{xcolor}
\usepackage{array}
\usepackage{tabularx}
\usepackage{xcolor,colortbl}
\usepackage{float}
\usepackage{array}
\usepackage{booktabs}
\usepackage[authoryear]{natbib}
\setcitestyle{authoryear, maxcitenames=2, maxbibnames=2}

\usepackage{revision_common}

\def\tsc#1{\csdef{#1}{\textsc{\lowercase{#1}}\xspace}}
\tsc{WGM}
\tsc{QE}
\tsc{EP}
\tsc{PMS}
\tsc{BEC}
\tsc{DE}

\begin{document}
\let\WriteBookmarks\relax
\renewcommand{\topfraction}{0.9}
\renewcommand{\bottomfraction}{0.5}
\renewcommand{\textfraction}{0.07}
\renewcommand{\floatpagefraction}{0.75}
\setcounter{topnumber}{3}
\setcounter{bottomnumber}{2}
\setcounter{totalnumber}{5}

\shorttitle{From S4 to Mamba: A Survey}

\shortauthors{Somvanshi et~al.}

\title [mode = title]{Advancing Intelligent Sequence Modeling: Evolution, Trade-offs, and Applications of State-Space Architectures from S4 to Mamba}

%
\author[1]{Shriyank Somvanshi}[orcid=0009-0008-3723-0607]
\ead{shriyank@txstate.edu}
\credit{Draft manuscript preparation}

\author[1]{Md Monzurul Islam}[orcid=0009-0007-3670-6100]
\ead{monzurul@txstate.edu}
\credit{Draft manuscript preparation}

\author[1]{Mahmuda Sultana Mimi}[orcid=0009-0007-8534-3633]
\ead{qnb9@txstate.edu}
\credit{Draft manuscript preparation}

\author[1]{Sazzad Bin Bashar Polock}[orcid=0009-0005-5505-4174]
\ead{pay28@txstate.edu}
\credit{Draft manuscript preparation}

\author[2]{Gaurab Chhetri}[orcid=0009-0000-0124-4814]
\cormark[1]
\ead{gaurab@txstate.edu}
\credit{Draft manuscript preparation}

\author[1]{Anandi Dutta, Ph.D.}[orcid=0009-0002-7279-7752]
\ead{anandi.dutta@txstate.edu}
\credit{Draft manuscript preparation}

\author[1]{Amir Rafe, Ph.D.}[orcid=0000-0002-4089-2088]
\ead{amir.rafe@txstate.edu}
\credit{Draft manuscript preparation}

\author[1]{Subasish Das, Ph.D.}[orcid=0000-0002-1671-2753]
\ead{subasish@txstate.edu}
\credit{Draft manuscript preparation}

\affiliation[1]{organization={Civil Engineering, Texas State University},
    addressline={601 University Drive}, 
    city={San Marcos},
    postcode={78666 TX}, 
    country={USA}}

\affiliation[2]{organization={Computer Science, Texas State University},
    addressline={601 University Drive}, 
    city={San Marcos},
    postcode={78666 TX}, 
    country={USA}}
    
\cortext[cor1]{Corresponding author}

\begin{abstract}
Structured State Space Models (SSMs) have become a prominent class of sequence models, developed against two long-standing difficulties: the sequential computation and gradient propagation limits of Recurrent Neural Networks (RNNs), and the quadratic time and memory cost of self-attention in Transformers. By combining structured recurrence with state-space representations, SSMs attain linear or near-linear scaling in sequence length and hold a constant-size recurrent state during autoregressive decoding, so that no key-value cache grows with context. This paper is a structured review of the lineage running from the Structured State Space Sequence model (S4), through its diagonal and simplified successors S4D, Diagonal State Spaces (DSS) and S5, to the selective models Mamba and Mamba-2, and on to SSM-attention hybrids. Rather than describing models one at a time, the analysis is organized around cross-cutting design dimensions: input-dependent selectivity, the recurrent and convolutional views and when each is preferable, diagonalization, cache behavior at inference, hardware-aware kernels, and hybridization. Structured state-space duality is treated as a central thread, because it accounts for the equivalence between the recurrent and the attention-like form and for why hybrid designs work. Reported efficiency and accuracy results are presented with their measurement configuration and evidence level, because observed speedups reflect the combined effects of the algorithm, implementation kernel and hardware rather than an intrinsic property of the model alone. SSMs are competitive with Transformers and more efficient in specific long-context regimes, whereas attention retains an advantage on tasks dominated by exact retrieval and associative recall. The review closes with limitations, failure modes, deployment considerations and open problems.
\end{abstract}


\begin{highlights}
  \item Structured review of the S4 to Mamba lineage with a stated selection protocol.
  \item Cross-cutting analysis of selectivity, scan versus convolution and cache behavior.
  \item State-space duality developed as the thread linking Mamba-2 to attention and hybrids.
  \item Reported results are tabulated with its configuration and an evidence level.
  \item Dedicated treatment of failure modes, deployment, quantization and edge memory.
\end{highlights}


\maketitle


\section{Introduction}
Traditional sequence  modeling architectures, such as Recurrent Neural Networks (RNNs) and Transformers, have demonstrated significant limitations in handling long-range dependencies, particularly in domains such as natural language processing (NLP), speech processing, vision, and time-series forecasting. RNNs suffer from the vanishing and exploding gradient problem, limiting their ability to retain information over extended sequences, and their inherently sequential nature restricts parallelization, making them inefficient for long-range tasks \citep{patro2024mamba}. While Long Short-Term Memory (LSTM) networks alleviate some of these issues through gating mechanisms, they still introduce additional complexity and computational overhead. Transformers, on the other hand, rely on self-attention mechanisms that capture global dependencies but scale poorly due to their quadratic time and space complexity (O(N²)  \citep{patro2024mamba}. Moreover, Transformers face challenges with fixed-length positional embeddings and excessive memory consumption, making them inefficient for processing extremely long sequences.  

Beyond these challenges, RNNs struggle with slow training times due to their sequential processing, while Transformers become impractical for extremely long inputs, such as document-level NLP tasks or long-duration speech modeling, due to their memory-intensive pairwise attention computations \citep{gu2021efficiently}. These issues have motivated the development of Structured State Space Models (SSMs), which offer a fundamentally different approach to sequence modeling. Relative to RNNs, SSMs mitigate rather than eliminate the gradient decay problem: a structured, and in the diagonal case explicitly parameterized, state transition keeps the spectrum of the recurrence under control, so that gradient signal decays far more slowly over long horizons instead of not decaying at all \citep{gu2021efficiently, gu2022parameterization}. Relative to Transformers, SSMs replace pairwise attention with a state-space recurrence whose cost grows linearly in the sequence length $T$ for a fixed state size $N$, rather than quadratically in $T$.

Structured recurrence and continuous-time parameterization are what make this practical. Stable reparameterizations of the state transition are what allow long-term memory to be retained without the exponential decay characteristic of unconstrained recurrent models \citep{wang2023stablessm}. Two distinct computational routes exploit that structure. When the recurrence is linear and time-invariant, the whole sequence can be produced as a single long convolution and evaluated in $O(T \log T)$ time with the fast Fourier transform, which is the route S4 takes at training time \citep{gu2021efficiently}. When the recurrence must be evaluated step by step, an associative parallel scan recovers parallelism across the sequence, which is the route S5 and the selective models take \citep{smith2022simplified}. The Structured State Space Sequence model (S4) introduced the parameterization that made the first route tractable, reducing the memory cost of the state transition while preserving the state-space formulation \citep{gu2021efficiently}.Additionally, hybrid architectures such as State sPace AugmenteD TransformEr (SPADE) incorporate SSMs into Transformer layers to enhance global information capture while maintaining local contextual refinement through efficient attention mechanisms \citep{zuo2022efficient}. 

On the Long Range Arena (LRA) benchmark, S4 was the first model to exceed chance on the Path-X task at sequence length 16{,}384, and it reported roughly 60 times the autoregressive generation throughput of a vanilla Transformer baseline, 48K against 0.8K tokens per second on WikiText-103 and 20.84 against 0.32 images per second on CIFAR-10, measured on a single A100 with the batch size raised to fill memory \citep{gu2021efficiently}. These results represent large-batch generation throughput rather than per-request latency and are specific to the evaluated model, baseline, tasks and hardware configuration. Accordingly, Section 9 reports efficiency results together with the conditions under which they were measured.

Selective SSMs make the state update depend on the input, so that the model can decide what to write into and what to discard from a fixed-size state \citep{gu2023mamba}. This is the change that closed much of the content-based reasoning gap between linear SSMs and attention, and it is examined as a design dimension in Section~\ref{sec:crosscut}. Models built this way have been applied across language, speech, vision, time-series, graph and multimodal settings; Section~\ref{sec:applications} reviews that evidence domain by domain, and is explicit about where the comparison against a strong Transformer baseline has actually been run and where it has not.

\subsection{Objective of the Survey}
SSMs have emerged as a promising alternative to Transformer-based architectures, offering efficient sequence modeling with linear computational complexity. The evolution of SSMs, beginning with the foundational S4 model, has led to significant advancements, including Mamba, S5, and Jamba, which incorporate selective mechanisms to improve performance across various domains. Early SSMs, such as S4, were primarily designed for long-range sequence modeling but had limitations in expressive power, particularly in content-based reasoning tasks \citep{muca2024theoretical}. The introduction of Mamba refined these models by integrating a selective state-space mechanism that allows input-dependent processing, significantly enhancing expressivity and accuracy \citep{muca2024theoretical}. Mamba-2 then reformulated the selective recurrence so that its core computation is dominated by matrix multiplications, which run on the matrix units of modern accelerators. Dao and Gu report that a dedicated kernel for this formulation is 2 to 8 times faster than Mamba's fused selective scan in a core-layer microbenchmark on a single A100 80GB PCIe device at state size 64, swept over sequence lengths from 512 to 512K \citep{dao2024transformers}. This is a kernel-level measurement of one layer rather than an end-to-end model speedup, and Section~\ref{sec:ssd} develops the duality result that makes it possible.

SSMs offer a computational advantage over Transformers, which scale quadratically in sequence length, making them expensive for long-form processing. Instead, models like Mamba and S5 achieve linear scaling, improving inference speed and reducing memory requirements while maintaining comparable accuracy \citep{muca2024theoretical, smith2022simplified}. S5 introduces a multi-input, multi-output (MIMO) SSM, enhancing parallelism and efficiency over S4, allowing it to match Transformer performance while using fewer computational resources \citep{smith2022simplified}. This advantage has led to SSMs being increasingly adopted in various real-world applications, such as speech processing \citep{miyazaki2024exploring, smith2022simplified}, medical imaging \citep{bansal2024comprehensive}, and NLP \citep{dao2024transformers}. While Transformers still dominate NLP tasks, hybrid models like Jamba, which combine Transformer layers with Mamba components, report throughput and memory advantages in long-context settings; the specific results and the conditions under which they were measured are given in the paragraph below and in Table~\ref{tab:empirical} \citep{team2024jamba}.

Despite these advancements, challenges remain. Training SSMs efficiently at scale is still an open problem compared to Transformer-based architectures, which benefit from a well-established optimization ecosystem \citep{dao2024transformers}. Hybrid architectures show what the combination buys. Jamba, which interleaves Mamba and attention layers at a ratio of seven to one and adds mixture-of-experts layers, reports approximately three times the throughput of Mixtral on a single A100 80GB in int8 at an 8K context with 512 output tokens \citep{lieber2024jamba}. Jamba-1.5 reports a key-value cache of 9 GB at a 256K context in 16-bit precision, against 80 GB for a 70B-parameter Transformer of comparable class \citep{team2024jamba}. The cache requirement, rather than the throughput multiplier, is the more durable advantage because it follows from the architecture rather than from a particular kernel. SSMs are also being explored for multimodal learning, where their efficiency could complement Transformer-based representations \citep{bansal2024comprehensive}. Future research directions include enhancing hardware optimization, improving selective state-space mechanisms, and expanding SSMs' applicability in multimodal and real-time processing tasks \citep{wang2024state}.

While these developments show the growing role of SSMs as a scalable alternative to Transformers, a consolidated analysis of their evolution, capabilities and limitations is needed to judge where they actually help. This paper is a structured review, and Section~\ref{sec:scope} states the scope, sources and selection criteria on which it rests. This review traces the lineage from S4 to Mamba-2 and the hybrids that follow it, and evaluates the design decisions along the way against Transformer baselines. It is organized around four objectives.
\vspace{-0.06in}

\begin{enumerate}
    \item \textbf{Lineage.} Follow the developmental path from continuous-time state-space theory and HiPPO through S4 and its diagonal successors to selective models, treating each step as a response to a specific limitation of the step before it.
    \item \textbf{Cross-cutting analysis.} Compare the model families along design dimensions that cut across them, namely selectivity, the scan and convolution views, diagonalization, inference-time cache behavior, hardware assumptions, context extrapolation and hybridization, rather than describing each model in isolation.
    \item \textbf{Evidence.} Separate training cost, autoregressive inference cost, and memory cost instead of collapsing them into a single complexity column, and report every performance and efficiency result together with the configuration under which it was obtained.
    \item \textbf{Applicability and limits.} Set out where SSMs are the better choice, where attention or a hybrid remains preferable, and what stands in the way of deployment.
\end{enumerate}

\subsection{Positioning relative to existing SSM and Mamba reviews}
\label{sec:positioning}

The state-space literature has been extensively reviewed, so any new survey must clearly define its distinct contribution. Table~\ref{tab:survey_comparison} sets this survey beside the main existing
ones. Read as a group, those surveys divide into two kinds. The first kind is
domain-scoped: vision \citep{zhang2024surveyvisualmamba, xu2024visualmamba}, medical
image analysis \citep{bansal2024comprehensive, heidari2024computation} and remote
sensing \citep{bao2025visionmambaremotesensing} each survey Mamba within one
application area, and each organizes by task. The second kind is a broad catalog
of models and results \citep{patro2024mamba, qu2024survey,
salam2025comprehensivemamba, wang2024state}, organized by block design, scanning
strategy or modality; the last of these frames the field explicitly as a search for a
replacement for the Transformer, which is a framing this review does not adopt and
Section~\ref{sec:crosscut} argues against. Both approaches are valuable, but the present paper takes a different analytical approach.

\setlength{\tabcolsep}{3pt}
\renewcommand{\arraystretch}{1.15}
\begin{longtable}{@{}>{\footnotesize\raggedright\arraybackslash}p{0.1568\linewidth} >{\footnotesize\raggedright\arraybackslash}p{0.0334\linewidth} >{\footnotesize\raggedright\arraybackslash}p{0.2090\linewidth} >{\footnotesize\raggedright\arraybackslash}p{0.0575\linewidth} >{\footnotesize\raggedright\arraybackslash}p{0.1097\linewidth} >{\footnotesize\raggedright\arraybackslash}p{0.0523\linewidth} >{\footnotesize\raggedright\arraybackslash}p{0.0753\linewidth} >{\footnotesize\raggedright\arraybackslash}p{0.2163\linewidth}@{}}
\caption{Comparison of this review with existing surveys of state-space and Mamba models.}
\label{tab:survey_comparison} \\
\toprule
\textbf{Survey} & \textbf{Year} & \textbf{Organizing basis} & \textbf{Covers to} & \textbf{Domain scope} & \textbf{Proto\-col} & \textbf{Depth} & \textbf{What it centers on} \\
\midrule
\endfirsthead
\ltconthead{8}
\toprule
\textbf{Survey} & \textbf{Year} & \textbf{Organizing basis} & \textbf{Covers to} & \textbf{Domain scope} & \textbf{Proto\-col} & \textbf{Depth} & \textbf{What it centers on} \\
\midrule
\endhead
\midrule
\ltcontfoot{8}
\endfoot
\bottomrule
\endlastfoot

Mamba-360 \citep{patro2024mamba} & 2025 &
Three model paradigms: gating, structural, recurrent; applications then grouped by domain &
Apr.\ 2024 & Multi-domain & No & Cata\-log &
Consolidating reported benchmark results across LRA, WikiText and ImageNet \\
\addlinespace

A Survey of Mamba \citep{qu2024survey} & 2026 &
Three axes: architecture advances, adaptation to data modalities, applications &
2026 (v8) & Cross-domain & No & Archi\-tecture &
How Mamba is adapted to each data modality \\
\addlinespace

State Space Model as a Transformer Alternative \citep{wang2024state} & 2024 &
Principles first, then existing state-space implementations, then applications grouped by domain &
Apr.\ 2024 & Multi-domain & No & Archi\-tecture &
Positioning state-space models as a general replacement for Transformers across modalities \\
\addlinespace

A Comprehensive Survey on Mamba \citep{salam2025comprehensivemamba} & 2025 &
Architectural variants of Mamba, then challenges &
Early 2025 & Architecture-centric & No & Cata\-log &
A short magazine-length orientation to the architecture and its open issues \\
\addlinespace

A Survey on Visual Mamba \citep{zhang2024surveyvisualmamba} & 2024 &
Backbone models separated from convolution-, recurrence- and attention-enhanced models; then by task &
Apr.\ 2024 & Vision only & No & Archi\-tecture &
Vision backbones and the task areas they serve \\
\addlinespace

Visual Mamba: A Survey and New Outlooks \citep{xu2024visualmamba} & 2024 &
Backbones, then applications by modality; scanning techniques classified separately &
Nov.\ 2024 & Vision only & No & Archi\-tecture &
Scanning strategy as the critical design choice for vision, over 200 papers \\
\addlinespace

Mamba for Medical Image Analysis \citep{bansal2024comprehensive} & 2024 &
Architecture first: pure Mamba, U-Net variants, and CNN, Transformer and GNN hybrids &
Oct.\ 2025 (v3) & Medical imaging & No & Archi\-tecture &
Medical architectures, scanning, datasets and results \\
\addlinespace

Computation-Efficient Era \citep{heidari2024computation} & 2024 &
Three-way classification by application, imaging modality and target organ &
Jun.\ 2024 & Medical imaging & \textbf{Yes} & Cata\-log &
The only prior survey in this group that states its databases and search phrases \\
\addlinespace

Vision Mamba in Remote Sensing \citep{bao2025visionmambaremotesensing} & 2025 &
Five dimensions from foundations through micro- and macro-architecture to benchmarking and open problems &
May 2025 & Remote sensing & No & Archi\-tecture &
About 120 remote-sensing studies, benchmarked by task \\
\addlinespace
\midrule
\addlinespace

\textbf{This review} & 2026 &
\textbf{Developmental lineage from HiPPO to Mamba-2, crossed with design dimensions that cut across families} &
2025 & Broad, plus dynamic and cross-modal settings & \textbf{Yes} & \textbf{Deri\-vation} &
\textbf{Why each family appeared, structured state-space duality as the connecting result, and cost separated into training, inference and cache} \\

\end{longtable}
\tabnote{``Protocol'' records whether the survey names the databases and search terms it used.
``Depth'' records the level of the technical treatment: \emph{catalog} tabulates models and reported
results, \emph{architecture} describes block structure and design choices, and \emph{derivation}
develops the underlying mathematics.}

Five distinctions help clarify the contribution of this survey and are stated in a form that readers can assess against the surveys discussed above. First, the organizing thread is the \emph{lineage} of state-space models rather than a taxonomy of completed architectures. Each family is presented in relation to a limitation or design challenge that motivated the next development: HiPPO provides a principled initialization, S4 makes the resulting operator computationally tractable, DSS and S4D show that much of this machinery can be simplified once the state matrix is diagonal, S5 restores parallelism through a scan, selectivity introduces content dependence while relaxing time invariance, and structured state-space duality recovers hardware efficiency affected by that change. Thus, alongside describing what the models are, the survey emphasizes the technical motivations underlying their progression. Second, structured state-space duality is examined as an important connecting result rather than only as a recent architectural development. Several earlier surveys predate Mamba-2, while others necessarily provide only limited discussion of it. Section~\ref{sec:ssd} therefore develops this connection in greater detail, and Section~\ref{sec:hybrids} uses it to interpret the operation of hybrid architectures. Third, computational cost is reported separately for training, autoregressive inference, and memory or cache behavior. These dimensions can differ substantially, and treating them as a single asymptotic measure may obscure the deployment relevance of a recurrent state whose size does not increase with context length. Fourth, each reproduced efficiency or accuracy result is accompanied by its measurement configuration and an evidence level. As discussed in Section~\ref{sec:empirical}, many reported results are based on kernel-level microbenchmarks conducted on particular hardware configurations and may not generalize directly across systems or deployment settings. Fifth, the technical discussion extends from architectural description to selected derivations, corresponding to the ``Depth'' column of Table~\ref{tab:survey_comparison}. In particular, Section~\ref{sec:ssd} derives the semiseparable representation of the state-space operator and the associated rank bound, which provides the basis for the capability discussions in Sections~\ref{sec:crosscut} and \ref{sec:limitations}. This derivation supports a more precise discussion of the structural conditions under which exact retrieval remains challenging for this model family. These distinctions are intended to complement, rather than replace, the strengths of the existing literature. This survey does not claim exhaustive coverage of any single application domain. For vision, medical imaging and remote sensing, the domain-specific surveys identified above remain the more comprehensive references and are cited accordingly rather than reproduced here.

\subsection{Scope, sources and paper selection}
\label{sec:scope}

This study adopts a structured review design rather than a systematic review in the technical sense. The review was not based on an exhaustive protocol-driven search and does not claim PRISMA compliance. The procedures described below define the scope of the literature corpus, the search strategy, and the criteria used to determine study eligibility. The literature search covered seven sources: Google Scholar, arXiv, DBLP, Semantic Scholar, Scopus, Web of Science and IEEE Xplore. Google Scholar, arXiv, DBLP and Semantic Scholar were used to identify the broad body of relevant literature, including conference papers and preprints that constitute a substantial share of research in this area. Scopus, Web of Science and IEEE Xplore were used to supplement coverage and establish publication status. Proceedings websites were also consulted when indexed records remained associated with preprint versions.

The search strategy combined the core terms `state space model'', `structured state space'', `selective state space'', `S4'' and `Mamba'' with architecture-related terms, including `hybrid'', `diagonal'', `duality'' and `scan''; task-related terms, including `long-range'', `language modeling'', `segmentation'', `forecasting'', `point cloud'' and `dynamic graph''; and modality-related terms, including `vision'', `speech'', `multimodal'' and ``medical''. Backward and forward citation searches were conducted from S4, Mamba and Mamba-2 to identify relevant studies not retrieved through keyword searches. Foundational work in control theory and recurrent modeling was included without a date restriction when necessary to establish the theoretical development of the field. Coverage of the modern state-space deep learning literature extends from HiPPO in 2020 through the end of 2025.

Studies were eligible for inclusion if they satisfied at least one of the following criteria: development of the theory of structured or continuous-time state-space sequence models; introduction of an S4-family or selective state-space architecture; integration of state-space layers with attention; adaptation of a state-space model to a specific data modality with comparison against a non-SSM baseline; or review of the state-space modeling literature. Adjacent subquadratic sequence models without a direct state-space lineage were included only where needed to provide technical context and are identified separately in Section~\ref{sec:adjacent}. Studies were excluded when the state-space component was incidental to the principal contribution, when no comparative baseline was reported, or when the work duplicated an already included publication. At the completion of the review, \ensuremath{21} cited works remained available only as preprints because no corresponding peer-reviewed version had been identified.

The final list comprises \ensuremath{103} works. Of these, \ensuremath{85} were published from 2020 onward and represent the core state-space literature, while the remaining \ensuremath{18} provide the necessary foundations in control theory, recurrent and convolutional networks and attention. Venue type was classified consistently from the bibliographic record: conference papers were identified through proceedings, journal articles through peer-reviewed journals, preprints where no peer-reviewed version was available, and all remaining works as books, book chapters or theses. Using this classification, the corpus includes \ensuremath{53} conference papers, \ensuremath{24} journal articles, \ensuremath{21} preprints and \ensuremath{5} books and book chapters. Figure~1 summarizes the temporal and venue-type distribution of the corpus.

\begin{figure}[pos=H]
\centering

\definecolor{VenConf}{HTML}{2A78D6}
\definecolor{VenJour}{HTML}{EB6834}
\definecolor{VenPre}{HTML}{1BAF7A}
\definecolor{VenBook}{HTML}{EDA100}

\def\hgtA{3.2}

\rmfamily
\resizebox{\linewidth}{!}{%
\begin{tikzpicture}[x=1cm, y=1cm,
    Gridline/.style={draw=black!12, line width=0.4pt},
    Axisline/.style={draw=black!45, line width=0.5pt},
    AxisLab/.style={font=\rmfamily\scriptsize, text=black},
    AxisTitle/.style={font=\rmfamily\small, text=black},
    ValLab/.style={font=\rmfamily\scriptsize, text=black},
    PanelTitle/.style={font=\rmfamily\small\bfseries, text=black}]

\begin{scope}
  \draw[Gridline] (0,0.000) -- (8.45,0.000);
  \node[AxisLab,anchor=east] at (-0.10,0.000) {0};
  \draw[Gridline] (0,1.000) -- (8.45,1.000);
  \node[AxisLab,anchor=east] at (-0.10,1.000) {10};
  \draw[Gridline] (0,2.000) -- (8.45,2.000);
  \node[AxisLab,anchor=east] at (-0.10,2.000) {20};
  \draw[Gridline] (0,3.000) -- (8.45,3.000);
  \node[AxisLab,anchor=east] at (-0.10,3.000) {30};
  \draw[Axisline] (0,0) -- (8.45,0);
  \draw[Axisline] (0,0) -- (0,3.32);
  \fill[VenConf,rounded corners=0.5pt] (0.420,0.000) rectangle (1.020,0.600);
  \fill[VenJour,rounded corners=0.5pt] (0.420,0.635) rectangle (1.020,1.235);
  \fill[VenPre,rounded corners=0.5pt] (0.420,1.270) rectangle (1.020,1.470);
  \fill[VenBook,rounded corners=0.5pt] (0.420,1.505) rectangle (1.020,1.905);
  \node[ValLab,anchor=south] at (0.720,1.960) {18};
  \node[AxisLab,anchor=north] at (0.720,-0.07) {$\leq$2019};
  \fill[VenConf,rounded corners=0.5pt] (1.440,0.000) rectangle (2.040,0.300);
  \fill[VenJour,rounded corners=0.5pt] (1.440,0.335) rectangle (2.040,0.535);
  \fill[VenPre,rounded corners=0.5pt] (1.440,0.570) rectangle (2.040,0.670);
  \fill[VenBook,rounded corners=0.5pt] (1.440,0.705) rectangle (2.040,0.805);
  \node[ValLab,anchor=south] at (1.740,0.860) {7};
  \node[AxisLab,anchor=north] at (1.740,-0.07) {2020};
  \fill[VenConf,rounded corners=0.5pt] (2.460,0.000) rectangle (3.060,0.400);
  \fill[VenJour,rounded corners=0.5pt] (2.460,0.435) rectangle (3.060,0.635);
  \node[ValLab,anchor=south] at (2.760,0.690) {6};
  \node[AxisLab,anchor=north] at (2.760,-0.07) {2021};
  \fill[VenConf,rounded corners=0.5pt] (3.480,0.000) rectangle (4.080,0.400);
  \fill[VenJour,rounded corners=0.5pt] (3.480,0.435) rectangle (4.080,0.535);
  \fill[VenPre,rounded corners=0.5pt] (3.480,0.570) rectangle (4.080,0.670);
  \node[ValLab,anchor=south] at (3.780,0.725) {6};
  \node[AxisLab,anchor=north] at (3.780,-0.07) {2022};
  \fill[VenConf,rounded corners=0.5pt] (4.500,0.000) rectangle (5.100,1.500);
  \fill[VenJour,rounded corners=0.5pt] (4.500,1.535) rectangle (5.100,1.735);
  \fill[VenPre,rounded corners=0.5pt] (4.500,1.770) rectangle (5.100,1.870);
  \node[ValLab,anchor=south] at (4.800,1.925) {18};
  \node[AxisLab,anchor=north] at (4.800,-0.07) {2023};
  \fill[VenConf,rounded corners=0.5pt] (5.520,0.000) rectangle (6.120,1.300);
  \fill[VenJour,rounded corners=0.5pt] (5.520,1.335) rectangle (6.120,1.735);
  \fill[VenPre,rounded corners=0.5pt] (5.520,1.770) rectangle (6.120,3.270);
  \node[ValLab,anchor=south] at (5.820,3.325) {32};
  \node[AxisLab,anchor=north] at (5.820,-0.07) {2024};
  \fill[VenConf,rounded corners=0.5pt] (6.540,0.000) rectangle (7.140,0.800);
  \fill[VenJour,rounded corners=0.5pt] (6.540,0.835) rectangle (7.140,1.335);
  \fill[VenPre,rounded corners=0.5pt] (6.540,1.370) rectangle (7.140,1.470);
  \node[ValLab,anchor=south] at (6.840,1.525) {14};
  \node[AxisLab,anchor=north] at (6.840,-0.07) {2025};
  \fill[VenJour,rounded corners=0.5pt] (7.560,0.000) rectangle (8.160,0.200);
  \node[ValLab,anchor=south] at (7.860,0.255) {2};
  \node[AxisLab,anchor=north] at (7.860,-0.07) {2026};
  \node[AxisTitle,anchor=north] at (4.22,-0.62) {Publication year};
  \node[AxisTitle,rotate=90,anchor=south] at (-0.66,1.60) {Number of references};
  \node[PanelTitle,anchor=south west] at (0,\hgtA+0.30) {(a)~~References by year and venue type};
\end{scope}

\begin{scope}[xshift=11.1cm, yshift=2.87cm]
  \draw[Axisline] (0,0.400) -- (0,-3.340);
  \fill[VenConf,rounded corners=0.5pt] (0,-0.230) rectangle (3.900,0.230);
  \node[AxisLab,anchor=east] at (-0.12,0.000) {Conference};
  \node[ValLab,anchor=west] at (4.000,0.000) {53~(51\%)};
  \fill[VenJour,rounded corners=0.5pt] (0,-1.210) rectangle (1.766,-0.750);
  \node[AxisLab,anchor=east] at (-0.12,-0.980) {Journal};
  \node[ValLab,anchor=west] at (1.866,-0.980) {24~(23\%)};
  \fill[VenPre,rounded corners=0.5pt] (0,-2.190) rectangle (1.545,-1.730);
  \node[AxisLab,anchor=east] at (-0.12,-1.960) {Preprint};
  \node[ValLab,anchor=west] at (1.645,-1.960) {21~(20\%)};
  \fill[VenBook,rounded corners=0.5pt] (0,-3.170) rectangle (0.368,-2.710);
  \node[AxisLab,anchor=east] at (-0.12,-2.940) {Book};
  \node[ValLab,anchor=west] at (0.468,-2.940) {5~(5\%)};
  \node[AxisTitle,anchor=north] at (1.95,-3.490) {Number of references};
  \node[PanelTitle,anchor=south west] at (-1.30,0.630) {(b)~~Totals by venue type};
\end{scope}

\begin{scope}[yshift=-1.42cm]
  \fill[VenConf,rounded corners=0.5pt] (3.550,-0.09) rectangle (3.890,0.09);
  \node[AxisLab,anchor=west] at (3.990,0) {Conference};
  \fill[VenJour,rounded corners=0.5pt] (5.750,-0.09) rectangle (6.090,0.09);
  \node[AxisLab,anchor=west] at (6.190,0) {Journal};
  \fill[VenPre,rounded corners=0.5pt] (7.950,-0.09) rectangle (8.290,0.09);
  \node[AxisLab,anchor=west] at (8.390,0) {Preprint};
  \fill[VenBook,rounded corners=0.5pt] (10.150,-0.09) rectangle (10.490,0.09);
  \node[AxisLab,anchor=west] at (10.590,0) {Book};
\end{scope}

\end{tikzpicture}%
}
\caption{Composition of the reference corpus by year and venue type.}
\tabnote{Counts are derived from the bibliography supplied with this submission. Works
published in 2019 or earlier are pooled, since they are foundational rather than part of the
state-space era. Color encodes venue type only.}
\label{fig:corpus}
\end{figure}

\subsection{Reproducibility Resources}
\label{sec:repo}

Given the pace of development in this area, a static reference list can become outdated quickly, while the methodological conditions attached to reported results are often lost when findings are reproduced across studies. To support continued verification and updating, the review corpus is released as a maintained open index at \url{https://github.com/pozapas/s4-to-mamba} under the MIT license.

The index extends beyond a conventional bibliography in four respects. First, papers are organized using the same model families and membership criteria defined in Section~\ref{sec:adjacent}, allowing the taxonomy to be examined and revised transparently. Second, each reported efficiency result is accompanied by the baseline, device, batch size and measurement scope provided in the original source, consistent with the reporting convention established in Section~\ref{sec:crosscut}. Third, the benchmark provenance documented in Section~9.1 is retained in full, including source attribution and averaging conventions. Fourth, the index includes dedicated resources for the dynamic and cross-modal settings discussed in Sections~\ref{sec:dyngraph} and \ref{sec:crossmodal}, distinguishing datasets and benchmarks on which state-space models have been evaluated from related problem settings for which evidence is not yet available. Contributions and corrections are accepted through the repository subject to verified metadata and explicit measurement conditions. Each entry also records whether a quantitative result originates from the primary authors or from an independent source, enabling subsequent revisions to be evaluated against the underlying evidence.

\section{Fundamentals of State Space Models (SSMs)}
\label{sec:foundations}
\subsection{Mathematical Foundation of SSMs}
        SSMs constitute a powerful mathematical approach broadly employed in diverse fields, including control theory, systems engineering, economics, and artificial intelligence, to describe complex dynamical systems. At their core, SSMs characterize the evolution of system states over time through first-order differential or difference equations, explicitly capturing the influence of external inputs and their relationship with observable outputs \citep{hangos2006analysis}. Distinguished from classical autoregressive models, SSMs offer substantial advantages, such as the intuitive handling of multivariate systems, straightforward interpretation of internal dynamics, and seamless transitions between continuous and discrete-time domains \citep{gu2021efficiently}. Recent advancements have expanded the computational efficiency and interpretability of SSMs by demonstrating equivalence among their continuous, recurrent, and convolutional representations, further solidifying their relevance in modern sequence modeling frameworks \citep{gu2021efficiently,wang2024state}.

        \subsubsection{General Formulation} 
\begin{figure}[pos=H]
            \centering
            \includegraphics[width=0.9\textwidth]{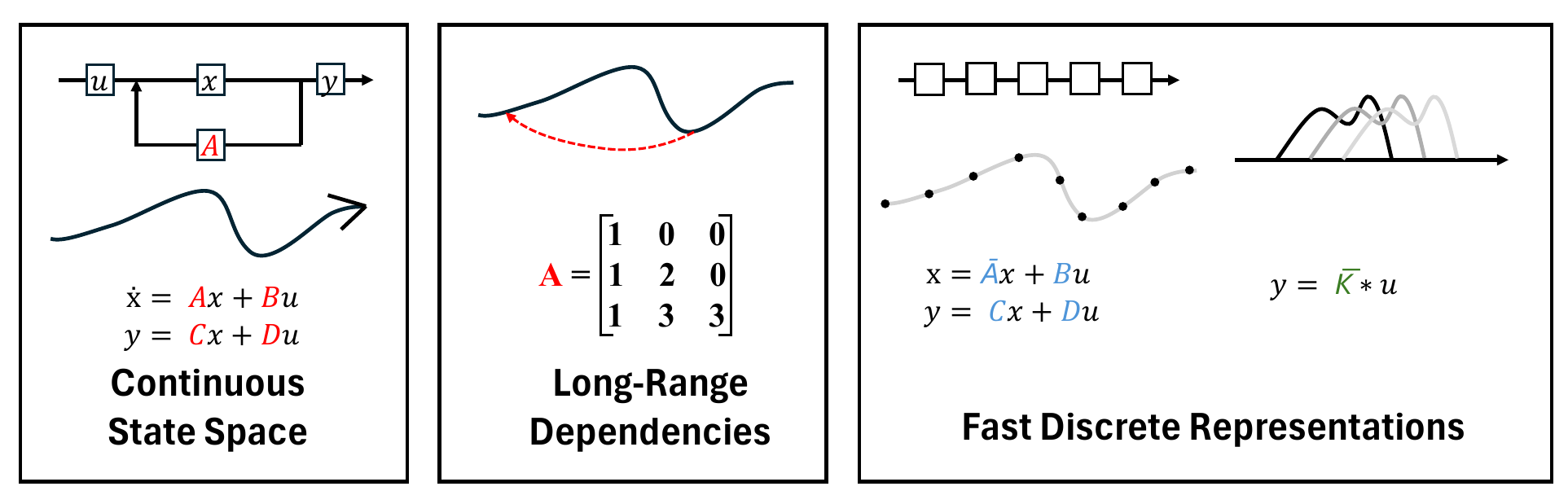}
            \caption{Conceptual Representation of State Space Models, adapted from \citep{gu2021efficiently}}
            
            \label{fig:Conceptual_Representation_of_SSMs}
        \end{figure}

Classical SSMs are mathematically defined by equations that describe the evolution of hidden states over time and the generation of observable outputs, as introduced by Kalman \citep{kalman1960new}. These models serve as a foundational framework for modeling dynamic systems across various scientific and engineering domains, such as control theory, signal processing, and economics. Classical SSMs encapsulate how a system's internal states evolve in response to external inputs and how these states influence observable outputs \citep{hangos2006analysis}. A notable strength of classical SSMs lies in their ability to explicitly represent both internal dynamics and external interactions using first-order differential equations, making them indispensable for capturing complex behaviors in dynamic systems. Figure~\ref{fig:Conceptual_Representation_of_SSMs} illustrates the fundamental structure of SSMs, showcasing three key perspectives: continuous-time state-space dynamics, long-range dependencies, and discrete-time representations.

        In the \textbf{left panel}, a continuous-time SSM is depicted, where an input \( u(t) \) influences the latent state \( x(t) \) via the system matrix \( A \), with matrices \( A, B, C, \) and \( D \) defining state evolution, input interaction, and output mapping.
        The \textbf{middle panel} illustrates how specialized \( A \) matrices enable deep SSMs to capture long-range dependencies, ensuring stable memory retention over long sequences, an improvement over traditional recurrent models.
        The \textbf{right panel} demonstrates how structured parameterization allows SSMs to transition into fast discrete representations, efficiently leveraging either a recurrent or convolutional framework. This enables scalability in deep learning applications like speech recognition, time-series forecasting, and NLP.
.        
        
        Recent advancements in deep learning have led to the development of SSMs, a specialized class of SSMs designed for efficient sequence modeling. Unlike classical SSMs, which focus on modeling physical and control systems, Structured SSMs such as S4, Mamba, and S5 leverage structured recurrence and diagonalized state-space representations to capture long-range dependencies in sequential data. These models have been evaluated in natural language processing, speech recognition and time-series forecasting, and the specific results supporting that evidence are collected with their measurement conditions in Table~\ref{tab:empirical}. Throughout this paper, we refer to "Structured SSMs" when discussing their role in deep learning, differentiating them from classical state-space models."

        \vspace{-1.5em}
        \begin{equation}
            \dot{x}(t) = A x(t) + B u(t), \quad y(t) = C x(t) + D u(t)
            \label{eq:state_space}
        \end{equation}
        
        Equation~\eqref{eq:state_space} represents the State-Space Representation of a dynamical system.

        Here, \( x(t) \in \mathbb{R}^{n} \) denotes the state vector encapsulating the system’s internal states at time \( t \), while \( u(t) \in \mathbb{R}^{m} \) represents the input or control vector externally influencing the system. The output vector \( y(t) \in \mathbb{R}^{p} \) describes observable quantities \citep{wang2024state}. The matrices \( A, B, C, D \) are constant and define system dynamics, input influence, state-to-output mapping, and direct input-to-output relationships, respectively. The structure and dimensions of these matrices depend on the complexity and nature of the system being modeled.
        
        where:
        \begin{itemize}
            \item \( x(t) \in \mathbb{R}^{n} \) is the state vector representing the internal state of the system=.
            \item \( u(t) \in \mathbb{R}^{m} \) is the input (control) vector influencing the state evolution=.
            \item \( y(t) \in \mathbb{R}^{p} \) is the output (observation) vector.
            \item \( A \in \mathbb{R}^{n \times n} \) is the state transition matrix, governing the dynamics of the system.
            \item \( B \in \mathbb{R}^{n \times m} \) is the control matrix, mapping input effects to the state.
            \item \( C \in \mathbb{R}^{p \times n} \) is the observation matrix, defining how the internal state maps to the output.
            \item \( D \in \mathbb{R}^{p \times m} \) is the feedthrough matrix, which directly relates the input to the output.
        \end{itemize}

        \subsubsection{Discrete-Time Representation of SSMs} SSMs originate from control theory, where they are traditionally formulated in continuous time using differential equations. However, for practical digital computation, deep learning applications, and real-time simulations, these continuous representations must be discretized. The process of converting continuous-time equations into discrete-time involves numerical discretization techniques such as the bilinear transform, Euler method, and Z-transform. Among these, the bilinear transform is often preferred due to its ability to preserve system stability while maintaining computational efficiency \citep{hamilton2020time,kailath1980linear}.
            
            A discrete-time SSM is mathematically defined as:
            \vspace{-0.5em} 
            \begin{equation}
                x_{t+1} = A_d x_t + B_d u_t, \quad y_t = C_d x_t + D_d u_t
            \end{equation}
            
            where \( A_d, B_d, C_d, D_d \) are the discretized counterparts of their continuous-time parameters. These parameters govern the evolution of the system's state over discrete time steps, enabling efficient sequential modeling of time-series data \citep{gu2021combining,gu2021efficiently}. The transformation from continuous to discrete form is commonly performed using matrix exponentials, formulated as:
            \vspace{-0.5em}
            \begin{equation}
                A_d = e^{A \Delta t}, \quad B_d = \int_{0}^{\Delta t} e^{A \tau} B d\tau
            \end{equation}
            
            where \( \Delta t \) represents the sampling interval used for discretization. This method ensures that the system’s temporal dynamics are accurately captured, allowing for efficient computation and numerical stability compared to simpler approximation methods \citep{smith2022simplified}.
            
            An alternative approach for discretization is using the bilinear transformation, which improves system stability while maintaining accuracy. The discrete transition matrix \( \bar{A} \) is computed as:
            \vspace{-0.5em}
            \begin{equation}
                \bar{A} = (I - \alpha \Delta t A)^{-1} (I + (1 - \alpha) \Delta t A)
            \end{equation}
            
            where:
            \begin{itemize}
                \item \( I \) is the identity matrix,
                \item \( A \) is the continuous-time state transition matrix,
                \item \( \alpha \) is a weighting parameter that determines the approximation method,
                \item \( \Delta t \) is the discretization step size.
            \end{itemize}
            
            This transformation ensures that the discrete-time system remains numerically stable while effectively approximating the continuous dynamics of the model. The bilinear transformation is particularly useful in machine learning applications, where preserving long-range dependencies and computational efficiency is crucial for model scalability.

            \begin{figure}[pos=H]
                \centering
                \includegraphics[width=0.9\textwidth]{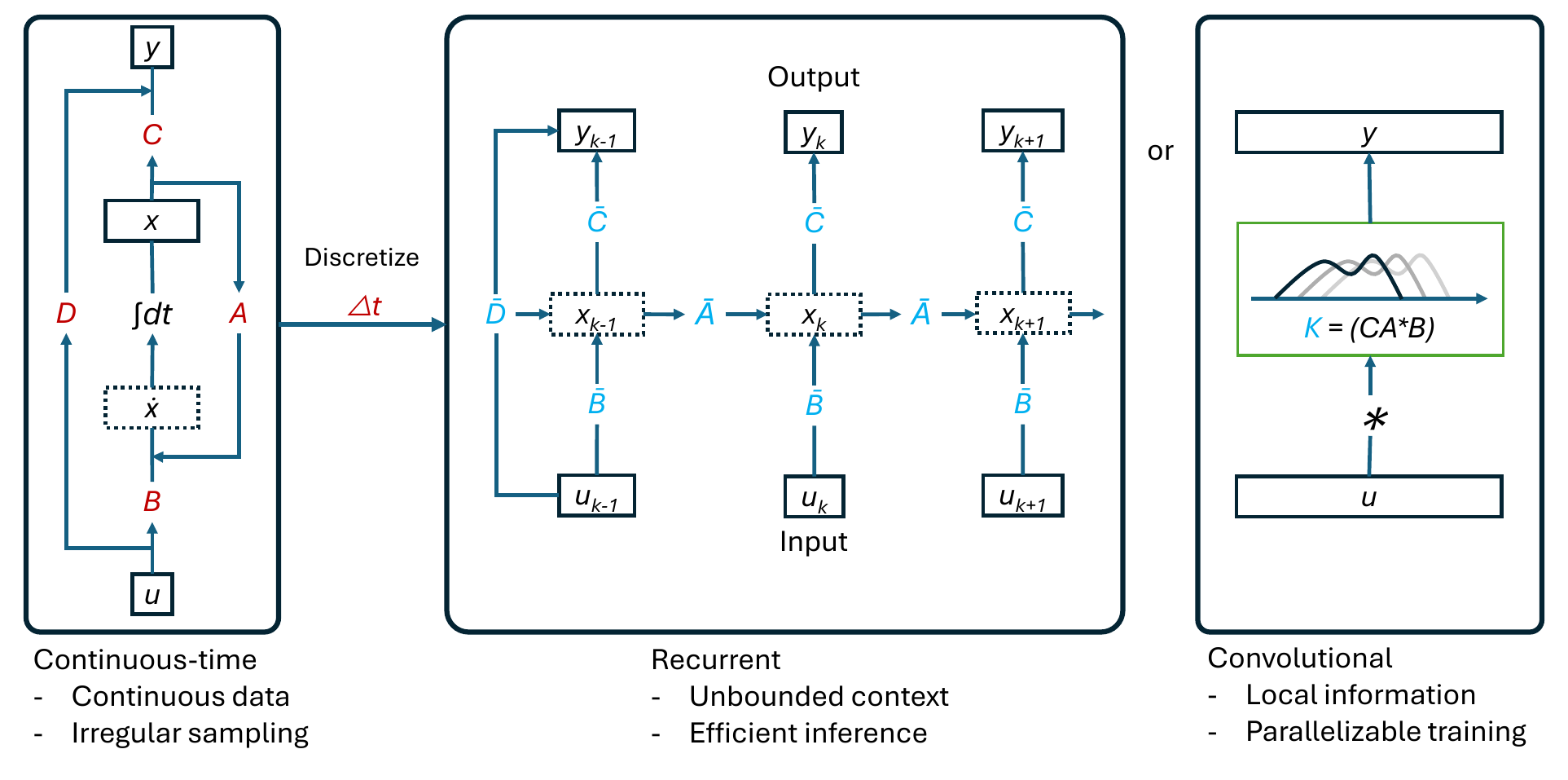}
                \caption{Three Views of Linear State Space Layer, adapted from \citep{gu2021combining}.}
                \label{fig:ssm_gu_lssl}
            \end{figure}

            As illustrated in Figure~\ref{fig:ssm_gu_lssl}, the discrete-time SSM can be implemented in either a recurrent or convolutional framework. In the recurrent formulation, the model updates its hidden state sequentially at each time step, following a structured recurrence similar to RNNs. This allows for effective memory retention and autoregressive sequence modeling, making it well-suited for applications like speech recognition, reinforcement learning, and finance \citep{gu2021combining,gu2021efficiently}. The structured recurrence also improves gradient propagation, addressing the vanishing gradient issue found in traditional RNNs.
            
            In contrast, the convolutional representation of discrete-time SSMs offers parallelized training and inference by applying a discretized convolutional kernel across the entire input sequence. This representation significantly reduces computational bottlenecks, making it ideal for large-scale sequence modeling tasks such as NLP, genomics, and time-series forecasting \citep{dao2024transformers}. Unlike standard convolutional neural networks (CNNs), which operate with fixed kernel sizes, convolutional SSMs use learned state-space parameterizations, allowing for more flexible and efficient long-sequence modeling.
            
            The dual formulation of discrete-time SSMs, recurrent and convolutional, demonstrates their versatility and adaptability to different computational constraints. Recurrent SSMs are more suitable for streaming and autoregressive tasks, where maintaining sequential order is essential. Convolutional SSMs, on the other hand, excel in batch processing and large-scale training, benefiting from depthwise computation and GPU acceleration. The choice between the two is therefore a deployment decision taken per phase rather than a property of the model, and Section~\ref{sec:crosscut} develops the rule for making it. As research continues to evolve, the discrete-time representation of SSMs is proving to be a powerful alternative to Transformers for many real-world applications. By leveraging structured state-space dynamics, these models can efficiently handle long sequences, bridging the gap between differential equation-based models, RNNs, and convolutional architectures \citep{gu2021combining}.

        \subsubsection{Convolutional and recurrent formulations} The convolutional perspective of SSMs transforms the input sequence through a structured, linear state transition, allowing for parallel computation over long sequences. As depicted in the provided image, SSMs can be rewritten in a discretized convolutional form, where the latent state evolution is governed by a transition matrix \( A \) and input transformation matrix \( B \) \citep{gu2021combining}. This results in a system response that can be computed efficiently using the following discrete-time formulation:
            \vspace{-0.5em}
            \begin{equation}
                h_t = A h_{t-1} + B u_t, \quad y_t = C h_t + D u_t
            \end{equation}
            
            where:
            \begin{itemize}
                \item \( h_t \) represents the hidden state,
                \item \( u_t \) is the input at time step \( t \),
                \item \( A, B, C, D \) are learned parameters that define state transitions.
            \end{itemize}
            
            In the convolutional interpretation, the state-space model behaves as an implicit filter, where the system response is equivalent to convolving the input sequence with a learned kernel:
            \vspace{-0.5em}
            \begin{equation}
                K = (C A^* B)
            \end{equation}
            
            This formulation enables efficient parallelization, similar to CNNs, making SSMs highly scalable for long-sequence processing \citep{gu2021combining, dao2024transformers}. Unlike standard CNNs, where the kernel size is fixed, SSMs implicitly parameterize kernels through structured matrices, offering a more flexible and learnable representation.

            Convolutional SSMs offer several advantages over traditional recurrent architectures. One key benefit is parallel computation: unlike RNNs, which require sequential updates, convolutional SSMs can process entire sequences in parallel, significantly reducing training time, making them efficient for handling large-scale datasets. Morover, another advantage is their ability to model long-range dependencies effectively. The implicit kernel representation enables better memory retention over long sequences, making these models well-suited for applications such as genomics, video processing, and speech recognition. By leveraging structured matrices, SSMs learn a kernel whose effective length is not fixed in advance, which is the specific respect in which they are less constrained than a convolution with a fixed kernel size. Finally, hardware efficiency is a crucial factor. Convolutional SSMs utilize structured matrix multiplications, allowing them to efficiently exploit modern hardware accelerators such as GPUs and TPUs. This computational advantage makes them ideal for high-performance machine learning applications where both scalability and efficiency are critical.

                The recurrent view is the same layer read the other way, and the two are worth setting side by side rather than in separate subsections, because Section~\ref{sec:crosscut} argues that choosing between them is a deployment decision taken per phase. In contrast to the convolutional approach, the recurrent perspective of SSMs treats the hidden state as an evolving memory, similar to RNNs but with structured linear recurrence. This formulation explicitly updates the state at each time step, preserving sequential information flow without the need for explicit convolutional filtering.
                
                The recurrent form of an SSM follows:
                \vspace{-0.5em}
                \begin{equation}
                    h_t = A h_{t-1} + B u_t \quad  y_t = C h_t
                \end{equation}
                
                where the state \( h_t \) is updated at each time step rather than applying a global convolution over the input. This design shares similarities with traditional RNNs but introduces structured state transitions that mitigate issues such as vanishing gradients \citep{gu2021combining, patro2024mamba}.
                
                A key difference between SSM recurrence and classical RNNs is that SSMs employ structured matrices for the transition dynamics. Unlike vanilla RNNs, where the hidden state evolution is unconstrained, SSMs impose a structured transformation on \( A \), allowing for more stable long-range dependency modeling \citep{gu2021combining, dao2024transformers}.
                
                Recurrent SSMs offer several advantages over conventional sequence models. First, they reduce computational complexity compared to Transformers, which scale quadratically with sequence length. SSMs maintain a linear complexity of \( O(T) \), making them efficient for long sequences. Second, their use of structured state-space matrices constrains the spectrum of the recurrence, which slows rather than eliminates the gradient decay that affects standard RNNs, as Section~\ref{sec:crosscut} discusses. Finally, SSM recurrence enables efficient sequential processing while preserving long-range dependencies. Whether that translates into an accuracy advantage is task dependent, and Section~\ref{sec:empirical} reports the evidence with the configurations under which it was obtained rather than asserting a general advantage.

        \subsubsection{Related sequence-modeling architectures}

        The three architectures that state-space layers are measured against are summarized here in the terms the comparison of Table~\ref{tab:ssm_comp} uses, namely cost, parallelism and what each can and cannot represent.

        \textbf{Recurrent neural networks.} RNNs have a rich history dating back to the early work of Elman \citep{elman1990finding}, who introduced the foundational \textit{Elman network} as a simple recurrent model designed for sequence processing. RNNs were developed to handle sequential data by employing recurrent connections that allow the hidden state to retain information from previous time steps, effectively capturing temporal dependencies \citep{patro2024mamba}. Early RNN models, including \textit{Jordan networks} \citep{jordan1997serial} and \textit{Elman networks} \citep{elman1990finding}, established the core principle of using hidden states to model sequential dependencies \citep{gu2021combining}. However, traditional RNNs struggled with vanishing gradients when learning long-range dependencies, leading to the development of architectures such as \textit{LSTM} by by Hochreiter \& Schmidhuber \citep{hochreiter1997long}.
        Mathematically, the hidden state \( h_t \) at time step \( t \) is updated based on the current input \( x_t \) and the previous hidden state \( h_{t-1} \), following the equation:
        \vspace{-0.5em}
        \begin{equation}
            h_t = \sigma(W_h h_{t-1} + W_x x_t + b)
        \end{equation}
        
        where \( W_h \), \( W_x \), and \( b \) are trainable parameters, and \( \sigma \) is typically a nonlinear activation function such as the hyperbolic tangent (\textit{tanh}). Despite their theoretical strength in capturing temporal dependencies, RNNs suffered from practical issues like the \textit{vanishing} and \textit{exploding gradient} problems, which severely restricted their effectiveness in modeling long sequences \citep{hochreiter1997long}. These limitations necessitated improvements, eventually leading to the introduction of gated architectures such as LSTMs and Gated Recurrent Units (GRUs) \citep{gu2023mamba}.

        \textbf{Convolutional neural networks.} CNNs, first introduced by \cite{lecun1989backpropagation}, were initially designed for computer vision tasks but later adapted for sequential data processing. CNNs leverage convolutional layers to efficiently capture local patterns by applying \textit{kernels (filters)} across input data. In sequence modeling, a \textit{one-dimensional convolution operation} is commonly defined as:
        \vspace{-0.5em}
        \begin{equation}
            y_i = \sum_{j=0}^{k-1} x_{i+j} w_j + b
        \end{equation}
        
        where \( x \) represents the input sequence, \( w \) denotes the kernel weights, \( b \) is a bias term, and \( k \) is the kernel size. CNNs exhibit \textit{high computational efficiency} due to their parallelization capabilities, making them more scalable than RNNs. However, their \textit{limited receptive fields} inherently restrict their ability to model long-range dependencies, requiring deeper architectures or advanced techniques such as \textit{dilated convolutions} to extend their contextual reach \citep{lecun1998gradient}.

        \textbf{Transformers.} The Transformer architecture, introduced by Vaswani et al.\citep{vaswani2017attention}, revolutionized sequence modeling, particularly in natural language processing, by introducing \textit{self-attention} mechanisms. Self-attention allows each element in a sequence to directly attend to every other element, effectively capturing \textit{global dependencies}. The core mathematical operation of self-attention is defined as:
        \vspace{-0.5em}
        \begin{equation}
            \text{Attention}(Q, K, V) = \text{softmax} \left(\frac{QK^T}{\sqrt{d_k}}\right) V 
        \end{equation}
        \vspace{-0.5em}
        \begin{equation}
            Q = W_q X, \quad K = W_k X, \quad V = W_v X
        \end{equation}

        where \( Q \), \( K \), and \( V \) are linear transformations of the input, representing \textit{queries}, \textit{keys}, and \textit{values}, respectively, and \( d_k \) is the dimensionality of the keys. Transformers mitigate the long-range dependency issues observed in RNNs and CNNs, enabling efficient parallel computation. However, they incur a quadratic computational complexity \( O(T^2) \) with respect to sequence length, posing scalability challenges for extremely long sequences \citep{vaswani2017attention}.

        \textbf{State space models.} SSMs originate from control theory and signal processing \citep{kalman1960new}, but recent advancements have positioned them as a subquadratic alternative to Transformers for sequence modeling, particularly for very long sequences. Unlike Transformers, which have quadratic complexity due to self-attention operations, SSMs provide a structured linear representation of latent state dynamics, reducing the asymptotic training cost and enabling efficient parallel computation \citep{gu2021efficiently,gu2021combining, wang2024state}. The mathematical formulation, including continuous and discrete-time representations and discretization techniques, is given in Section~\ref{sec:foundations} above; Section~\ref{sec:crosscut} takes up the question of when that asymptotic advantage is the one that binds in practice.

\subsection{Comparison with Other Sequence Models}

A concise comparison of RNNs, Transformers, CNNs, and SSMs in terms of computational complexity, strengths, and weaknesses is provided in Table \ref{tab:ssm_comp}.

RNNs process sequences efficiently with linear complexity \( O(T) \), updating hidden states at each step based on prior states and inputs. However, they accumulate unnecessary information over long sequences, leading to inefficiencies and vanishing gradient issues, which hinder their ability to model long-range dependencies effectively \citep{hochreiter1997long}. While LSTMs and GRUs introduce gating mechanisms to mitigate these limitations, they still scale poorly with sequence length due to their reliance on storing past states \citep{cahuantzi2023comparison}.

Transformers \citep{vaswani2017attention} address this limitation by using self-attention to model global interactions across sequence elements. However, this flexibility comes at a cost: quadratic complexity \( O(T^2) \), making Transformers computationally expensive, particularly for long sequences. CNNs, originally designed for vision tasks, have been adapted for sequence modeling using 1D convolutions over temporal data \citep{bhat2024mathematical}. CNNs efficiently capture local patterns in sequences, making them computationally efficient. However, their fixed kernel sizes limit their ability to model long-range dependencies unless large receptive fields or stacked layers are used, which increases computational cost \citep{agrawal2020using}.

In contrast, SSMs, including architectures such as S4 and Mamba, offer an efficient alternative with linear complexity \( O(T) \) while maintaining strong long-range dependency modeling. Unlike Transformers, which rely on self-attention, SSMs utilize structured state transitions to model dependencies more efficiently \citep{gu2021combining,gu2021efficiently}. Moreover, Selective SSMs dynamically update their hidden states, retaining only relevant information rather than storing full past states like RNNs. This enables efficient memory compression, reducing computational overhead while maintaining scalability \citep{muca2024theoretical}. Unlike CNNs, which rely on fixed receptive fields and struggle with long-range dependencies, SSMs do not require predefined kernel sizes and can adaptively model dependencies across varying timescales \citep{ding2024dygmamba}. Their structured approach ensures scalability and efficiency, making them a strong contender for sequence modeling tasks that demand both long-term memory and computational efficiency.

By balancing memory retention, efficiency, and scalability, Selective SSMs provide a more flexible and computationally efficient framework than RNNs, CNNs, and Transformers. Their ability to selectively retain critical information while discarding irrelevant data makes them particularly effective for long-sequence tasks such as NLP, time-series forecasting, and signal processing.

\setlength{\tabcolsep}{3pt}
\renewcommand{\arraystretch}{1.2}
\begin{longtable}{@{}>{\footnotesize\raggedright\arraybackslash}p{0.0989\linewidth} >{\footnotesize\raggedright\arraybackslash}p{0.0989\linewidth} >{\footnotesize\raggedright\arraybackslash}p{0.0659\linewidth} >{\footnotesize\raggedright\arraybackslash}p{0.0829\linewidth} >{\footnotesize\raggedright\arraybackslash}p{0.1223\linewidth} >{\footnotesize\raggedright\arraybackslash}p{0.1223\linewidth} >{\footnotesize\raggedright\arraybackslash}p{0.1436\linewidth} >{\footnotesize\raggedright\arraybackslash}p{0.1755\linewidth}@{}}
\caption{Sequence-model families compared along separated cost axes.}
\label{tab:ssm_comp} \\
\toprule
\textbf{Family} &
\textbf{Training cost} &
\textbf{Infer.\ cost / token} &
\textbf{Decoding state} &
\textbf{Execution mode (train / decode)} &
\textbf{Hardware assumption} &
\textbf{Where it is strong} &
\textbf{Where it fails} \\
\midrule
\endfirsthead
\ltconthead{8}
\toprule
\textbf{Family} &
\textbf{Training cost} &
\textbf{Infer.\ cost / token} &
\textbf{Decoding state} &
\textbf{Execution mode (train / decode)} &
\textbf{Hardware assumption} &
\textbf{Where it is strong} &
\textbf{Where it fails} \\
\midrule
\endhead
\midrule
\ltcontfoot{8}
\endfoot
\bottomrule
\endlastfoot

\textbf{RNN / LSTM} \citep{hochreiter1997long, cahuantzi2023comparison} &
$O(Td^2)$ &
$O(d^2)$ &
$O(d)$, constant &
Sequential recurrence in both phases &
None beyond dense matrix-vector products; poor device utilization &
Streaming; small constant memory &
Gradient decay over long horizons; no parallelism in $T$, so wall-clock training scales badly \\
\addlinespace

\textbf{CNN} \citep{agrawal2020using} &
$O(Tkd^2)$ &
$O(kd^2)$ &
$O(kd)$, constant &
Parallel convolution / sliding window &
Standard convolution kernels; well served by matrix units &
Local structure; fully parallel &
Receptive field grows only with depth, so long-range dependencies need many layers \\
\addlinespace

\textbf{Transformer} \citep{vaswani2017attention} &
$O(T^2d)$ &
$O(Td)$, grows with context &
$O(Td)$, \textbf{grows linearly with context} &
Parallel attention / cached incremental attention &
Matrix units plus sufficient memory bandwidth for the cache; benefits from fused attention kernels &
Exact retrieval and in-context learning; mature tooling &
Quadratic training cost; the key-value cache is the binding constraint at long context \\
\addlinespace
\midrule
\addlinespace

\textbf{S4 family, LTI} \citep{gu2021efficiently, gupta2022diagonal, gu2022parameterization} &
$O(Td\log T)$ via FFT convolution &
$O(Nd)$ &
$O(Nd)$, constant &
Global convolution by FFT / step-by-step recurrence &
Efficient long FFT; the two phases run different kernels, so both must be implemented &
Long smooth signals; strong on LRA &
Time-invariant, so it cannot condition on content; weak on token-level filtering and copying \\
\addlinespace

\textbf{S5} \citep{smith2022simplified} &
$O(TNd)$ via parallel scan &
$O(Nd)$ &
$O(Nd)$, constant &
Associative parallel scan / recurrence &
Scan primitive with good memory locality; no FFT required &
Removes the FFT while keeping parallelism; multi-input multi-output state &
Still time-invariant, so the same content-dependence limit applies \\
\addlinespace

\textbf{Selective SSM (Mamba)} \citep{gu2023mamba} &
$O(TN^2)$ via selective scan &
$O(Nd)$ &
$O(Nd)$, constant &
Fused selective scan / recurrence &
Custom fused kernel and a memory hierarchy that keeps the state on chip; does \emph{not} reach matrix units &
Content-dependent gating of a fixed-size state; constant-cost decoding &
A fixed state bounds exact recall; scan kernels leave matrix units idle \\
\addlinespace

\textbf{Mamba-2 / SSD} \citep{dao2024transformers} &
$O(TN^2)$, dominated by matrix products &
$O(Nd)$ &
$O(Nd)$, constant &
Block decomposition of a semiseparable matrix / recurrence &
Matrix units, the same assumption attention makes; this is what the duality buys &
Same asymptotics as Mamba at much higher hardware utilization, permitting larger $N$ &
State transition restricted to scalar times identity, a deliberate loss of expressivity \\
\addlinespace

\textbf{SSM-attention hybrid} \citep{lieber2024jamba, ren2025samba, dong2025hymba} &
Between the two, set by the attention fraction &
Between the two &
Constant part plus a small cache from the attention layers &
Both mechanisms in one pass, by layer or in parallel heads &
Both sets of kernels must be present and interleaved efficiently &
Recovers exact retrieval at a bounded cache cost &
Inherits the tuning burden of both; the attention fraction is another hyperparameter \\
\addlinespace

\textbf{Adjacent subquadratic} \citep{peng2023rwkv, sun2023retnet, poli2023hyena} &
$O(Td^2)$ or $O(Td\log T)$ &
$O(d)$, or convolution history &
$O(d)$ constant, or none &
Parallel or chunkwise form / recurrence, where one exists &
Varies; the convolutional members need long FFTs, the recurrent members need a scan &
Constant-time decoding without a state-space derivation &
Mixing is fixed by position rather than learned from content, except in the gated members \\

\end{longtable}
\tabnote{``Execution mode'' records how the sequence is traversed at training and at decoding time, since several families change mechanism between the two. ``Hardware assumption'' records what a family needs from the device to reach its stated cost. $T$ is the sequence length, $N$ the state size, $d$ the model width and $k$ the convolution kernel width.}

\section{The Foundation of Modern SSMs}
\subsection{Introduction to S4}

The advancement of deep learning has led to remarkable progress in sequence modeling tasks such as NLP, time series forecasting, and speech recognition. However, conventional models like RNNs and transformers face significant challenges when handling extremely long sequences. RNNs struggle with vanishing and exploding gradients, making them inefficient for long-range dependencies \citep{bengio1994learning, hochreiter1997long}. Meanwhile, transformers, which rely on self-attention mechanisms, suffer from quadratic complexity $O(N^2)$ for sequence length, limiting their scalability for long-sequence modeling \citep{vaswani2017attention}. To address these limitations, structured SSMs have gained attention as an alternative, particularly the S4 model, which provides a novel approach for handling long-range dependencies efficiently \citep{gu2021efficiently}.

S4 is a structured state-space model that significantly improves upon previous RNN and attention-based methods by leveraging continuous-time state-space formulations \citep{gu2022train}. The foundation of S4 lies in classical state-space models, which have been extensively used in control theory and signal processing \citep{kalman1960new, ljung1987theory}. These models define system dynamics through differential equations, which are then discretized for computation \citep{chen2018neural}. While traditional SSMs were impractical for deep learning due to their computational overhead, S4 introduces a structured formulation that allows for fast and scalable computation \citep{gu2020hippo}. This innovation enables S4 to efficiently capture long-range dependencies without suffering from gradient degradation, making it a promising solution for long-sequence modeling tasks.

\subsubsection{First Structured SSM Designed for Long-Sequence Modeling}

S4 represents a paradigm shift in sequence modeling by introducing a mathematically grounded and computationally efficient alternative to traditional architectures. A key innovation in S4 is the use of the High-Order Polynomial Projection Operator (HiPPO) framework, which provides a mathematically rigorous approach to preserving long-range dependencies in sequential data~\citep{gu2020hippo}. HiPPO enables continuous memory retention through an optimal polynomial projection mechanism, addressing the limitations faced by earlier sequence models such as LSTMs~\citep{hochreiter1997long}, GRUs~\citep{chung2014empirical}, and vanilla RNNs~\citep{bengio1994learning}. 

\begin{figure}[pos=H]
    \centering
    \includegraphics[width=0.98\linewidth]{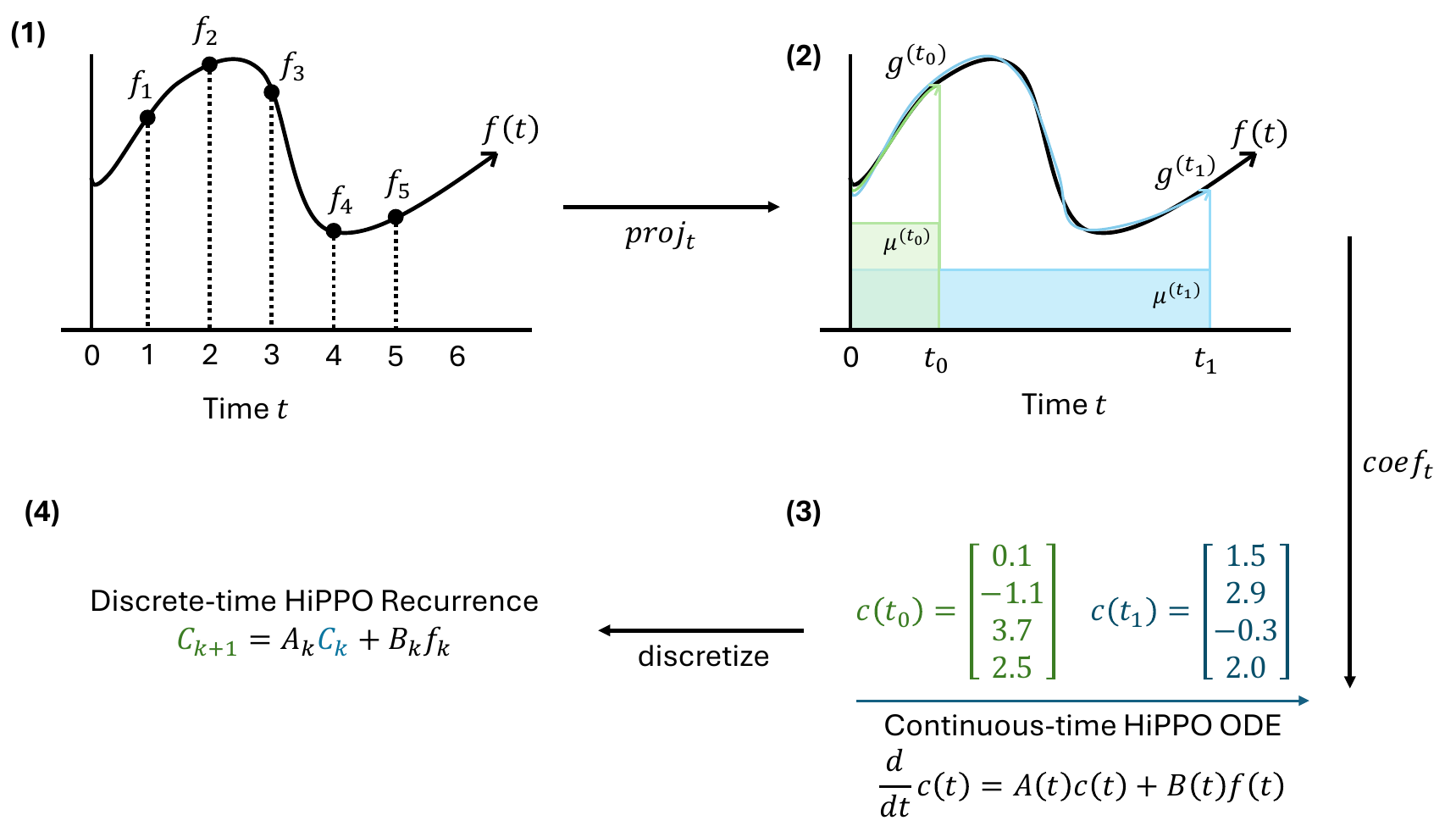}
    \caption{HiPPO Framework for Online Function Approximation, 
    adapted from \citep{gu2020hippo}}
    \label{fig:HiPPO}
\end{figure}

Figure \ref{fig:HiPPO} illustrates the HiPPO framework, which optimally projects past information onto a polynomial basis to maintain a compressed yet expressive memory representation. In this process, a function 
$f(t)$ is approximated by projecting it onto polynomial bases using a weighting measure $\mu(t)$. The function's history is then represented as a set of evolving coefficients governed by a structured ODE:
\vspace{-0.5em}
\[
\frac{d}{dt} c(t) = A(t)c(t) + B(t)f(t).
\]

To ensure efficient computation, this continuous ODE is discretized into a recurrence relation:
\vspace{-0.5em}
\[
c_{k+1} = A_k c_k + B_k f_k.
\]

This approach allows sequence models to retain information across long time horizons efficiently, a key principle that later influenced S4’s structured state-space design.

Unlike previous architectures that relied on explicit gating mechanisms or self-attention, S4 leverages structured matrices for efficient long-term information retention~\citep{katharopoulos2020transformers, tay2022efficient}. Another breakthrough of S4 is its ability to achieve subquadratic computational complexity while maintaining high expressivity. Transformers, while powerful, require $\mathcal{O}(N^2)$ complexity due to self-attention computations, making them infeasible for very long sequences~\citep{brown2020language, vaswani2017attention}. In contrast, S4 achieves $\mathcal{O}(N \log N)$ complexity by leveraging structured state-space computations, significantly reducing computational overhead~\citep{gu2022train}. This advantage has positioned S4 as a more scalable alternative to transformers, particularly in time series forecasting, speech recognition, and text processing~\citep{bain2021frozen, lim2021temporal}.

S4’s efficiency is also attributed to its use of diagonal plus low-rank parameterization, which allows it to approximate long-sequence dependencies while remaining computationally stable~\citep{gu2022train}. This contrasts with previous state-space representations that required expensive matrix multiplications, making them impractical for deep learning applications. By introducing structured recurrence, S4 can effectively learn from long sequential data without explicit recurrence mechanisms like those in standard RNNs~\citep{graves2013speech}. One of the most important contributions of S4 is its compatibility with modern hardware accelerators, such as GPUs and TPUs, which enable efficient parallelization~\citep{raffel2020exploring, shoeybi2019megatron}. Many earlier sequence models, including attention-based transformers, require significant memory overhead, limiting their feasibility for long-sequence processing~\citep{raffel2020exploring}. S4, however, benefits from its structured state-space computations, allowing it to handle sequences of tens of thousands of tokens without excessive memory consumption~\citep{dai2019transformer, wang2020linformer}.

Beyond efficiency, S4 was applied across several domains in the period following its release. It has been used for long-range time-series forecasting, where the comparison of interest is against attention-based forecasters such as the temporal fusion transformer~\citep{lim2021temporal}, and for biomedical signal analysis, where long electroencephalogram and electrocardiogram recordings are exactly the regime in which a linear-time model is attractive~\citep{tang2023modeling}. These are reported results in individual studies rather than controlled comparisons across a common protocol, and Section~\ref{sec:empirical} explains why that distinction matters throughout this literature. Despite its advantages, implementing S4 is not without challenges. One of the primary difficulties is the optimization of structured state-space matrices, which requires specialized initialization techniques to maintain numerical stability~\citep{gu2022train}. Unlike standard architectures such as transformers, where training dynamics are well-understood, S4 introduces complexities in parameter tuning that necessitate novel training techniques such as spectral normalization~\citep{miyato2018spectral}.

S4’s success has inspired the development of several extensions. Structured State-Space Sequence Model with Simplified Representations (S5) introduces a streamlined version of S4 with fewer hyperparameters, improving ease of implementation~\citep{smith2022simplified}. Additionally, hybrid models that combine S4 with self-attention mechanisms have been explored to further enhance their modeling capabilities~\citep{smith2023convolutional}. These developments indicate a growing interest in structured state-space approaches as a replacement for traditional sequence models.

\subsection{Key Innovations in S4}

The S4 model introduces several groundbreaking innovations that distinguish it from traditional sequence models such as RNNs, LSTMs, and Transformers. These innovations primarily revolve around structured recurrence, efficient convolutional formulations, and FFT acceleration, all of which contribute to making S4 scalable, efficient, and well-suited for long-sequence modeling tasks~\citep{gu2021efficiently}.

\subsubsection{State-space Formulation (Structured Recurrence)}
One of the most fundamental innovations in S4 is its state-space formulation, which provides a structured alternative to traditional recurrence-based models. Unlike standard RNNs, which rely on hidden states that evolve sequentially over time, S4 models sequences using continuous-time state-space equations, allowing for a more mathematically rigorous representation of long-range dependencies~\citep{gu2022train}. Traditional SSMs have long been used in control theory, signal processing, and dynamical systems, but their direct application to deep learning has been limited due to their computational inefficiency~\citep{kalman1960new, ljung1987theory}. S4 overcomes these challenges by introducing a structured representation that enables efficient parallelization while retaining the expressiveness of SSMs~\citep{chen2018neural, lim2021temporal}. The core equation governing S4’s structured recurrence is:
\vspace{-0.5em}
\begin{equation}
    \dot{x}(t) = A x(t) + B u(t)
\end{equation}
\vspace{-2.5em}
\begin{equation}
    y(t) = C x(t) + D u(t)
\end{equation}

where $A$, $B$, $C$, and $D$ are learnable structured matrices that define the evolution of the internal state $x(t)$ over time~\citep{gu2021efficiently}. Unlike conventional RNNs that require sequential updates of hidden states, S4 converts these state-space operations into an efficient convolutional representation, allowing for fast parallel computations~\citep{gu2022train, gu2021combining}. The advantage of this structured formulation is that it implicitly maintains long-range dependencies, addressing one of the fundamental weaknesses of both recurrent architectures and self-attention mechanisms~\citep{katharopoulos2020transformers, wang2020linformer}.

\begin{figure}[pos=H]
    \centering
    \includegraphics[width=0.98\linewidth]{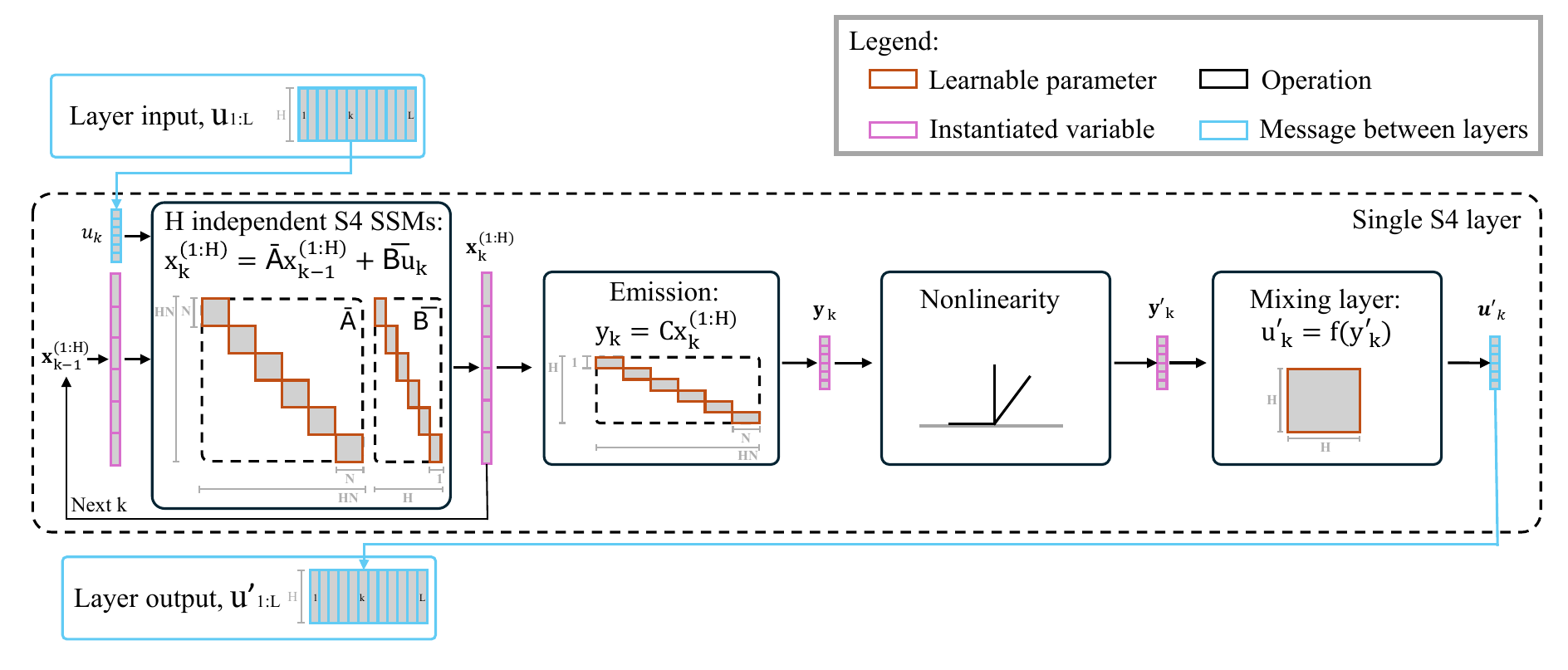}
    \caption{S4 Layer Structure adapted from \citep{smith2022simplified}}
    \label{fig:S4_Layer}
\end{figure}

This structured recurrence mechanism is illustrated in Figure~\ref{fig:S4_Layer}, which provides an overview of the S4 layer architecture. The model is formulated as a large state-space system (SSM) with state size \( HN \), where each independent SSM operates in parallel, leveraging block-diagonal state, input, and output matrices for efficient sequence modeling. The architecture integrates HiPPO-based initialization to approximate long-range dependencies and employs a Diagonal Plus Low-Rank parameterization to derive an optimized convolutional kernel. This parameterization enables efficient memory retention and fast sequence processing while reducing computational complexity. By structuring the recurrence in this manner, S4 efficiently models long-range dependencies with reduced memory overhead, making it a scalable and expressive alternative to traditional sequence modeling architectures.

\subsubsection{Efficient Convolutions Replacing RNN-style Recurrence}
S4 replaces traditional RNN-style recurrence with convolutional operations, allowing it to handle long sequences in a much more efficient manner~\citep{gu2021efficiently}. While classical recurrent architectures update states sequentially, S4 instead applies a convolutional filter over the input sequence, which effectively replaces recurrence with a learned convolutional response. This shift is significant because convolutions are inherently parallelizable, unlike recurrence, which requires sequential dependencies. In contrast, RNNs update hidden states one step at a time, preventing parallel execution and leading to inefficiencies when processing long sequences~\citep{hochreiter1997long}. The convolutional nature of S4 also enables smoother long-range dependencies, allowing the model to capture patterns spanning thousands of tokens without requiring explicit memory gating mechanisms like those used in LSTMs or GRUs~\citep{chung2014empirical}. This makes S4 particularly advantageous for tasks such as video understanding~\citep{bain2021frozen}, protein sequence modeling~\citep{jumper2021highly}, and large-scale language modeling~\citep{gu2022train}.

\subsubsection{FFT Acceleration for Parallelized Computations}
Another major innovation in S4 is its use of FFT acceleration, which enables highly efficient sequence processing~\citep{gu2021efficiently}. By converting state-space operations into the frequency domain using FFT, S4 significantly reduces computational overhead and improves scalability. FFT acceleration allows S4 to compute convolutions in $\mathcal{O}(N \log N)$ time, making it dramatically faster than traditional sequence models. This optimization is particularly important in real-time applications, such as speech recognition and time series forecasting, where rapid inference is critical~\citep{lim2021temporal}. Furthermore, FFT-based parallelization enables S4 to be highly optimized for GPU and TPU architectures, making it practical for large-scale deployments~\citep{raffel2020exploring, shoeybi2019megatron}. This sets S4 apart from traditional RNNs, which suffer from poor hardware utilization due to their inherently sequential nature.

\subsection{Strengths \& Limitations}
While S4 introduces a range of powerful innovations, it also comes with certain challenges and trade-offs. This section discusses both the strengths and limitations of the model.

\textbf{Strengths.} One of the primary advantages of S4 is its ability to process extremely long sequences efficiently, scaling as $O(T\log T)$ at training time against the $O(T^2)$ of self-attention and the sequential $O(T)$ of an RNN~\citep{gu2021efficiently}. Its structured recurrence and FFT acceleration enable it to handle sequences of tens of thousands of tokens, making it ideal for applications like DNA sequence modeling~\citep{smith2023convolutional}, weather forecasting~\citep{lim2021temporal}, and long-context NLP tasks~\citep{zhang2024chain}. Unlike RNNs, which suffer from sequential computation bottlenecks, S4 is highly parallelizable, making it suitable for GPU and TPU acceleration~\citep{huang2019gpipe, shoeybi2019megatron}. This makes S4 a practical choice for large-scale machine learning applications. S4's structured recurrence gives an implicit memory whose decay is controlled by the parameterization rather than learned from scratch, which is what allows it to retain dependencies over horizons where an LSTM's gradient signal has already vanished~\citep{gu2021combining}. This is particularly useful for tasks requiring long-term context understanding, such as speech-to-text processing~\citep{radford2023robust} and financial time series analysis~\citep{tay2020long}.

\textbf{Limitations.} One of the major drawbacks of S4 is its complex initialization process. Training S4 requires specialized parameter tuning to maintain numerical stability, as the structured state-space formulation can be sensitive to improper initialization~\citep{gu2022train}. Spectral normalization techniques have been explored to mitigate this issue, but further research is needed to simplify S4’s optimization dynamics~\citep{gogianu2021spectral}. Unlike attention-based models, where attention weights provide interpretability, S4’s state-space formulation lacks an easily interpretable mechanism for analyzing how it processes sequential information~\citep{zhang2024chain}. This makes it challenging to understand how the model prioritizes different parts of an input sequence, particularly in applications such as medical decision-making. While S4 is designed for efficiency, it requires carefully optimized implementations to utilize its computational benefits fully~\citep{gu2022train}. This can be a limitation in environments where high-performance GPUs/TPUs are unavailable.


\section{Evolution of S4-Based Models}
\subsection{Mamba: Optimized Successor to S4}
The rapid advancement of sequence modeling has led to the widespread adoption of Transformers, yet their quadratic computational complexity poses significant scalability challenges, particularly for long-sequence tasks. SSMs, such as S4, have emerged as promising alternatives by leveraging structured state-space dynamics to enhance efficiency in sequence modeling. However, S4's reliance on static parameters and its computational overhead in handling long-range dependencies limit its scalability. Mamba, a novel Selective State Space Model (SSSM), builds upon S4 by introducing dynamic parameterization and a hardware-aware parallel algorithm, enabling linear-time complexity and substantial improvements in computational efficiency and memory utilization \citep{dao2024transformers}. Unlike S4, which depends on fixed state-space updates, Mamba introduces a data-dependent selection mechanism that dynamically filters inputs, improving long-range dependency modeling \citep{wang2024graph}. It also employs an efficient hardware-aware algorithm that optimizes memory hierarchy and reduces IO operations, achieving linear-time complexity (O(L)), whereas Transformers suffer from quadratic complexity (O(L²)) \citep{wang2024graph}. Additionally, parallel scanning, kernel fusion, and strategic recalculation further enhance computational speed while maintaining accuracy \citep{heidari2024computation}.

\begin{figure}[pos=H]
    \centering
    \includegraphics[width=0.98\linewidth]{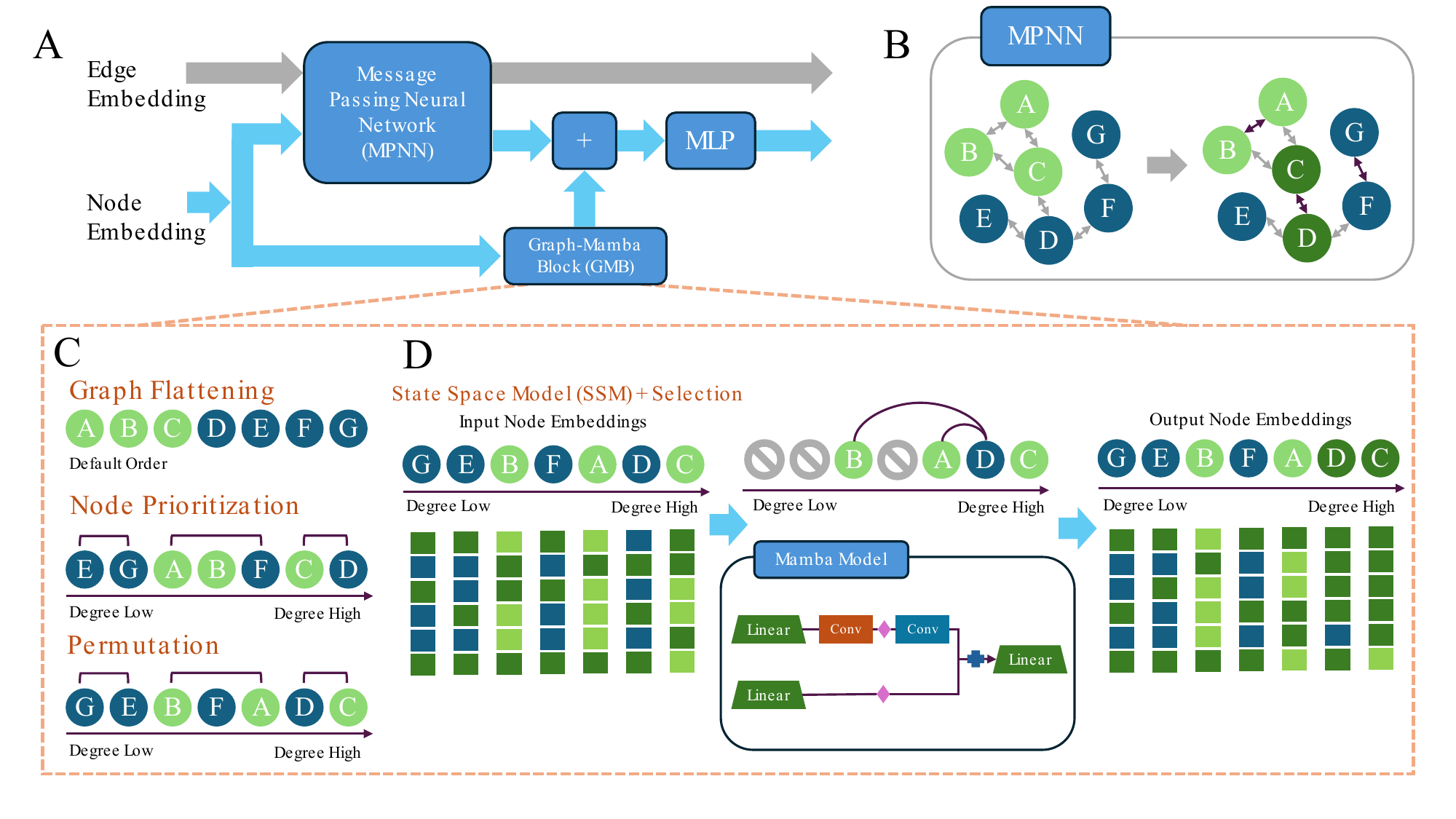}
    \caption{Architecture of Graph-Mamba adapted from \citep{wang2024graph}}
    \label{fig:Graph_mamba_architecture}
\end{figure}

Unlike Transformers, which rely on full self-attention mechanisms, Mamba employs a Gated State Space Model (GSSM) that selectively propagates and forgets information, improving sequence retention and long-range dependency modeling. This selective propagation is further enhanced by input-dependent matrices, which replace costly self-attention mechanisms while retaining the computational efficiency of state-space models \citep{zhang2024surveyvisualmamba}. Additionally, Mamba’s Selective Scan Algorithm optimizes training efficiency by improving GPU memory usage and reducing IO overhead, allowing for faster computation compared to traditional SSMs \citep{zhang2024surveyvisualmamba}. The resultusually quoted for this is a fivefold throughput gain, and it needs its conditions. Gu and Dao report 4 to 5 times the autoregressive inference throughput of a similarly sized Transformer, measured at a large batch size on a single A100 80GB PCIe device with a 2048-token prompt and 128 generated tokens \citep{gu2023mamba}. It is a throughput measurement at a batch size chosen to fill the device, not a single-request latency reduction, and Section~\ref{sec:empirical} reports it alongside the other efficiency results in this literature under the same convention. By eliminating the need for computationally expensive attention operations, Mamba provides a hardware-efficient alternative for sequence modeling, making it highly suitable for long-sequence tasks and high-resolution data processing \citep{zhang2024surveyvisualmamba}. These advantages extend to graph-based modeling, as seen in Graph-Mamba, whose authors report reductions in GPU memory of up to 74 percent relative to the attention module it replaces, on the long-range graph benchmarks they evaluate \citep{wang2024graph}.As with the other reductions quoted in this review, this result comes from the originating authors and is specific to that baseline and those datasets. Specifically, Graph-Mamba replaces the attention module in the GraphGPS framework with the Graph-Mamba Block (GMB), which integrates an edge-based Message Passing Neural Network (MPNN) and a node-centric Mamba model for efficient graph learning. By incorporating node prioritization, permutation-based sparsification, and input-dependent context filtering, Graph-Mamba enhances computational scalability and long-range dependency modeling (as illustrated in Figure. \ref{fig:Graph_mamba_architecture}). 

\begin{figure}[pos=H]
    \centering
    \includegraphics[width=0.98\linewidth]{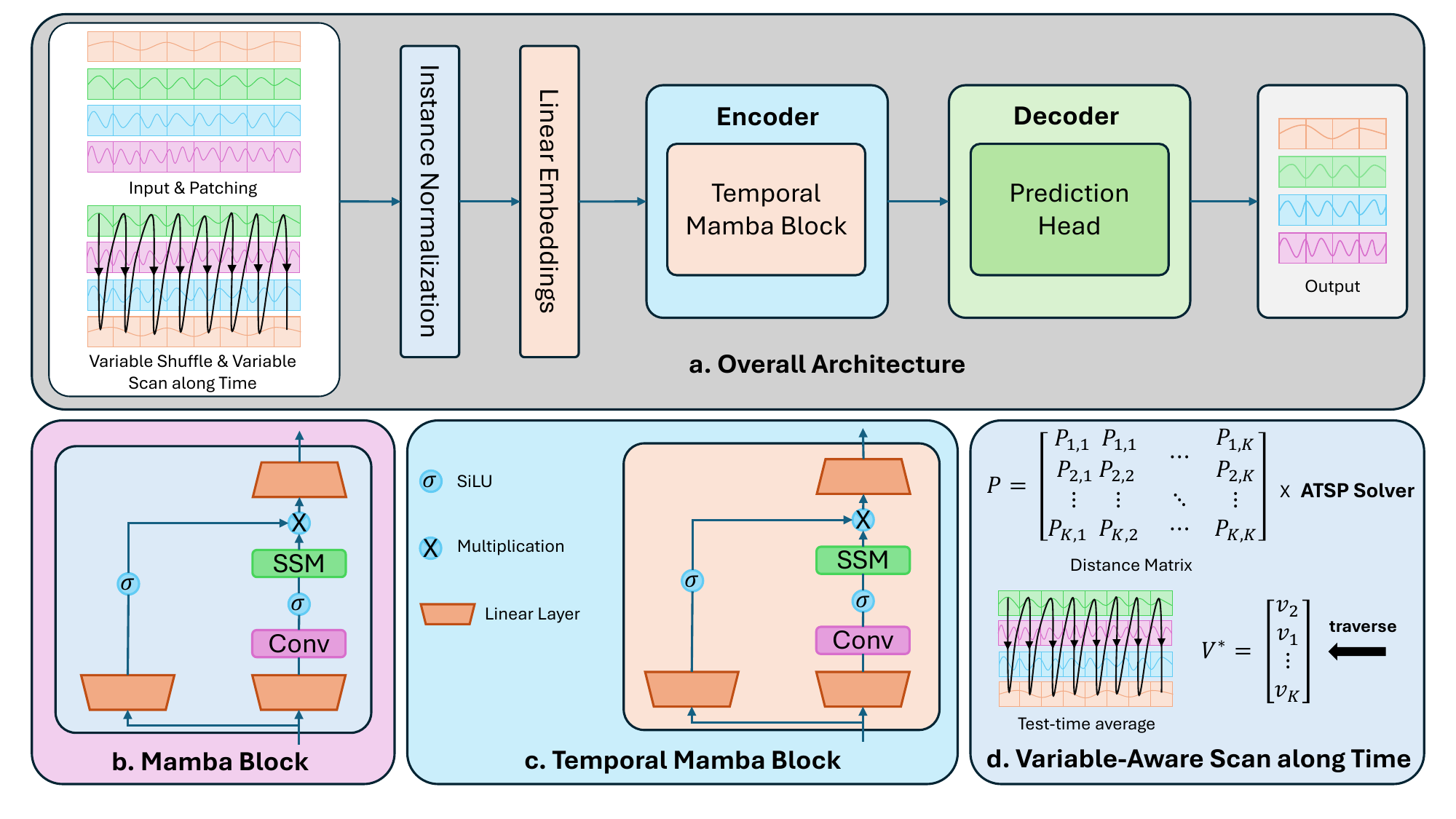}
    \caption{Architecture of MambaTS adapted from \citep{cai2024mambats}}
    \label{fig:MambaTS_architecture}
\end{figure} 

Additionally, in multivariate time-series forecasting, selective state-space models have been reported to improve on the Transformer forecasters they were compared against by mitigating permutation-invariant biases and improving variable selection through an optimized scan mechanism \citep{cai2024mambats}. The comparison is the originating authors' own, on the forecasting datasets they selected, and it is not evidence of a general advantage over attention in this domain. This advantage is exemplified in MambaTS, a specialized adaptation designed for time-series modeling. As illustrated in Figure \ref{fig:MambaTS_architecture}, MambaTS consists of an embedding layer, instance normalization, multiple stacked Temporal Mamba Blocks (TMBs), and a prediction head. It employs patching and tokenization to reduce redundancy, while Variable Scan along Time (VST) arranges tokens in a structured manner to improve long-range dependency modeling. The encoder integrates SSM-based sequence modeling and gated non-linearity, enhancing efficiency and accuracy in capturing temporal patterns. Finally, a channel-independent linear decoder generates predictions, ensuring scalability and computational efficiency.

\subsection{Multi-Input, Multi-Output SSM}
SSMs have emerged as a powerful alternative to Transformers for sequence modeling, offering linear computational complexity and improved efficiency in handling long-range dependencies. Among these, the S5 introduces key architectural advancements that enable MIMO processing, distinguishing it from its predecessor, S4 \citep{smith2022simplified}. Unlike S4, which relies on a bank of independent single-input, single-output (SISO) SSMs, S5 integrates a single MIMO SSM, streamlining computation and enhancing information exchange across multiple channels \citep{smith2022simplified}. Additionally, S5 replaces S4’s convolutional and frequency-domain approach with a parallel scan operation, optimizing its efficiency for multi-modal data processing. This parallel scan mechanism eliminates the need for computationally expensive FFT operations, enabling S5 to scale efficiently while maintaining state transitions across multiple input-output channels \citep{smith2022simplified}.

\begin{figure}[pos=H]
    \centering
    \includegraphics[width=0.98\linewidth]{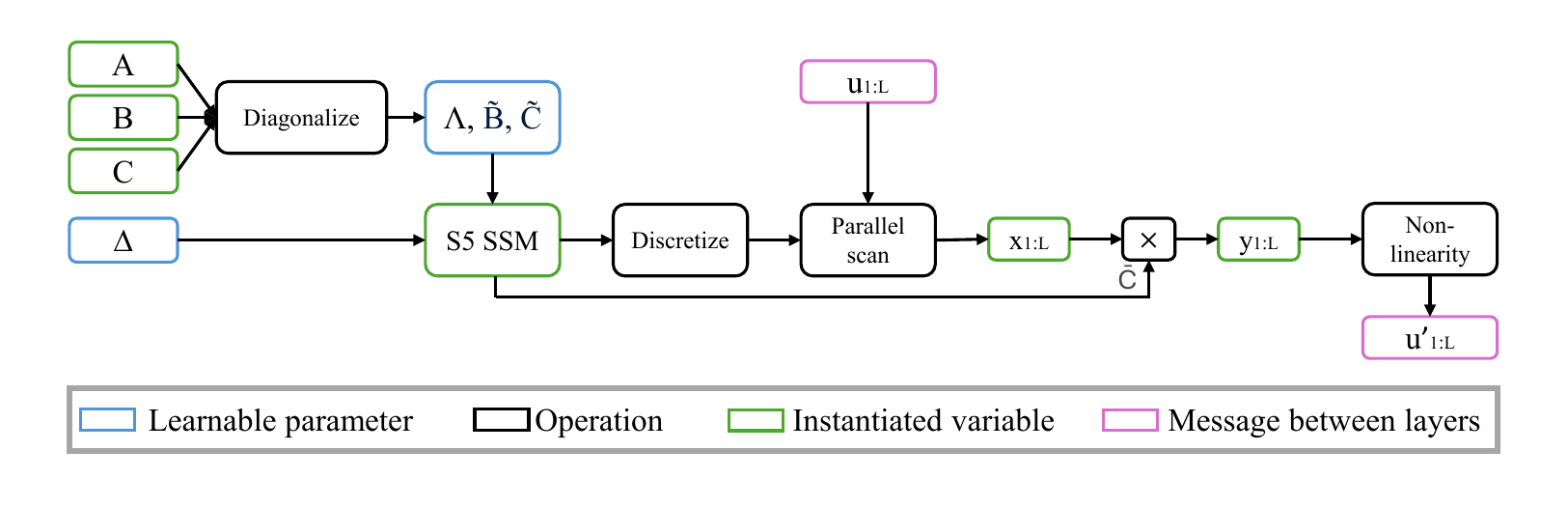}
    \caption{Computational Components of the S5 Layer With Parallel Scan on a Diagonalized Linear SSM for Sequence Modeling, Followed by a Nonlinear Activation adapted from \citep{smith2022simplified}}
    \label{fig:S5_architecture}
\end{figure}

The computational structure of S5 is illustrated in Figure \ref{fig:S5_architecture}. As shown, S5 enhances efficiency by utilizing a parallel scan operation on a diagonalized linear SSM, allowing it to compute outputs recurrently in the time domain rather than relying on frequency-domain convolutions \citep{smith2022simplified}. The transition from independent SISO state models in S4 to a unified MIMO framework in S5 enables more efficient computation of long sequences while preserving computational complexity. Additionally, S5 leverages HiPPO initialization schemes, a key feature inherited from S4, to optimize its internal state representations for improved sequence modeling. After computing the SSM outputs, a nonlinear activation function is applied to produce the final layer outputs, ensuring efficient transformation of sequential dependencies.

The Mamba architecture further refines these structured state space models by introducing Selective State Space Mechanisms (S6), which dynamically adjust hidden states based on input context, improving scalability and enabling efficient parallelization \citep{dao2024transformers}. Mamba also depends on a hardware-aware implementation: its fused selective-scan kernel is reported to be 20 to 40 times faster than a naive scan written in PyTorch, measured on a single A100 80GB PCIe at model dimension 1024, state size 16 and batch size 1 in bfloat16 \citep{gu2023mamba}. The comparison is against an unoptimized reference implementation rather than against a Transformer, and it is a kernel measurement rather than an end-to-end one. Compared to S5, Mamba’s associative parallel scans extend beyond structured sequence modeling, improving scalability in long-sequence tasks \citep{smith2022simplified}. This makes Mamba particularly effective for multi-modal processing, as seen in architectures like VL-Mamba and Coupled Mamba, which introduce state-coupling mechanisms to enhance cross-modal information fusion \citep{li2024coupled, qiao2024vl}.

Coupled Mamba lets the state of one modality's recurrence influence another's while keeping the per-modality dynamics separate, which preserves the parallel evaluation that a fully coupled formulation would lose \citep{li2024coupled}. KalMamba further extends Mamba by integrating probabilistic inference with Kalman filtering, allowing efficient time-parallel belief state estimation for reinforcement learning applications, although it is not explicitly focused on multi-modal fusion \citep{becker2024kalmamba}. Furthermore, SiMBA, a Mamba-based variant, introduces Einstein FFT (EinFFT) for spectral channel mixing, specifically optimizing Mamba for vision and multivariate time-series tasks, thereby addressing scalability concerns in high-dimensional data processing \citep{patro2024simba}.

\subsection{Mamba-2 and structured state-space duality}
\label{sec:ssd}

Selectivity solved an expressivity problem and created a hardware problem. Once
$\mathbf{A}$, $\mathbf{B}$ and $\mathbf{C}$ depend on the input, the recurrence is no
longer time-invariant, the global convolution view is unavailable, and training must
fall back on a scan. Mamba's scan is a carefully engineered fused kernel, but a scan
is an associative reduction, and associative reductions do not use the
matrix-multiplication units that dominate the arithmetic throughput of a modern
accelerator. Structured state-space duality (SSD) is the result that recovers those
units without giving up selectivity \citep{dao2024transformers}.

\subsubsection{The sequence transformation as a structured matrix}

Any linear sequence transformation can be written as multiplication by a single
matrix $\mathbf{M}$, with $y = \mathbf{M}x$. For a causal state-space model with state
size $N$, that matrix is not arbitrary. Writing out the recurrence and collecting
terms gives
\begin{equation}
\label{eq:ssm_matrix}
\mathbf{M}_{ji} \;=\; \mathbf{C}_j^{\top}\,\mathbf{A}_{j}\mathbf{A}_{j-1}\cdots\mathbf{A}_{i+1}\,\mathbf{B}_i ,
\qquad j \geq i ,
\end{equation}
and zero above the diagonal. A lower-triangular matrix of this form is
\emph{$N$-semiseparable}: every submatrix drawn entirely from the region on or below
the diagonal has rank at most $N$, the state size. The rank bound is not incidental,
it is the matrix-level statement of what a state-space model does, namely route all
information from the past to the future through a state of fixed width. Dao and Gu
prove the converse as well, so that structured state-space models and semiseparable
matrices in sequentially semiseparable representation are the same object viewed two
ways.

The duality follows immediately. Computing $y = \mathbf{M}x$ by materializing
$\mathbf{M}$ and multiplying costs $O(T^2)$, and this is the quadratic,
attention-like mode. Computing it by exploiting the rank structure, that is by
carrying the state forward, costs $O(TN^2)$ and is the linear, recurrent mode. The
two modes are not two models. They are two orderings of the same contraction, and
which one is preferable is a question about sequence length, state size and hardware
rather than about modeling.

\subsubsection{Attention as a special case}

The same argument runs in the other direction. Masked kernel attention computes
$\mathbf{Y} = (\mathbf{L} \circ \mathbf{Q}\mathbf{K}^{\top})\mathbf{V}$, where
$\mathbf{L}$ is the causal mask. Ordinary causal linear attention takes $\mathbf{L}$
to be the lower-triangular matrix of ones, and the reason linear attention is fast is
that multiplying by that particular mask is a cumulative sum. Dao and Gu observe that
nothing requires the mask to be all ones: any $\mathbf{L}$ admitting subquadratic
multiplication preserves the efficient mode. Replacing the mask with an arbitrary
structured matrix gives \emph{structured masked attention}, of which causal linear
attention is the all-ones case and the decay mask of retentive networks is another.

Setting $\mathbf{A}_t = a_t \mathbf{I}$, a scalar times the identity, makes
Equation~\eqref{eq:ssm_matrix} factor as $\mathbf{M} = \mathbf{L} \circ
(\mathbf{C}\mathbf{B}^{\top})$ with $\mathbf{L}_{ji} = a_j a_{j-1} \cdots a_{i+1}$, a
$1$-semiseparable matrix. Evaluating that product naively is exactly quadratic masked
kernel attention. This is the precise sense in which a selective state-space model
with scalar state transitions \emph{is} a form of attention, and conversely
$1$-semiseparable structured masked attention is a special case of a diagonal
state-space model.

\subsubsection{What changes between Mamba and Mamba-2}

Mamba-2 buys this correspondence with a deliberate restriction, and
Table~\ref{tab:mamba1_vs_mamba2} sets out what is traded. Mamba's $\mathbf{A}_t$ is
diagonal, so each channel evolves with its own decay rate. Mamba-2 collapses the
diagonal to a single scalar per head per step. Expressivity falls, and the authors
say so plainly. What is bought is that the block decomposition of the semiseparable
matrix becomes dominated by dense matrix multiplications on $N \times N$ blocks,
which run on the accelerator's matrix units. The asymptotic training cost is
unchanged at $O(TN^2)$; what changes is the fraction of that cost the hardware
executes at peak rate.

The restriction also makes a larger state affordable. Mamba's released models use
$N = 16$; Mamba-2 raises the state by roughly a factor of eight or more, with
$N = 64$, $128$ and $256$ all appearing across its experiments. Since the recurrent
state is the model's entire memory of the past, this is a direct increase in capacity
rather than a bookkeeping change, and it is visible on associative-recall tasks, where
Dao and Gu report that moving from $N = 16$ to $N = 64$ and then $N = 256$ improves
performance consistently.

\begin{table}
\centering
\caption{Differences between Mamba and Mamba-2.}
\label{tab:mamba1_vs_mamba2}
\footnotesize
\setlength{\tabcolsep}{4pt}
\renewcommand{\arraystretch}{1.2}
\begin{tabular}{@{}p{0.24\linewidth} p{0.33\linewidth} p{0.33\linewidth}@{}}
\toprule
 & \textbf{Mamba (S6)} & \textbf{Mamba-2 (SSD)} \\
\midrule
State transition $\mathbf{A}_t$ &
Diagonal, one decay rate per channel &
Scalar times identity, one decay per head per step \\
\addlinespace
State size $N$ &
$16$ in the released models &
Raised roughly eightfold or more; $64$, $128$ and $256$ used across experiments \\
\addlinespace
Head dimension $P$ &
$P = 1$; every channel has independent dynamics &
$P > 1$, typically $64$ or $128$, analogous to multi-head attention \\
\addlinespace
Training algorithm &
Fused associative selective scan in custom CUDA &
Block decomposition of the semiseparable matrix, dominated by dense matrix products \\
\addlinespace
Hardware units used &
Does not use matrix-multiplication units &
Runs the dominant work on matrix-multiplication units \\
\addlinespace
Training cost &
$O(TN^2)$ &
$O(TN^2)$, unchanged asymptotically \\
\addlinespace
Reported kernel speed &
Baseline &
$2$ to $8\times$ faster than the Mamba scan\evorig, core-layer microbenchmark, one A100 80GB PCIe, $N = 64$, sequence lengths $512$ to $512$K; batch size \nr \\
\addlinespace
Expressivity &
Higher, per-channel dynamics &
Deliberately reduced, traded for hardware utilization \\
\bottomrule
\end{tabular}
\tabnote{Evidence markers: \keyorig; \keynr.}
\end{table}

The efficiency claim attached to this reformulation needs its conditions stated,
because it is frequently repeated without them. A dedicated SSD kernel is reported to
be 2 to 8 times faster than Mamba's fused selective scan. That measurement is a
core-layer microbenchmark on a single A100 80GB PCIe device at state size $64$, swept
over sequence lengths from $512$ to $512$K, and the range is the spread across that
sweep rather than a single operating point. It is not an end-to-end model speedup, the
paper does not report the batch size used, and it should not be merged with the
separate comparison against FlashAttention-2 in the same figure. Dao and Gu also note
the counter-case: at sequence lengths around $2$K a complete Mamba-2 model can still
train more slowly than a parameter-matched Transformer, because half of the
Transformer's layers are multilayer perceptrons, which are close to ideal for matrix
units.

\subsubsection{Why this matters for hybrids}

The duality is the reason the hybrid designs of Section~\ref{sec:hybrids} are a
principled construction rather than an empirical convenience. Two consequences follow. The systems consequence is that attention and state-space layers stop being separate
engineering problems. Once the state-space layer is expressed as structured masked
attention, the tensor and sequence parallelism strategies, the kernel fusion patterns
and the memory layouts developed for attention transfer to it, which lowers the cost
of putting both in one model.

The modeling consequence is sharper. The rank bound that defines
$N$-semiseparability is also a capacity bound: everything the model can carry forward
from position $i$ to position $j$ passes through $N$ numbers. Attention has no such
bound, since it retains every key and value, which is why its memory grows with
context and why it does not fail at exact retrieval. Stating it this way makes the
division of labor in a hybrid explicit. State-space layers supply linear-time
throughput and a constant-size decoding state; a small number of attention layers
supply the unbounded, exact lookup that a fixed-width state provably cannot. This is
also why the empirically preferred attention fraction is small rather than zero or
large: Dao and Gu find roughly ten percent of layers to be the best setting in their
own ablation, where a $350$M-parameter model on the Pile reaches perplexity $8.26$
with six attention blocks against $8.60$ for pure Mamba-2 and $8.68$ for a strong
Transformer recipe.


\section{SSM-attention hybrids}
\label{sec:hybrids}

Pure state-space models and pure Transformers exhibit complementary limitations, as clarified by the duality discussed in Section~\ref{sec:ssd}. A state-space layer summarizes the input history within a fixed-dimensional state of width $N$. This enables linear-time sequence processing and keeps the decoding state independent of context length, but it also limits the amount of information that can be retrieved exactly. In contrast, attention preserves all keys and values, supporting direct retrieval from the full context at the cost of a cache that grows linearly with sequence length. Hybrid architectures combine these strengths by inserting a small number of attention layers into a state-space backbone, thereby improving retrieval while retaining much of the computational and memory efficiency of state-space models. The architectures discussed below differ primarily in how these two components are integrated, and these design choices lead to distinct tradeoffs in efficiency, memory use, and retrieval capability.

\subsection{How the two mechanisms are combined}

Four arrangements appear in the literature, and they differ in where the attention sits
rather than in what it does.

\paragraph{Interleaving in depth.} Jamba alternates Mamba and attention layers at a
ratio of seven to one, in blocks of eight layers, and inserts mixture-of-experts
layers every second layer with sixteen experts and top-two routing, giving 12B active
of 52B total parameters at a 256K context \citep{lieber2024jamba}. It reports roughly
three times the throughput of Mixtral on a single A100 80GB in int8 at an 8K context
with 512 output tokens. Jamba-1.5 scales the same recipe and reports the deployment-relevant metric: a key-value cache of 9~GB at a 256K context in 16-bit precision, compared with 80~GB for a 70B-parameter Transformer of comparable class \citep{team2024jamba}. Note that Jamba-1.5 reports a latency curve rather than a bare throughput multiplier, so the widely repeated ``three times faster'' claim belongs to Jamba and not to Jamba-1.5.

\paragraph{Sharing one attention block.} Zamba takes the interleaving idea to its
sparsest limit. A single global block, self-attention followed by a multilayer
perceptron, has its weights shared and is reapplied roughly every six Mamba blocks,
with its input concatenated with the original embedding and a separate output
projection per application \citep{glorioso2024zamba}. Attention is thus present at
several depths while being paid for once in parameters. The model is 7B parameters
over 80 layers, trained on 1T tokens.

\paragraph{Sequential stacking with local attention.} Samba stacks Mamba, then
sliding-window attention with a window of 2048, then a multilayer perceptron, and
repeats \citep{ren2025samba}. Because the attention is windowed rather than global,
the per-step cost is bounded regardless of context, which is what allows the 3.8B
model, trained on 3.2T tokens at a 4K context, to extrapolate to 256K. It reports
3.73 times the prompt-processing throughput of Llama-3 at 1.6B scale on a 128K prompt,
single A100 in bfloat16.

\paragraph{Parallel heads within a layer.} Hymba differs from all of the above in that
it does not interleave at all \citep{dong2025hymba}. Attention heads and state-space
heads run \emph{in parallel inside the same layer} on the same input, and their
outputs are normalized and combined with learned scalars. The two mechanisms therefore
see identical inputs rather than each other's outputs, which the authors argue lets
attention supply high-resolution recall while the state-space heads supply a summarized
context. Combined with sliding-window attention, it reports an 11.67-fold cache
reduction and 3.49 times the throughput of Llama-3.2-3B at an 8K sequence length,
batch size 128, on an A100 with a 16-bit cache.

\paragraph{Gated linear recurrence with local attention.} Griffin sits slightly
outside the state-space family proper. It interleaves two blocks of a gated linear
recurrent unit with one block of local multi-query attention at a window of 1024
\citep{de2024griffin}, and RecurrentGemma is the released model built on that recipe,
with a window of 2048 and 2B and 9B variants trained on 2T tokens
\citep{botev2024recurrentgemma}. It is included here because the hybridization pattern
is the same even though the recurrence is not derived from a continuous-time
state-space formulation.

\subsection{Obtaining a hybrid without pretraining}

A hybrid need not be trained from
scratch, which matters more for adoption than any of the architectural choices above.
MOHAWK distils a pretrained Transformer into a Mamba-2-based student in three
stages, aligning the sequence-mixing matrices, then the hidden states, then
transferring the remaining weights with knowledge distillation, using about 3.0B
tokens in total \citep{bick2024transformers}. Its hybrid variant retains four of
twenty-four attention layers and uses 5B tokens. This matters for the practical
argument of Section~\ref{sec:limitations}: the cost of moving to a state-space
backbone need not be a full pretraining run.

\subsection{What the pattern shows}

Two things are consistent across these designs.
The attention fraction is small and deliberately so, from one layer in eight in Jamba
down to a single shared block in Zamba, and the ablation in Mamba-2 puts the optimum
near ten percent of layers. And the reported advantage that survives across papers is
the cache, not the throughput multiplier. Throughput results vary with kernel, precision, batch size, and device, and the values quoted above were each measured under different conditions. The cache reduction follows from the architecture itself, which is why it is the claim this review treats as durable.

\section{S4-family variants}
\label{sec:variants}

The Structured State-Space Sequence model revolutionized long-range sequence
modeling, but its diagonal-plus-low-rank parameterization is intricate, its
initialization is delicate, and its time invariance prevents content-dependent
filtering. The models in this section are the direct responses to those three
difficulties. Each retains a structured state-space recurrence with a learned state
transition, which is the membership rule used throughout this review and stated in the
caption of Table~\ref{tab:model_families}. Two boundaries are worth drawing before the descriptions begin. Models that reach
subquadratic sequence mixing without such a recurrence, including Hyena, RWKV, Mega and
the Toeplitz networks, are treated in Section~\ref{sec:adjacent} and are not S4
descendants, whatever their names suggest. Models that keep the recurrence unchanged and
adapt it to a data modality, such as U-Mamba and FusionMamba, are architectural
applications rather than architectural variants and are treated in
Section~\ref{sec:applications} beside comparable work in their own domains.

\subsection{Multi-Dimensional S4 (S4ND)}
        
S4ND extends S4 to multi-dimensional signals like images and videos by converting state-space equations to partial differential equations (PDEs) \citep{nguyen2022s4nd}. Unlike S4, which operates on one-dimensional sequences, S4ND characterizes multi-dimensional signals in terms of:
        \vspace{-0.5em}
        \begin{equation}
            \frac{\partial x}{\partial t^{(1)}} = A^{(1)} x + B^{(1)} u,
        \end{equation}
        \vspace{-1em}
        \begin{equation}
            \frac{\partial x}{\partial t^{(2)}} = A^{(2)} x + B^{(2)} u.
        \end{equation}
        
Discretization of the above PDEs allows S4ND to construct multi-dimensional convolution kernels that can effectively process high-dimensional data. S4ND can be used to substitute Conv2D and self-attention layers to improve performance on vision tasks. Its authors report a 1.5 percentage point gain over a Vision Transformer baseline on ImageNet-1k and a 4 percentage point gain over a 3D ConvNeXt on HMDB-51 video classification \citep{nguyen2022s4nd}. Both are drop-in substitutions into an otherwise unchanged backbone, reported by the originating authors under their own training recipe, and neither has been reproduced independently.
        
One of its strengths is its resolution-invariant, implicitly learned convolutional kernels that encourage scale generalization. A band-limiting extension also minimizes aliasing, which the authors demonstrate in a deliberately narrow zero-shot setting: trained on 8$\times$8 images and tested at 32$\times$32 without retraining, the model exceeds a two-dimensional convolutional baseline by 40 percentage points \citep{nguyen2022s4nd}. The size of that margin reflects how badly a fixed-size kernel transfers across resolutions rather than a general 40 percent advantage, and it does not carry over to matched-resolution training. While S4ND's multiple-state matrix support increases computational cost, its ability to capture spatial dependency makes it an interesting candidate for medical imaging and video recognition tasks.

\subsection{Diagonal State Spaces (DSS)}

DSS simplifies the S4 model by constraining the state matrix \( A \) to be diagonal, significantly reducing computational load while maintaining most of S4's performance \citep{gupta2022diagonal}. The state-space method is formulated as:
        \vspace{-0.5em}
        \begin{equation}
            x'(t) = A x(t) + B u(t), \quad y(t) = C x(t).
        \end{equation}
        
DSS avoids low-rank corrections in S4, resulting in a more efficient implementation. The convolution kernel is reduced to:
        \vspace{-0.5em}
        \begin{equation}
            K_l = \sum_{n=1}^{N} C_n e^{\lambda_n \Delta l} B_n.
        \end{equation}
        
This design allows DSS to achieve competitive performance with S4 on long-range sequence modeling tasks while being easier to implement and optimize \citep{gupta2022diagonal}. However, the use of diagonal matrices limits interactions between state dimensions, reducing expressiveness in complex sequence tasks. Despite this, DSS performs comparably to S4 on benchmarks such as the LRA and speech classification \citep{gupta2022diagonal}.

\subsection{Diagonal S4 (S4D)}

S4D strengthens DSS by incorporating S4 initialization methods while retaining a diagonal state matrix \citep{gu2022parameterization}. It generalizes DSS's diagonal structure but enhances long-range dependency modeling through specialized initialization techniques. The convolution kernel is given by:
        \vspace{-0.5em}
        \begin{equation}
            K_l = \sum_{n=0}^{N-1} C_n \lambda_n^{l} B_n.
        \end{equation}
        
S4D balances expressiveness and computational cost by maintaining the advantages of DSS while avoiding the computational burden of full-rank state matrices. It is compatible with Vandermonde matrix-based kernel computation, retaining the same theoretical complexity as S4 but with a simpler implementation \citep{gu2022parameterization}.
        
Compared to DSS, S4D supports more flexible initialization schemes, such as HiPPO matrix-based methods, leading to improved performance on long-range sequence modeling tasks. Its authors report results on image, audio and time-series benchmarks in which the best S4D initialization exceeds DSS, at lower implementation cost \citep{gu2022parameterization}. The margin depends on which initialization is used, and Table~\ref{tab:empirical} reports the three variants separately for that reason.

\subsection{Liquid S4}
        
Liquid-S4 augments S4 with an input-sensitive state transition scheme, drawing inspiration from Liquid Time-Constant (LTC) networks \citep{hasani2022liquid}. The state-space representation is modified as:
        \vspace{-0.5em}
        \begin{equation}
            x'(t) = (A + B \odot f(x,u)) x + B \odot f(x,u).
        \end{equation}
        
Here, \( f(x,u) \) is a dynamically changing nonlinear function with respect to the input, enabling the model to selectively filter or emphasize information. This responsiveness allows Liquid-S4 to better handle variable-length dependencies compared to regular SSMs.
        
Compared with standard S4, which relies on a fixed state transition matrix, Liquid-S4's adaptive mechanism enables generalization across more diverse input sequences. Its authors report improved Long Range Arena results over S4 and stronger speech-command recognition than a convolutional baseline at fewer parameters \citep{hasani2022liquid}. As with the other Long Range Arena results collected in Section~\ref{sec:empirical}, these are the originating authors' own results under their own protocol. However, its non-linearity imposes additional computational costs and may require more regularization to ensure stable training.

\subsection{Gated State Space Models (GSS)} 
GSS models enhance state-space modeling efficiency and scalability by incorporating gating mechanisms to regulate information flow selectively \citep{mehta2022long}. Building upon S4 and DSS, GSS addresses computational inefficiencies by dynamically activating state updates, improving performance in autoregressive sequence modeling.

The GSS architecture consists of a gated state-space module that controls information flow between dense layers and state-space components, allowing efficient long-range dependency modeling \citep{mehta2022long}. By using fast Fourier transform convolutions, GSS keeps the $O(T \log T)$ training cost of the diagonal state-space layer it builds on rather than the quadratic cost of attention. Its authors report training roughly two to three times faster than DSS at comparable accuracy, and zero-shot generalization to sequences of up to 65K tokens; both results are the originating authors' own and were obtained in their language-modeling setting rather than under an independent protocol.

A hybrid variant of GSS integrates self-attention layers, enhancing short-range modeling while maintaining state-space efficiency\citep{mehta2022long}. Despite its advantages, GSS faces trade-offs in expressivity and speed, as it remains less optimized than highly tuned Transformer models \citep{gu2023mamba, mehta2022long}. Ongoing research focuses on refining hybridization strategies and optimizing GPU and TPU implementations for broader applicability.

\subsection{Hungry Hungry Hippos (H3)}

H3 is the model that first closed most of the language-modeling gap between
state-space layers and attention, and it did so before selectivity existed
\citep{fu2022hungry}. Its authors diagnosed the gap concretely rather than
generically: a linear time-invariant state-space layer fails at two synthetic
abilities that language modeling depends on, namely recalling a token seen earlier in
the sequence and comparing tokens across the sequence.

The response is architectural. H3 stacks two state-space layers with a
multiplicative interaction between them. A shift matrix in the first layer gives the
model an explicit short window over recent tokens, while a diagonal state-space layer in
the second carries information over the full sequence, and the multiplicative gate lets
the second layer's output be modulated by the first. The composition is what supplies an
approximate lookup that neither layer provides alone. The authors also contributed
FlashConv, a fused kernel for the state-space convolution, which is a direct ancestor of
the hardware-aware kernels that Section~\ref{sec:ssd} discusses.

H3 belongs in this section rather than among the adjacent models because both of
its mixing layers are structured state-space recurrences. Its principal limitation is
that the lookup it provides is approximate and architecturally fixed: the window is a
design constant rather than a function of the input. Selectivity, one year later,
replaced that fixed construction with an input-dependent one, which is why H3 is best
read as the last and most successful attempt to obtain content-dependence without
giving up time invariance. Considering together, these variants converge on a single conclusion that
Section~\ref{sec:crosscut} develops: once the state matrix is diagonal and the
initialization is derived rather than guessed, most of the original S4 machinery is
unnecessary, and the remaining limitation is time invariance itself.

\section{Adjacent subquadratic sequence models}
\label{sec:adjacent}

The models in this section reach subquadratic sequence mixing without implementing a
structured state-space recurrence with a learned state transition. They are included
because they share the design pressure that produced the state-space family and are
routinely benchmarked against it, but they are not state-space descendants and this
review does not treat them as such. Stating the boundary matters: several published
taxonomies place these models inside the state-space family, which makes the family's
defining property impossible to recover from the taxonomy itself. Applying the rule
consistently also moves two models the other way, since the gated state-space layer and
H3 are state-space recurrences despite being commonly grouped with the models below, and
both are therefore described in Section~\ref{sec:variants}.

\subsection{Linear recurrent models}

\subsubsection{RWKV}
The RWKV model is a hybrid neural architecture that combines the parallelized training
efficiency of Transformers with the sequential inference benefits of RNNs
\citep{peng2023rwkv}. By employing a linear attention mechanism, RWKV efficiently scales
to long sequences while significantly reducing computational complexity. This design
allows it to overcome the quadratic memory and compute requirements of Transformers
while retaining expressiveness \citep{peng2023rwkv, li2024survey}. The hidden state is
updated through
\vspace{-0.3em}
\begin{equation}
    p_t = \omega \odot p_{t-1} + e^{k_t} \odot v_t, \quad q_t = \omega \odot q_{t-1} + e^{k_t},
\end{equation}
\vspace{-1.5em}
\begin{equation}
    h_t = \sigma(r_t) \odot \frac{p_t}{q_t},
\end{equation}

so that sequences are processed with constant memory at inference.

RWKV integrates key architectural elements from both paradigms. It employs a Receptance
Vector (R) to regulate memory retention, a Weight Decay Vector (W) to modulate historical
influence, and Key (K) and Value (V) vectors similar to attention mechanisms in
Transformers. These components are structured within stacked residual blocks, where
time-mixing and channel-mixing sub-blocks facilitate efficient long-range dependency
modeling. The time-mixing mechanism employs a weighted decay function to manage token
interactions, while channel-mixing enhances expressivity across feature dimensions
\citep{peng2023rwkv, li2024survey}.

The model has undergone several refinements, with RWKV-4 introducing time- and
channel-mixing mechanisms, RWKV-5 (Eagle) incorporating multi-head matrix-valued states to
improve expressiveness, and RWKV-6 (Finch) optimizing data-driven interpolation for
enhanced sequence modeling \citep{li2024survey}. These iterations have broadened RWKV's
applicability across NLP, computer vision and audio processing, including high-resolution
medical image segmentation, 3D point cloud perception and automatic speech recognition.

Because the recurrence is linear and its decay is parameterized rather than learned
through a saturating nonlinearity, RWKV is far less exposed to the gradient decay that
limits a standard RNN, and it can be trained in parallel in the manner of a Transformer.
Its authors report perplexity competitive with Transformers of comparable parameter count
on The Pile and WikiText-103, at lower inference cost \citep{peng2023rwkv}. These are the
originating authors' results at the scales they trained, and the comparison has not been
re-run here. The structural cost is the one this review returns to throughout: RWKV
compresses the past into a single fixed hidden state, so exact recall of a specific
earlier token is bounded in the same way it is for a selective state-space model.

\subsubsection{Retentive networks}
The retentive network is the closest of the adjacent models to the state-space
family, and comparing the two makes the boundary drawn in this section concrete
\citep{sun2023retnet}. It replaces attention with a retention operator that admits three
computational forms of the same function: a parallel form used for training, a recurrent
form that decodes in constant time and constant memory, and a chunkwise recurrent form
that processes a long input in blocks, encoding each block in parallel while summarizing
across blocks recurrently. The first two are precisely the dual views that
Section~\ref{sec:foundations} develops for a linear time-invariant state-space layer, and
the third anticipates the block decomposition that structured state-space duality later
formalizes in Section~\ref{sec:ssd}.

The reason it is nonetheless adjacent is that its decay is fixed rather than
learned. Retention applies a scalar decay to a relative position, so the operator is a
particular structured mask rather than a state transition estimated from data. In the
vocabulary of Section~\ref{sec:ssd} this makes it an instance of structured masked
attention with a decay mask, which is one of the cases the duality result covers
explicitly. The practical consequence is the one the whole section turns on: a retentive
network inherits constant-time decoding and the associated cache advantage, so on the
deployment axis of Section~\ref{sec:crosscut} it behaves like a state-space model, but it
has no continuous-time derivation, no learned state matrix and no principled
initialization theory to inherit.

\subsubsection{Linear recurrent units}
The linear recurrent unit was introduced to answer a specific question: how much
of a deep state-space layer's advantage comes from the state-space formulation itself,
and how much from properties that a plain recurrent network could also have
\citep{orvieto2023resurrecting}. The answer was that a standard recurrent network can
be brought close to S4-level performance on the Long Range Arena by three changes,
namely removing the nonlinearity from the recurrence, using a complex diagonal
parameterization with a stable exponential form, and normalizing the hidden state at
initialization. Linearity is what allows the recurrence to be evaluated by a parallel
scan, so the parallel training that a state-space layer enjoys is recovered without
the continuous-time derivation. The construction has since been carried into
applied settings, for instance sequential recommendation, where the long user
histories and the strict latency budget of a serving system both favor a linear
recurrence \citep{yue2024linear}.

This result is more important to the lineage than its citation count suggests,
because it separates two things that are easily conflated. Parallelizable training and
stable long-range gradient flow come from linearity and from a controlled spectrum,
not from the HiPPO theory. What the state-space derivation contributes is a principled
initialization, which is a real contribution but a narrower one than the framing of the
earlier literature implies. The linear recurrent unit is therefore treated here as an
adjacent model, since it has no state-space derivation, while being the clearest
available diagnostic of what that derivation actually buys.

\subsubsection{Hierarchically gated recurrent networks}
The hierarchically gated recurrent network is a gated linear recurrent model,
described by its authors as such rather than as a state-space model
\citep{qin2023hgrn}. It places a forget gate inside a linear recurrence and adds one
structural constraint: the gate has a learnable lower bound that increases
monotonically with depth. Lower layers may therefore forget quickly and model local
structure, while upper layers are forced to retain information over long spans. The
model is evaluated on language modeling, image classification and the Long Range
Arena. HGRN2 subsequently expands the recurrent state while keeping the gating scheme
\citep{qin2024hgrn2}.

It is included in this section, and not among the state-space families, for two
reasons. Its recurrence is not derived from a discretized continuous-time system, and
its decay behavior is imposed by an explicit architectural constraint rather than
arising from the parameterization of a state matrix. It is nonetheless directly
relevant, because the monotonic lower bound is an alternative answer to the question
selectivity also addresses: how a fixed-width recurrent state should decide what to
keep. The state-space family answers it by conditioning the transition on the input;
HGRN answers it by conditioning on depth.

\subsection{Long-convolution and gated-attention models}

\subsubsection{Hyena}
Hyena serves as a replacement for self-attention, utilizing long convolutions coupled
with gating mechanisms to achieve sub-quadratic complexity without sacrificing
expressiveness \citep{poli2023hyena}. The model output is given by
\vspace{-0.5em}
\begin{equation}
    y(t) = g(t) \odot (W *_l x)(t),
\end{equation}

where \( (W *_l x) \) represents a convolutional kernel and \( g(t) \) is a
data-dependent gating function. Unlike self-attention, which performs pairwise
interactions between all tokens, Hyena integrates data-dependent implicit long
convolutions, preserving sequence-wide context while avoiding quadratic cost.
Hyena has demonstrated competitive performance in long-context modeling tasks
such as language modeling on WikiText-103 and The Pile, with its authors reporting
perplexity comparable to Transformers at roughly 20 percent less training compute at the
scales they evaluated \citep{poli2023hyena}. The compute reduction is measured at the
sequence lengths and model sizes in that paper, and the margin narrows as sequence length
falls.

Hyena is placed here rather than among the S4 variants because its kernel is
produced by an implicit parameterization, typically a small feedforward network over
positional encodings, rather than by a state transition applied recurrently. There is no
state to carry at inference, which is why its decoding behavior is that of a long
convolution and not that of a recurrence. Its remaining weakness is the one that
motivated selectivity: it may fall short of attention when token-specific retrieval is
required.

\subsubsection{Mega}
Mega addresses two limitations of Transformers, their weak inductive bias and the
quadratic cost of attention, by introducing moving-average equipped gated attention
\citep{ma2022mega}. It refines its hidden state as
\vspace{-0.5em}
\begin{equation}
    y_t = \sigma(q_t \odot k_t) \odot v_t,
\end{equation}

where \( (q_t, k_t, v_t) \) are learned feature representations and \( \sigma \) is an
element-wise gating function. The key ingredient is a multi-dimensional damped
exponential moving average folded into the attention computation, which smooths the input
with a learnable decay factor and supplies the position-aware local bias that plain
attention lacks. Mega further employs single-head gated attention, reaching expressivity
comparable to multi-head self-attention at lower cost, and the Mega-chunk variant applies
the mechanism chunk-wise to reduce the quadratic term to near-linear scaling.

Mega is evaluated by its authors on the Long Range Arena, WMT16 translation, and
image and speech classification, and reports competitive results across that range at the
scales tested \citep{ma2022mega}, including gains over Transformer and S4 baselines on
WikiText-103. Its breadth of evaluation is unusual for this group of models, which is
worth noting given how narrowly most are assessed. It is adjacent rather than
state-space because the exponential moving average is a fixed, input-independent smoothing
operator attached to an attention block, not a learned state transition carrying the
sequence. Tuning the decay parameters is the practical difficulty, and the gating
mechanism adds hyperparameters that require calibration.

\subsubsection{Toeplitz Neural Networks (TNN)}
The TNN is an efficient sequence modeling architecture that replaces traditional attention
mechanisms with Toeplitz matrices, reducing computational complexity from quadratic to
log-linear \citep{qin2023toeplitz}. Unlike Transformers, which depend on pairwise token
relations and position embeddings, TNN models sequences based solely on relative
positional relations, enabling scalability to sequences up to 14K tokens while maintaining
strong performance. TNN's architecture is built on Gated Toeplitz Units (GTU) and Gated Linear Units (GLU),
with the Toeplitz Neural Operator (TNO) handling token mixing purely based on relative
positional information. The Relative Position Encoder (RPE) enables dynamic position
embeddings, allowing generalization across different sequence lengths without retraining
\citep{qin2023toeplitz}. Additionally, TNN integrates an exponential decay bias, similar
to ALiBi, to enhance extrapolation for long sequences.

The Toeplitz network is evaluated on the Long Range Arena, on autoregressive and
bidirectional language modeling, and on ImageNet-1K, with its authors reporting
perplexity improvements over the Transformer and state-space baselines they compare
against \citep{qin2023toeplitz}. Its mixing depends only on relative position, so it
cannot be content aware, which is the same limitation that time invariance imposes on the
S4 family and which selectivity was introduced to remove.

Table~\ref{tab:model_families} collects the models discussed in
Sections~\ref{sec:variants} and \ref{sec:adjacent} together with the foundational,
selective and hybrid families of the preceding sections, grouped by family and described
along the same fields. The grouping carries the inclusion rule: models appear under the
S4 or selective headings only when they implement a structured state-space recurrence
with a learned state transition, models that reach subquadratic sequence mixing by other
means are listed separately as adjacent, and models that leave the recurrence unchanged
while fitting it to a modality are listed as domain adaptations. Quantitative results are
kept out of that table deliberately and reported in Table~\ref{tab:empirical}, where each
number can be shown with the configuration under which it was measured.

\setlength{\tabcolsep}{3pt}
\renewcommand{\arraystretch}{1.2}
\begin{longtable}{@{}>{\footnotesize\raggedright\arraybackslash}p{0.1135\linewidth} >{\footnotesize\raggedright\arraybackslash}p{0.2101\linewidth} >{\footnotesize\raggedright\arraybackslash}p{0.1135\linewidth} >{\footnotesize\raggedright\arraybackslash}p{0.0852\linewidth} >{\footnotesize\raggedright\arraybackslash}p{0.0545\linewidth} >{\footnotesize\raggedright\arraybackslash}p{0.1135\linewidth} >{\footnotesize\raggedright\arraybackslash}p{0.2328\linewidth}@{}}
\caption{State-space models and adjacent sequence models, grouped by family.}
\label{tab:model_families} \\
\toprule
\textbf{Model} & \textbf{Core mechanism} & \textbf{Training cost} & \textbf{Decoding state} & \textbf{Input-dep.} & \textbf{Typical domain} & \textbf{Principal limitation} \\
\midrule
\endfirsthead
\ltconthead{7}
\toprule
\textbf{Model} & \textbf{Core mechanism} & \textbf{Training cost} & \textbf{Decoding state} & \textbf{Input-dep.} & \textbf{Typical domain} & \textbf{Principal limitation} \\
\midrule
\endhead
\midrule
\ltcontfoot{7}
\endfoot
\bottomrule
\endlastfoot

\multicolumn{7}{@{}l}{\textbf{Foundational theory}}\\
\addlinespace
HiPPO \citep{gu2020hippo, gu2022train} &
Optimal projection of the input history onto an orthogonal polynomial basis; supplies the initialization, not a layer &
\nr &
$O(N)$ &
No &
Theory; online function approximation &
A principle rather than an architecture; the projection is only optimal for the measure it is derived under \\
\addlinespace
LSSL \citep{gu2021combining} &
Unifies the recurrent, convolutional and continuous-time views of a linear state-space layer &
$O(TN^2d)$ &
$O(Nd)$ &
No &
Sequential benchmarks; proof of concept &
Computationally impractical: the state transition is dense, which S4 was written to fix \\
\addlinespace
\midrule

\multicolumn{7}{@{}l}{\textbf{S4 family: linear time-invariant structured state spaces} (Section~\ref{sec:variants})}\\
\addlinespace
S4 \citep{gu2021efficiently} &
Diagonal-plus-low-rank state matrix, evaluated as a global convolution by fast Fourier transform &
$O(Td\log T)$ &
$O(Nd)$ &
No &
Long-range benchmarks; raw audio; long time series &
Intricate parameterization and initialization; time invariance prevents content-dependent filtering \\
\addlinespace
DSS \citep{gupta2022diagonal} &
Drops the low-rank correction; purely diagonal state matrix with a closed-form kernel &
$O(Td\log T)$ &
$O(Nd)$ &
No &
Long-range benchmarks; speech classification &
Sensitive to the parameterization of the diagonal; no coupling between state dimensions \\
\addlinespace
S4D \citep{gu2022parameterization} &
Diagonal state matrix with a principled initialization derived from the HiPPO analysis &
$O(Td\log T)$ &
$O(Nd)$ &
No &
Image, audio and time-series benchmarks &
Remains linear and time-invariant; complex eigenvalues complicate implementation \\
\addlinespace
S5 \citep{smith2022simplified} &
Single multi-input multi-output state, evaluated by an associative parallel scan rather than by convolution &
$O(TNd)$ &
$O(Nd)$ &
No &
Long-range benchmarks &
The HiPPO matrix is not stably diagonalizable, so initialization relies on an approximation \\
\addlinespace
S4ND \citep{nguyen2022s4nd} &
Extends the state-space kernel to multiple spatial dimensions for images and video &
$O(Td\log T)$ per axis &
$O(Nd)$ &
No &
Image classification; video classification &
Multiple state matrices and higher parameter count; little benefit for one-dimensional data \\
\addlinespace
Liquid-S4 \citep{hasani2022liquid} &
State transition modulated by an input-dependent correction term drawn from liquid time-constant networks &
$O(Td\log T)$ &
$O(Nd)$ &
Partly &
Long-range benchmarks; speech commands &
Nonlinear dynamics are harder to train; per-step overhead for the correction \\
\addlinespace
GSS \citep{mehta2022long} &
Gating applied to a diagonal state-space layer, with a fast Fourier transform convolution &
$O(Td\log T)$ &
$O(Nd)$ &
Partly &
Autoregressive language modeling &
Still behind well-tuned Transformers on some tasks \\
\addlinespace
BiGS \citep{wang2023bigs} &
Bidirectional gated state-space layer used as a masked-language-model encoder in place of attention &
$O(Td\log T)$ &
$O(Nd)$ &
Partly &
Masked language modeling; GLUE &
Static layers with no pairwise interaction, so its inductive bias differs from a Transformer encoder \\
\addlinespace
H3 \citep{fu2022hungry} &
Two stacked state-space layers, a shift and a diagonal, combined multiplicatively to approximate lookup &
$O(Td\log T)$ &
$O(Nd)$ &
No &
Language modeling; synthetic recall tasks &
The lookup window is an architectural constant, not a function of the input; a perplexity gap persists at larger scales \\
\addlinespace
\midrule

\multicolumn{7}{@{}l}{\textbf{Selective state spaces}}\\
\addlinespace
Mamba (S6) \citep{gu2023mamba} &
State transition, input and output projections all functions of the input; fused selective scan &
$O(TN^2)$ &
$O(Nd)$ &
\textbf{Yes} &
Language modeling; genomics; audio &
Loses the convolutional view; the scan kernel cannot use matrix units; fixed state bounds exact recall \\
\addlinespace
Mamba-2 (SSD) \citep{dao2024transformers} &
State transition restricted to a scalar times the identity, exposing the semiseparable structure and a matrix-product algorithm &
$O(TN^2)$, on matrix units &
$O(Nd)$ &
\textbf{Yes} &
Language modeling; associative recall &
Expressivity deliberately reduced relative to Mamba; short sequences may still favor a Transformer \\
\addlinespace
\midrule

\multicolumn{7}{@{}l}{\textbf{SSM-attention hybrids} (Section~\ref{sec:hybrids})}\\
\addlinespace
Jamba \citep{lieber2024jamba} &
Mamba and attention layers interleaved seven to one, with mixture-of-experts layers &
Mixed &
Constant plus a small cache &
Yes &
Long-context language modeling &
Two architectures to tune; the attention ratio is an extra hyperparameter \\
\addlinespace
Zamba \citep{glorioso2024zamba} &
A single weight-shared attention and perceptron block reapplied at intervals through a Mamba backbone &
Mixed &
Constant plus a small cache &
Yes &
General language modeling &
Sharing constrains what the attention block can specialize to \\
\addlinespace
Samba \citep{ren2025samba} &
Mamba, sliding-window attention and a perceptron stacked in sequence &
Mixed &
Constant plus a bounded window cache &
Yes &
Long-context language modeling; extrapolation &
Windowed attention gives no exact recall beyond the window \\
\addlinespace
Hymba \citep{dong2025hymba} &
Attention heads and state-space heads run in parallel within one layer and are fused &
Mixed &
Constant plus a bounded window cache &
Yes &
Small-scale and on-device language modeling &
Both mechanisms are paid for at every layer rather than at selected depths \\
\addlinespace
Griffin, Recurrent\-Gemma \citep{de2024griffin, botev2024recurrentgemma} &
Gated linear recurrence interleaved with local multi-query attention &
Mixed &
Constant plus a bounded window cache &
Yes &
General language modeling &
The recurrence is not derived from a continuous-time state space, so the family label is loose \\
\addlinespace
\midrule

\multicolumn{7}{@{}l}{\textbf{Domain adaptations}, the recurrence unchanged and fitted to a modality (Section~\ref{sec:applications})}\\
\addlinespace
U-Mamba \citep{ma2024umamba} &
Mamba blocks inserted into a U-Net encoder alongside convolutional layers &
$O(TN^2)$ per block &
$O(Nd)$ &
Yes &
Biomedical image segmentation &
Evaluated on small datasets against author-selected baselines \\
\addlinespace
Fusion\-Mamba \citep{xie2024fusionmamba} &
Cross-modal fusion module built on a selective state-space block &
$O(TN^2)$ per block &
$O(Nd)$ &
Yes &
Multimodal medical and infrared-visible image fusion &
Added dynamic convolution raises cost; no controlled comparison across fusion methods \\
\addlinespace
MambaTS \citep{cai2024mambats} &
Selective scan over patched multivariate series, with a variable-ordering correction &
$O(TN^2)$ &
$O(Nd)$ &
Yes &
Multivariate long-horizon forecasting &
Variable order is arbitrary, so the scan imposes structure the data does not have \\
\addlinespace
Graph-Mamba \citep{wang2024graph} &
Mamba block substituted for attention inside a graph-Transformer framework &
$O(TN^2)$ &
$O(Nd)$ &
Yes &
Long-range graph benchmarks &
Requires a node ordering; results are reported against one attention baseline \\
\addlinespace
VL-Mamba, Coupled Mamba \citep{qiao2024vl, li2024coupled} &
State-space backbone for vision-language input; coupled per-modality states &
$O(TN^2)$ &
$O(Nd)$ &
Yes &
Vision-language and multimodal fusion &
Cross-modal alignment is a retrieval operation, which a fixed-width state bounds \\
\addlinespace
MambaTab \citep{ahamed2024mambatab} &
Mamba block applied to the feature vector of a tabular record &
$O(TN^2)$, $T$ the feature count &
$O(Nd)$ &
Yes &
Tabular classification &
No sequential structure in the data to exploit; reported gains rest on parameter efficiency \\
\addlinespace
\midrule

\multicolumn{7}{@{}l}{\textbf{Adjacent subquadratic models}, included for context and \emph{not} state-space descendants (Section~\ref{sec:adjacent})}\\
\addlinespace
RWKV \citep{peng2023rwkv} &
Linear attention recast as a recurrence with learned exponential decay; trains in parallel, runs as an RNN &
$O(Td^2)$ &
$O(d)$ &
Yes &
Language modeling; several applied modalities &
A fixed-width state is an information bottleneck; the advantage disappears at short context \\
\addlinespace
RetNet \citep{sun2023retnet} &
Retention operator with a fixed scalar decay; parallel, recurrent and chunkwise recurrent forms of one function &
$O(T^2d)$ parallel form &
$O(d)$ &
No &
Language modeling at scale &
Decay is fixed by relative position rather than learned, so the operator cannot condition on content \\
\addlinespace
LRU \citep{orvieto2023resurrecting} &
Plain recurrent network made linear, complex-diagonal and normalized at initialization &
$O(TNd)$ via scan &
$O(Nd)$ &
No &
Long-range benchmarks; recommendation &
No continuous-time derivation, so initialization is empirical rather than derived \\
\addlinespace
HGRN \citep{qin2023hgrn, qin2024hgrn2} &
Gated linear recurrence whose forget-gate lower bound increases monotonically with depth &
$O(Td^2)$ &
$O(d)$ &
Yes &
Language modeling; image classification &
Decay is imposed by depth rather than by content; not derived from a state-space system \\
\addlinespace
Hyena \citep{poli2023hyena} &
Implicitly parameterized long convolutions interleaved with elementwise gating &
$O(Td\log T)$ &
Convolution history &
Partly &
Language modeling; genomics &
Many hyperparameters; no recurrent state, so no constant-cost decoding \\
\addlinespace
Mega \citep{ma2022mega} &
Single-head gated attention with an exponential moving average positional bias; chunked for linear cost &
$O(Tcd)$, chunk $c$ &
$O(cd)$ &
Yes &
Long-range benchmarks; translation; speech &
Chunking makes long-range interaction indirect; extra hyperparameters \\
\addlinespace
TNN \citep{qin2023toeplitz} &
Token mixing by Toeplitz matrices built from relative position alone &
$O(Td\log T)$ &
\nr &
No &
Language modeling; image classification &
Mixing depends only on relative position, so it cannot be content aware \\

\end{longtable}
\tabnote{Membership rule: a model is placed in the S4 or selective families only if it implements a
structured state-space recurrence with a learned state transition; models reaching subquadratic mixing by
other means are grouped as adjacent, and models that leave the recurrence unchanged and fit it to a
modality are grouped as domain adaptations. Costs are per layer, with $T$ the sequence length, $N$ the
state size and $d$ the model width. Evidence marker: \keynr.}

\section{Cross-cutting analysis: what actually matters in state-space design}
\label{sec:crosscut}

The preceding sections trace the development of the state-space lineage. This section examines the major design decisions across that lineage and evaluates their implications for model capacity, computational efficiency, long-context behavior and deployment. Each subsection synthesizes the available evidence around a related set of design dimensions and distinguishes established findings from questions that remain open.

\subsection{Selectivity and computational execution}

The transition from S4 to Mamba represents the principal qualitative change in the lineage. Earlier developments, including diagonal parameterizations and parallel scan implementations, retain a linear time-invariant recurrence in which the same operator is applied at every sequence position. By making $\mathbf{A}$, $\mathbf{B}$ and $\mathbf{C}$ functions of the input, selective state-space models replace a fixed filter with an input-dependent mechanism that determines which information enters, remains in or is removed from the recurrent state.

Selectivity addresses a specific limitation of time-invariant models rather than providing a universal advantage. Its benefits are most evident in tasks requiring token-level filtering, copying or content-addressed lookup, because a fixed operator cannot condition its behavior on the content being processed. The benefit is less pronounced for smooth long-range signals, where S4-family models already perform strongly and input-dependent dynamics may provide limited additional value. This distinction helps explain why S4-family architectures remain competitive on the Long Range Arena but perform less effectively than selective models in language modeling.

The computational consequence follows directly from the loss of time invariance. A time-invariant recurrence admits a global convolutional representation, whereas input-dependent parameters remove that representation and require sequential dependence to be resolved through a scan. Section~\ref{sec:ssd} explains how structured state-space duality restores efficient matrix-based computation under this constraint.

This distinction also determines the appropriate execution mode. A linear time-invariant state-space layer can be evaluated through convolution, recurrence or scan, but these formulations serve different computational objectives. The convolutional formulation is well suited to training on fixed-length sequences because the fast Fourier transform provides $O(T \log T)$ complexity with parallel computation across positions. The recurrent formulation is preferable for autoregressive inference because each token is generated with constant computation and constant memory relative to context length. The scan formulation preserves parallelism when time invariance is unavailable, which explains its adoption in S5 and its necessity in selective architectures. Accordingly, training and inference efficiency should not be represented by a single performance measure. The same model may use different execution algorithms across stages, and training throughput, decoding throughput and memory consumption therefore characterize distinct computational properties.

\subsection{Diagonalization and state-space expressivity}

S4 introduced a diagonal-plus-low-rank parameterization because the HiPPO matrix is not stably diagonalizable, making the low-rank correction necessary to retain the original initialization structure. DSS subsequently demonstrated that a purely diagonal state matrix could achieve comparable performance, while S4D provided a theoretical basis for constructing diagonal initializations from the same underlying formulation. As shown in Table~\ref{tab:empirical}, diagonal variants achieve performance close to that of the corrected model on the Long Range Arena while requiring substantially less implementation complexity.

The subsequent development of the field indicates that diagonalization has become the prevailing design choice. Later architectures use diagonal transitions or impose still stronger structural restrictions, while the diagonal-plus-low-rank formulation remains primarily relevant for understanding the historical and theoretical development of S4. The available evidence therefore supports treating diagonalization as the practical resolution of this parameterization question rather than presenting the two forms as equally active alternatives. Expressivity in state-space models also requires a different interpretation from that commonly applied to Transformers or convolutional networks. In those architectures, capacity is generally associated with width, depth and parameter count. In a state-space layer, however, the recurrent state provides the only channel through which information from an earlier position can influence a later one. State width therefore imposes a central constraint on the information that can be retained and transmitted across the sequence.

This perspective clarifies the role of recurrent nonlinearity. The linear recurrent unit discussed in Section~\ref{sec:adjacent} shows that removing nonlinearity from the recurrence does not necessarily produce a corresponding loss in long-range performance when the model uses a stable complex-diagonal parameterization, appropriate initialization and nonlinear transformations between layers. Recurrent nonlinearity may therefore contribute less to effective capacity than expected while limiting parallel execution.

The same perspective explains the design trade-off introduced by Mamba-2. Restricting the state transition from a diagonal matrix to a scalar times the identity reduces the range of independently representable dynamics. However, the resulting computational structure permits a state approximately eight times larger at comparable cost. Because state width directly determines the amount of information that can be propagated through the recurrence, the expanded state can compensate for the reduced flexibility of the transition operator. The relevant trade-off is therefore between per-channel dynamical flexibility and the width of the recurrent state, rather than between parameter counts alone.

\subsection{Decoding efficiency and deployment constraints}

The principal deployment advantage of state-space models is more clearly expressed through decoding-state behavior than through training complexity alone. Although the quadratic training cost of attention is consequential, training is highly parallel and performed over a bounded number of optimization steps. During autoregressive deployment, by contrast, a Transformer's key-value cache grows with context length. At sufficiently long contexts, memory consumption can become a more restrictive constraint than arithmetic cost because exact attention requires retaining the keys and values associated with previous positions.

A state-space model instead maintains a recurrent state of fixed width. Consequently, both the per-token decoding cost and the memory required for that state remain constant with respect to context length. The reported results illustrate the practical importance of this property: Jamba-1.5 reports 9~GB of cache at a 256K context against 80~GB for a comparable Transformer, and Hymba reports an elevenfold cache reduction against a 3B Transformer. These reductions arise from the architecture's fixed-state recurrence and are therefore more transferable across hardware configurations than kernel-specific throughput measurements.

For this reason, constant-state decoding provides the most stable basis for evaluating the deployment efficiency of this model family. Reported speedups remain informative, but their interpretation depends on the baseline architecture, hardware, sequence length, software implementation and measurement scope. Table~\ref{tab:empirical} therefore reports efficiency results together with the conditions under which they were measured. A throughput multiplier without its baseline, device and evaluation scope does not provide a sufficient basis for comparison across studies.

Deployment analysis must nevertheless account for both architectural and ecosystem constraints. Attention currently benefits from mature kernels, established serving frameworks, quantization support and widely used parameter-efficient fine-tuning methods. These advantages can change as software and hardware support evolves. In contrast, the fixed-size decoding state of a state-space model is an intrinsic architectural property. Evaluations based only on current tooling may therefore understate the long-term deployment potential of state-space architectures, whereas evaluations based only on asymptotic complexity may overstate their present operational readiness. Section~\ref{sec:limitations} separates these considerations and examines the extent to which current adoption remains constrained by the surrounding ecosystem.

\subsection{Long-context extrapolation and retrieval limits}

State-space models can extend to sequence lengths beyond those observed during training without encountering the fixed positional range associated with learned positional embeddings. Because the recurrence is not indexed by absolute position, increasing the sequence length does not inherently move the model beyond a learned positional domain. Samba's ability to train at 4K and extrapolate to 256K at constant throughput provides a representative example of this behavior.

Long-context extrapolation, however, should not be interpreted as equivalent to exact long-range recall. The rank bound derived in Section~\ref{sec:ssd} establishes that information transmitted from position $i$ to position $j$ must pass through a state of dimension $N$. A fixed-width state therefore cannot support unbounded exact retrieval, regardless of training scale. Increasing $N$ from 16 to 256, as in Mamba-2, increases the amount of information that can be retained but does not remove the underlying constraint. The reported monotonic improvement in associative recall with increasing $N$ is consistent with this interpretation. The same limitation has implications for multimodal modeling. State-space backbones are computationally attractive for multimodal inputs because concatenated or aligned streams can produce long sequences for which linear scaling is beneficial. The architectures reviewed in Section~\ref{sec:applications} differ primarily in the degree of interaction permitted among modalities, ranging from a shared backbone over concatenated tokens to coupled modality-specific recurrences in which the state of one modality influences another while preserving separable dynamics.

The expected performance of these architectures depends on the form of cross-modal interaction required by the task. Summarizing multiple streams into a shared contextual representation is compatible with a bounded recurrent state. In contrast, aligning a specific element in one modality with a specific element in another requires a form of content-addressed retrieval. Section~\ref{sec:ssd} indicates that such retrieval remains constrained by the rank of the state-space operator. This analysis therefore suggests that state-space fusion may be well suited to joint contextual summarization but less effective for tasks requiring precise cross-modal correspondence. Existing multimodal studies generally report aggregate task performance without separating these mechanisms, leaving this distinction as a testable hypothesis.

\subsection{The role of attention in hybrid architectures}

State-space models and attention provide complementary capabilities rather than interchangeable solutions. Attention remains advantageous when a task depends on exact retrieval from long contexts, copying specific spans or performing in-context learning over discrete tokens, because these operations require direct access to individual earlier positions. Attention may also remain preferable at short context lengths, where its quadratic cost is limited and the mature implementation ecosystem discussed in Section~\ref{sec:limitations} provides a practical advantage. State-space models are better aligned with settings in which context is long, decoding memory is constrained, inputs are continuous or smoothly varying, or deployment must operate within fixed memory limits. Section~\ref{sec:limitations} translates these distinctions into more explicit architectural guidance.

The emergence of hybrid architectures reflects this division of capabilities. Across independently developed systems, a relatively small proportion of attention layers within a state-space backbone often performs better than either a purely recurrent or purely attention-based design. Jamba uses one attention layer in eight, Zamba shares a single attention block, Samba and Hymba use bounded attention windows, and the Mamba-2 ablation places the strongest performance near ten percent of layers.

Structured state-space duality provides a theoretical explanation for this convergence. Attention and state-space layers can be represented within the same broader operator class, but state-space propagation imposes a rank constraint that limits exact retrieval. Attention layers supply this missing capability and need only appear frequently enough to support the retrieval operations required by the task. The consistent preference for a small but nonzero proportion of attention therefore suggests that future progress is more likely to depend on the placement, frequency and sparsity of attention within recurrent backbones than on its complete removal.

\section{Empirical comparison and what it does and does not establish}
\label{sec:empirical}

A survey of this area is expected to report numbers on shared benchmarks. This section
provides them, and states plainly where the shared benchmark is less shared than it
appears. Table~\ref{tab:empirical} reports Long Range Arena results,
language-modeling perplexity and throughput for the core lineage. Three conventions
govern it. Every cell names the paper it was taken from, because these rows were not
produced under one protocol. Every cell carries an evidence tag distinguishing a result reported by the model's own authors from one copied from earlier work. A cell
that no primary source reports is marked \nr\ rather than filled from a secondary
source.

\begingroup
\footnotesize
\setlength{\tabcolsep}{4pt}
\renewcommand{\arraystretch}{1.2}
\begin{longtable}{@{}>{\footnotesize\raggedright\arraybackslash}p{0.1639\linewidth} >{\footnotesize\raggedright\arraybackslash}p{0.1421\linewidth} >{\footnotesize\raggedright\arraybackslash}p{0.0765\linewidth} >{\footnotesize\raggedright\arraybackslash}p{0.0656\linewidth} >{\footnotesize\raggedright\arraybackslash}p{0.0710\linewidth} >{\footnotesize\raggedright\arraybackslash}p{0.0656\linewidth} >{\footnotesize\raggedright\arraybackslash}p{0.0820\linewidth} >{\footnotesize\raggedright\arraybackslash}p{0.0710\linewidth} >{\footnotesize\raggedright\arraybackslash}p{0.1257\linewidth}@{}}
\caption{Long Range Arena results reported by the state-space lineage.}
\label{tab:empirical} \\
\toprule
\textbf{Model} & \textbf{Source of row} & \textbf{List\-Ops} & \textbf{Text} & \textbf{Retr.} & \textbf{Image} & \textbf{Path\-finder} & \textbf{Path-X} & \textbf{Avg.\ (conv.)} \\
\midrule
\endfirsthead
\ltconthead{9}
\toprule
\textbf{Model} & \textbf{Source of row} & \textbf{List\-Ops} & \textbf{Text} & \textbf{Retr.} & \textbf{Image} & \textbf{Path\-finder} & \textbf{Path-X} & \textbf{Avg.\ (conv.)} \\
\midrule
\endhead
\midrule
\ltcontfoot{9}
\endfoot
\bottomrule
\endlastfoot

Transformer &
\citet{tay2020long}\evorig &
36.37 & 64.27 & 57.46 & 42.44 & 71.40 & fail & 54.39 (\texttt{5}) \\
\addlinespace

S4 &
\citet{gu2021efficiently}\evorig &
58.35 & 76.02 & 87.09 & 87.26 & 86.05 & 88.10 & 80.48 (\texttt{6/50}) \\
\addlinespace

S4, updated hyper\-parameters &
\citet{gu2022parameterization}\evthird &
59.60 & 86.82 & 90.90 & 88.65 & 94.20 & 96.35 & 86.09 (\texttt{6/50}) \\
\addlinespace

DSS &
\citet{gupta2022diagonal}\evorig &
\multicolumn{6}{@{}l}{\emph{softmax variant; the source does not separate per-task scores}} & 81.88 (\texttt{6/0}) \\
\addlinespace

S4D &
\citet{gu2022parameterization}\evorig &
\multicolumn{6}{@{}l}{\emph{LegS 84.89, Inv 85.50, Lin 78.39}} & (\texttt{6/50}) \\
\addlinespace

S5 &
\citet{smith2022simplified}\evorig &
\multicolumn{6}{@{}l}{\emph{S5 87.46, Inv 86.08, Lin 80.90}} & (\texttt{6/50}) \\

\end{longtable}
\tabnote{Accuracy in percent. ``Avg.\ (conv.)'' records the averaging convention the source used:
\texttt{5} is the benchmark's own five-task mean excluding Path-X, \texttt{6/50} averages six tasks
scoring a Path-X failure as 50, and \texttt{6/0} scores a failure as 0. Rows come from different papers
under different protocols; see Section~7. Evidence markers: \keyorig; \keythird.}
\endgroup

Table~\ref{tab:lm} carries the second group. The selective models, the hybrids, and the encoder report no Long Range Arena results, so they have no rows to fill in Table~\ref{tab:empirical}; what they do report is language-modeling quality, throughput, and cache behavior, and those are given here together with the configuration under which each result was obtained.

\begingroup
\footnotesize
\setlength{\tabcolsep}{4pt}
\renewcommand{\arraystretch}{1.2}
\begin{longtable}{@{}>{\footnotesize\raggedright\arraybackslash}p{0.1452\linewidth} >{\footnotesize\raggedright\arraybackslash}p{0.1192\linewidth} >{\footnotesize\raggedright\arraybackslash}p{0.3111\linewidth} >{\footnotesize\raggedright\arraybackslash}p{0.3733\linewidth}@{}}
\caption{Language-modeling, throughput and cache results, with measurement configurations.}
\label{tab:lm} \\
\toprule
\textbf{Model} & \textbf{Source of row} & \textbf{Reported result} & \textbf{Configuration under which it was measured} \\
\midrule
\endfirsthead
\ltconthead{4}
\toprule
\textbf{Model} & \textbf{Source of row} & \textbf{Reported result} & \textbf{Configuration under which it was measured} \\
\midrule
\endhead
\midrule
\ltcontfoot{4}
\endfoot
\bottomrule
\endlastfoot

S4 &
\citet{gu2021efficiently}\evorig &
WikiText-103 test perplexity 20.95 at 249M parameters, against 20.51 for a 247M Transformer. Generation throughput about 60$\times$ that Transformer &
48K against 0.8K tokens per second, single A100, batch size raised to fill memory. A throughput ratio at large batch, not a latency reduction \\
\addlinespace

BiGS &
\citet{wang2023bigs}\evorig &
Matches BERT pretraining accuracy on the GLUE average; pretrains at a 4096-token context without approximation &
Equal pretraining token budget. The only state-space work in this corpus reporting GLUE, so no cross-variant comparison is possible \\
\addlinespace

Mamba &
\citet{gu2023mamba}\evorig &
Pile validation perplexity 6.22 at 2.8B parameters. Inference throughput 4 to 5$\times$ a similarly sized Transformer &
300B tokens, context 2048. Throughput at large batch, A100 80GB PCIe, prompt 2048, 128 generated tokens \\
\addlinespace

Mamba-2 &
\citet{dao2024transformers}\evorig &
Pile validation perplexity 6.09 at 2.7B parameters, ahead of Mamba-2.8B at 6.22 and of Pythia-6.9B. Core-layer kernel 2 to 8$\times$ the Mamba scan &
300B tokens on the same data. Kernel result is a core-layer microbenchmark, A100 80GB PCIe, $N=64$, lengths 512 to 512K; batch size \nr \\
\addlinespace

Mamba-2 with attention &
\citet{dao2024transformers}\evorig &
Pile perplexity 5.95, ahead of pure Mamba-2 at 6.09 and a Transformer recipe at 6.13 &
58 state-space and 6 attention layers at 2.7B parameters, 300B tokens \\
\addlinespace

Jamba &
\citet{lieber2024jamba}\evorig &
About 3$\times$ the throughput of Mixtral &
12B active of 52B parameters, 256K context. Single A100 80GB, int8, 8K context, 512 output tokens \\
\addlinespace

Jamba-1.5 &
\citet{team2024jamba}\evorig &
Key-value cache 9~GB at a 256K context, against 80~GB for a 70B Transformer &
16-bit precision. Reports a latency curve rather than a throughput multiplier \\
\addlinespace

Samba &
\citet{ren2025samba}\evorig &
3.73$\times$ the prompt-processing throughput of Llama-3 at 1.6B scale; extrapolates from a 4K training context to 256K &
3.8B parameters, 3.2T tokens. 128K prompt, single A100, bfloat16 \\
\addlinespace

Hymba &
\citet{dong2025hymba}\evorig &
11.67$\times$ cache reduction and 3.49$\times$ throughput against Llama-3.2-3B &
8K sequence, batch 128, A100, 16-bit cache \\

\end{longtable}
\tabnote{Rows come from different papers and are not a controlled comparison. Models here report no Long Range Arena results; those that do are in Table~\ref{tab:empirical}. Evidence markers: \keyorig; \keynr.}
\endgroup

\subsection{Interpretive constraints in cross-study comparison}
\label{sec}

Four issues limit direct comparison across the reported results. First, the frequently cited Long Range Arena average of $86.09$ should not be attributed to the original S4 experiments. The appendix of the S4 paper indicates that the corresponding table was later updated using results from S4D and the HiPPO follow-up, whereas the S4 authors originally reported an average of $80.48$ with Path-X at $88.10$. Accordingly, Table~\ref{tab:empirical} reports $80.48$ for S4 and assigns $86.09$ to the source from which it was obtained. Second, the reported averages are based on different aggregation conventions. The Long Range Arena paper averages over five tasks and excludes Path-X, yielding a Transformer baseline of $54.39$. In contrast, the S4, S4D and S5 papers average over six tasks and assign a score of $50$ to Path-X failure, producing a Transformer average of $53.66$ from the same five task scores. The DSS paper assigns $0$ to its own failed tasks while reproducing the $53.66$ Transformer baseline, resulting in two scoring conventions within the same comparison. Table~\ref{tab:empirical} therefore identifies the averaging convention used for each row.

Third, nominally similar results were obtained under different experimental protocols. DSS reports unidirectional state-space models, whereas S4D and S5 use bidirectional configurations. Sequence lengths also vary across studies for the same benchmark tasks, and the Text baseline reported by Long Range Arena reflects a best-of-four selection across input lengths. In addition, the relevant papers do not report parameter counts for their Long Range Arena models, preventing verification that the compared models are matched in scale. Fourth, neither Mamba nor Mamba-2 reports Long Range Arena results. No ListOps or Pathfinder results are provided in either paper, and the corresponding entries are therefore marked \nr\ here. Their omission also indicates that later long-context language models were evaluated using benchmarks other than Long Range Arena; consequently, Mamba results should not be inserted into this benchmark without a clearly identified external source.

Beyond these four, only one benchmark in this corpus is genuinely shared, and it is not the one most often asked for. A natural request of a survey is a table of several state-space variants evaluated on
several common benchmarks. The Long Range Arena is the only benchmark in this corpus that
more than two of these models report, which is why it occupies most of
Table~\ref{tab:empirical} and why its hazards are set out at such length above. The
general language understanding evaluation suite is reported by exactly one state-space
work, the bidirectional gated encoder of \citet{wang2023bigs}, which matches BERT
pretraining accuracy on the GLUE average at an equal pretraining budget while pretraining
at a 4096-token context without approximation. One row is not a comparison, so this review
does not construct a GLUE column and invite the reader to read differences off it. The
absence is itself a finding: the selective models were developed and evaluated as
autoregressive language models, and the encoder-style understanding benchmarks were left
behind rather than contested.

\subsection{Interpretation of the empirical evidence}
The evidence supports three conclusions. First, diagonalization imposed little apparent performance cost: S4D and DSS recover S4-level Long Range Arena performance without the diagonal-plus-low-rank machinery, which helps explain why diagonal parameterizations became standard. Second, the benchmark appears to have lost discriminatory power. Once several architectures cluster in the mid-eighties and the Transformer baseline fails one task outright, the remaining differences are comparable to the variation introduced by protocol choices. Although none of the papers explicitly describes Long Range Arena as saturated, this interpretation follows from three converging observations: the benchmark authors' caution regarding their baselines, the S4 authors' instruction to consult the appendix before citing the reported table, and the subsequent abandonment of the benchmark by later models.

Third, the more durable distinction lies in inference behavior rather than benchmark accuracy. The Long Range Arena results separate these models by only a few points under nonuniform protocols, whereas the decoding-state column of Table~\ref{tab:ssm_comp} separates them by asymptotic class. That distinction, rather than the benchmark scores alone, motivates the deployment interest discussed in Section~\ref{sec:limitations}. The table does not, however, support a general claim that state-space models outperform Transformers. In the language-modeling results, S4 comes within $0.44$ perplexity of a parameter-matched Transformer on WikiText-103 while generating substantially faster, and Mamba-2 at 2.7B parameters outperforms a strong Transformer recipe trained under the same conditions. These remain two specific results at two specific scales, and Section~\ref{sec:crosscut} identifies the task classes in which the comparison favors attention.

\section{Applications}
\label{sec:applications}

Domain adaptations of state-space models now outnumber the architectural papers, and
a survey cannot cover them exhaustively without becoming a catalog. This section
takes a narrower brief. For each domain it asks the same question: what property of a
state-space model is actually being used, and against what baseline has the benefit
been demonstrated. Where controlled comparisons are unavailable, this limitation is stated explicitly. Three properties do the work in almost every case. Linear scaling in sequence length
makes long inputs tractable. A constant-size decoding state makes memory predictable.
Input-dependent selectivity allows a fixed state to be allocated to whatever the task
finds salient. Adaptations that succeed generally exploit one of these; adaptations
that report gains without exploiting any of them warrant scepticism.

Table~\ref{tab:domains} states that assessment domain by domain before the discussion
begins, so that the ordering of the subsections below is visible as a claim rather than
as an accident of how much has been published. The domains are ordered by the strength
of the evidence, not by volume of work, and the two orderings differ sharply: medical
imaging has by far the largest literature and only moderate evidential weight, while
speech has a smaller literature and the cleanest architectural argument.

\begingroup
\setlength{\tabcolsep}{4pt}
\renewcommand{\arraystretch}{1.2}
\begin{longtable}{@{}>{\footnotesize\raggedright\arraybackslash}p{0.1660\linewidth} >{\footnotesize\raggedright\arraybackslash}p{0.1660\linewidth} >{\footnotesize\raggedright\arraybackslash}p{0.1874\linewidth} >{\footnotesize\raggedright\arraybackslash}p{0.1981\linewidth} >{\footnotesize\raggedright\arraybackslash}p{0.2142\linewidth}@{}}
\caption{Application domains, ordered by strength of evidence.}
\label{tab:domains} \\
\toprule
\textbf{Domain} & \textbf{Property exploited} & \textbf{Representative work} & \textbf{Compared against} & \textbf{Evidence} \\
\midrule
\endfirsthead
\ltconthead{5}
\toprule
\textbf{Domain} & \textbf{Property exploited} & \textbf{Representative work} & \textbf{Compared against} & \textbf{Evidence} \\
\midrule
\endhead
\midrule
\ltcontfoot{5}
\endfoot
\bottomrule
\endlastfoot

Speech and audio &
Linear scaling; constant decoding state for streaming &
S4 on raw waveforms \citep{gu2021efficiently}; systematic Mamba evaluation \citep{miyazaki2024exploring} &
Transformer and convolutional speech baselines &
\textbf{Strong.} The longest raw sequences of any common modality, and the streaming advantage is architectural \\
\addlinespace

Time series and forecasting &
Linear scaling; genuinely sequential data &
MambaTS \citep{cai2024mambats} &
Transformer forecasters on standard multivariate benchmarks &
\textbf{Moderate.} Good architectural fit, but forecasting benchmarks are small and sensitive to the evaluation protocol \\
\addlinespace

Medical and volumetric imaging &
Linear scaling over very long token sequences &
U-Mamba \citep{ma2024umamba}; FusionMamba \citep{xie2024fusionmamba}; two dedicated surveys \citep{bansal2024comprehensive, heidari2024computation} &
U-Net, SegResNet, UNETR, SwinUNETR &
\textbf{Broad but shallow.} Many studies, mostly small datasets and author-selected baselines, few independent reproductions \\
\addlinespace

Vision and point clouds &
Linear scaling at high spatial resolution &
S4ND \citep{nguyen2022s4nd}; Point Cloud Mamba \citep{zhang2025pointcloudmamba}; DGMamba \citep{long2024dgmamba} &
Vision Transformers; convolutional backbones &
\textbf{Moderate.} The scan order is an imposed assumption, and it is the dominant design variable rather than the recurrence \\
\addlinespace

Anomaly detection &
Parameter efficiency at long spatial context &
MambaAD \citep{he2024mambaad} &
Per-class and multi-class detection baselines &
\textbf{Moderate.} The multi-class setting makes the parameter argument meaningful \\
\addlinespace

Graphs, including dynamic graphs &
Fixed-size summary updated as observations arrive &
Graph-Mamba \citep{wang2024graph}; DyGMamba \citep{ding2024dygmamba} &
Graph Transformers on long-range graph benchmarks &
\textbf{Thin.} Mostly static graphs with an imposed serialization; structural evolution proper is not yet addressed \\
\addlinespace

Multimodal and cross-modal &
Linear scaling over concatenated streams &
Coupled Mamba \citep{li2024coupled}; VL-Mamba \citep{qiao2024vl} &
Vision-language Transformers &
\textbf{Thin.} The efficiency argument holds; the alignment operation is retrieval, which the rank bound limits, and this is untested \\
\addlinespace

Language beyond English &
None specific; tests the capacity of a fixed-width state &
Indic structured question answering \citep{vats2025multilingual} &
Transformer baselines in the same languages &
\textbf{Thin and important.} Morphologically richer languages place more demand per token on a fixed state \\
\addlinespace

Structured and tabular data &
\textbf{None of the three}; there is no sequence to exploit &
MambaTab \citep{ahamed2024mambatab}; crash-severity classifiers \citep{Somvanshi2025YM, Somvanshi2025} &
Gradient-boosted trees; TabNet, FT-Transformer, TransTab &
\textbf{Weakest.} Reported gains rest on parameter efficiency rather than on sequence modeling, and results are mixed \\

\end{longtable}
\tabnote{``Property exploited'' names which of the three properties set out at the head of Section~\ref{sec:applications} the adaptation actually uses. ``Evidence'' is this review's own assessment, not a claim by the cited authors.}
\endgroup

\subsection{Naturally sequential signals: speech and time series}

These are the domains where the inductive bias is not an assumption but a description of the data. Speech has the longest raw sequences of any common modality. S4 demonstrated early
that a state-space model could classify raw waveforms directly rather than
hand-engineered features \citep{gu2021efficiently}, which is a result about the
architecture rather than about a pipeline: a sixteen-kilohertz waveform is exactly the
regime in which quadratic attention is unaffordable and a linear recurrence is not.
Systematic evaluation of Mamba across speech tasks finds it competitive with
Transformer baselines while retaining the streaming advantage that constant-state
decoding provides \citep{miyazaki2024exploring}. The streaming property is worth
separating from the accuracy result, because it is architectural: a recogniser whose
decoding state does not grow can run indefinitely on a fixed memory budget, which no
attention-based recogniser can do without windowing.

Forecasting is the other natural fit, since the data are genuinely sequential and often
long. MambaTS adapts the selective mechanism to multivariate forecasting and, in doing
so, engages with a problem that does not arise in language: the ordering of variables is
arbitrary, so a scan across them imposes a structure the data do not have, and the model
must be designed not to depend on it \citep{cai2024mambats}. This is the pattern that
distinguishes a serious domain adaptation from a substitution. The adaptation identifies
a mismatch between the architecture's assumptions and the data, and modifies the
architecture rather than reporting around the mismatch.

The caution for both domains is the same and is evidential rather than architectural.
Forecasting benchmarks are small, sensitive to the normalization and windowing choices,
and have a documented history of apparent improvements that do not survive a controlled
re-run. The architectural argument here is stronger than the benchmark evidence, and
those two things should be kept apart.

\subsection{Spatial and volumetric data: vision, medical imaging and anomaly detection}

Vision was the first domain to adopt state-space backbones at scale, and it is also
where the adaptation is least natural, because images have no canonical sequence
order. The design problem is therefore the scan order rather than the recurrence, a
point made explicitly by the vision surveys, which treat scanning strategy as the
critical design axis \citep{xu2024visualmamba, zhang2024surveyvisualmamba}. S4ND takes
the alternative route of extending the kernel to multiple dimensions directly rather
than serializing the image \citep{nguyen2022s4nd}, which is the more principled response
and, tellingly, the less widely adopted one. Point Cloud Mamba applies a state-space
backbone to point-cloud learning, where the linear scaling matters because point counts
are large and irregular \citep{zhang2025pointcloudmamba}. DGMamba addresses domain
generalization, arguing that the hidden states and the scanning pattern of a vision
state-space model are themselves sources of domain-specific overfitting, and modifying
both \citep{long2024dgmamba}. The second is the more interesting for this review,
because it identifies a failure mode specific to the architecture rather than
transferring a generic technique onto it.

Medical imaging is the most heavily worked domain, with two dedicated surveys covering
it \citep{bansal2024comprehensive, heidari2024computation}. The attraction is
concrete: volumetric scans produce very long token sequences, and the quadratic cost
of attention is the binding constraint. U-Mamba is representative. It places Mamba
blocks inside a U-Net encoder so that convolutional layers capture local texture while
the recurrence carries context across distant regions, and its discrete-time formulation
is the standard one,
\begin{equation}
x_k = A_d x_{k-1} + B_d u_k, \qquad y_k = C_d x_k ,
\end{equation}
with \( A_d, B_d, C_d \) the discretized parameters \citep{ma2024umamba}. Its
authors report gains over the convolutional baselines U-Net and SegResNet and the
Transformer baselines UNETR and SwinUNETR across the segmentation tasks they evaluate,
and the network self-configures its hyperparameters per dataset, which removes one source
of tuning advantage but not the baseline-selection question.

\begin{figure}[pos=H]
    \centering
    \includegraphics[width=0.98\linewidth]{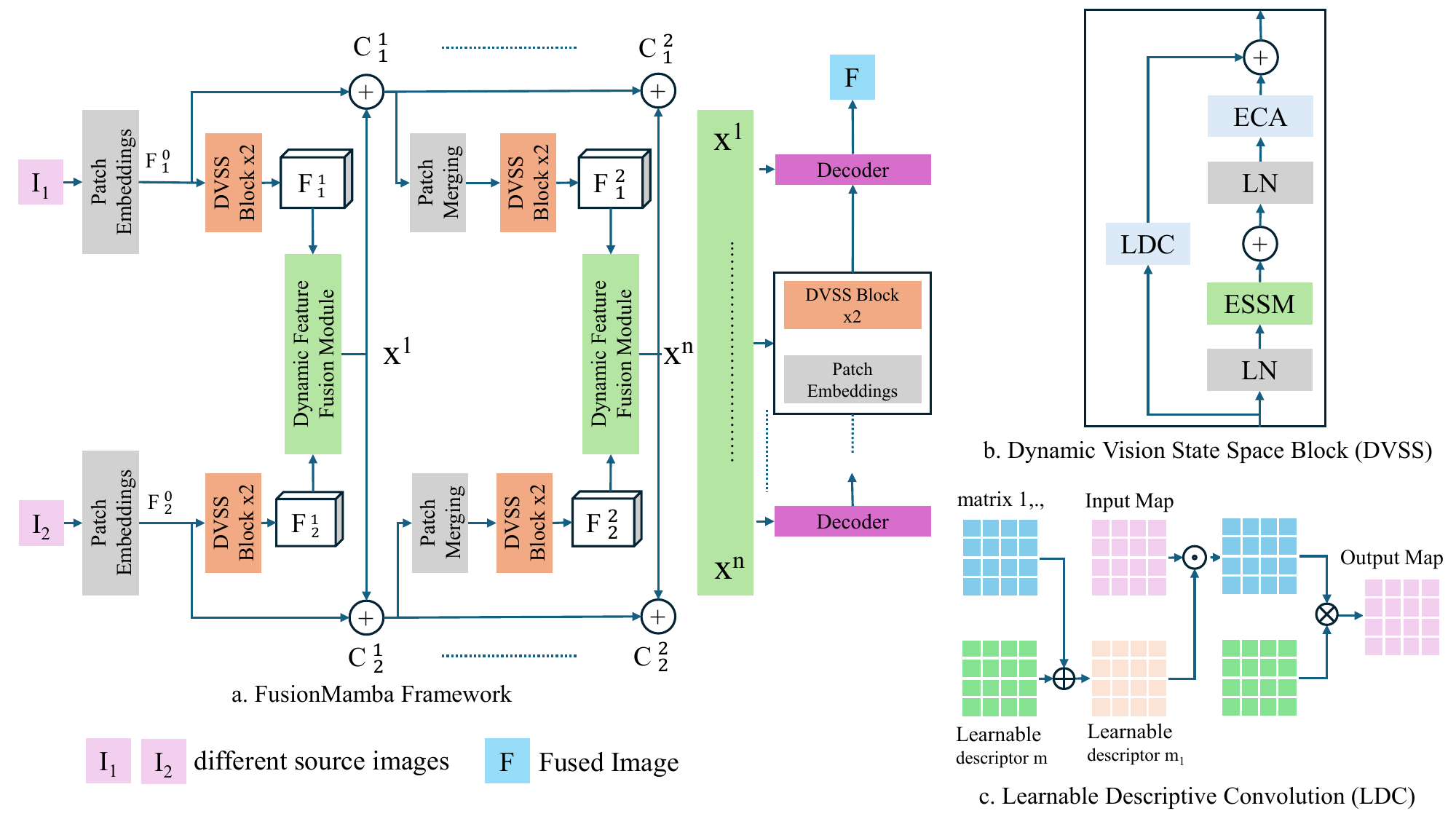}
    \caption{Overview of the FusionMamba framework adapted from \citep{xie2024fusionmamba}}
    \label{fig:fusionmamba}
\end{figure}

FusionMamba applies the same reasoning to multimodal fusion. A dynamic feature fusion
module adjusts texture and disparity perception across modalities, and a cross-modal
fusion block built on a selective state-space layer enhances inter-modal correlation
while suppressing redundancy \citep{xie2024fusionmamba}. Figure~\ref{fig:fusionmamba}
shows the arrangement: two source images \(I_1\) and \(I_2\) pass through dynamic vision
state-space blocks for feature extraction, learnable descriptive convolution refines
texture, the fusion module combines the streams, and a decoder of state-space and
patch-expanding blocks reconstructs the fused image \(F\). Its authors report improvements over the convolutional and Transformer
baselines they evaluated on infrared-visible, computed-tomography with magnetic-resonance,
positron-emission with magnetic-resonance, and single-photon-emission with
magnetic-resonance fusion. No controlled comparison across fusion methods under a shared
protocol has been run, here or in the surveys.

MambaAD extends the same spatial argument to multi-class unsupervised anomaly detection,
using a pyramidal state-space decoder in which one model covers multiple object classes
rather than one model per class \citep{he2024mambaad}. The property exploited is
parameter efficiency at long spatial context, and the multi-class setting is what makes
the comparison meaningful, since per-class models trade the problem for storage. A caution belongs here rather than in the limitations section, because it is specific
to this literature. Medical-imaging results are typically reported on small datasets
against baselines the authors selected, and the surveys above collect those reports
rather than re-running them. The evidence that state-space backbones help in medical
imaging is broad but shallow, and Table~\ref{tab:domains} records it as such.

\subsection{Language, multimodality and cross-modal interaction}
\label{sec:crossmodal}

Three settings belong together because they turn on the same question: whether a state
of fixed width has the capacity the task demands. Most of the language-modeling evidence for state-space models is English. Evaluation
of state-space models for structured question answering across Indic languages is one
of the few systematic looks at whether the architecture's behavior carries across
morphologically richer languages and lower-resource settings
\citep{vats2025multilingual}. This matters for the survey's argument because the
fixed-width state is a capacity constraint, and languages with richer morphology place
more demand on that capacity per token. The question is open and under-studied, and it is
one of the few places where the rank bound of Section~\ref{sec:ssd} makes a prediction
that the literature has not yet tested.

Cross-modal work asks the state to carry information from more than one stream.
Coupled Mamba allows the state of one modality's recurrence to influence another's
while keeping the per-modality dynamics separate, which preserves the parallel
evaluation that a fully coupled formulation would lose \citep{li2024coupled}, and
VL-Mamba substitutes a state-space backbone into a vision-language model
\citep{qiao2024vl}. The efficiency argument is straightforward, since concatenated
multimodal sequences are long. The open question is different and is worth stating:
cross-modal reasoning often requires aligning specific elements across streams, which is
a retrieval operation, and Section~\ref{sec:crosscut} gives the reason to expect a
fixed-width state to be weak at exactly that. The reviewed cross-modal work does not yet
test this directly.

A related gap appears where the relationship between modalities changes over time.
Surveys of multimodal fake-news detection organize that field around exactly this
cross-modal interaction \citep{li2025mfndsurvey}, and the interaction is often what
carries the signal. State-space models are a plausible fit, since a recurrent state can
in principle track a drifting relationship, and the drift is precisely what a static
fusion mechanism cannot represent. As with dynamic graphs below, the fit is argued rather
than demonstrated, and this review does not claim otherwise.

\subsection{Dynamic graphs and temporally evolving structure}
\label{sec:dyngraph}

This is the setting in which the architectural case for a recurrent state is strongest
and the empirical record is thinnest, so it is worth separating the two carefully. The problem setting is established independently of state-space models. Representation
learning on dynamic graphs must track structure that evolves over time
\citep{gao2023dgcn}, and community detection in dynamic graphs must do so while
resolving structure at multiple scales, since a community that splits or merges changes
the object being represented rather than merely its features
\citep{teng2026multiscale}. Neither of those works uses a state-space model; they are
cited here as the problem statement, not as evidence about state-space methods. The architectural connection is direct, and it is the cleanest such argument in this
section. A dynamic graph is a sequence of structural observations, and maintaining a
fixed-size summary that is updated as observations arrive is exactly what a recurrent
state does. A Transformer over the same stream must either re-encode the history at each
step or maintain a cache that grows with the number of observations, so the asymptotic
argument of Section~\ref{sec:crosscut} applies here with unusual directness: the quantity
that grows is the interaction history, and the quantity that does not grow is the state.

Two adaptations illustrate this potential across distinct graph-learning settings. Graph-Mamba replaces the attention module within a graph-Transformer framework with a selective state-space block and introduces node prioritization and permutation-based sparsification to impose a meaningful scan order \citep{wang2024graph}. DyGMamba provides a more direct evaluation in continuous-time dynamic graphs by composing two state-space layers: one models the sequence of a node's historical interactions, and the other captures temporal evolution. This design enables selective retention of relevant temporal patterns rather than uniform weighting of the full interaction history. Across seven continuous-time datasets---Wikipedia, Reddit, MOOC, LastFM, Enron, UCI and Social Evolution---the authors report competitive dynamic link-prediction performance against ten baselines, including JODIE, DyRep, TGAT, TGN, CAWN and DyGFormer, while using a lower computational budget over the same temporal horizon \citep{ding2024dygmamba}.

The current evidence indicates a promising but still limited application base, with two specific research gaps. Existing evaluations focus on dynamic link prediction, where the objective is to determine whether an edge will appear and a fixed-width state is well suited to summarizing a node's interaction history. More demanding settings remain unresolved. The first is structural evolution, in which changing community structure must be represented directly rather than recomputed independently at each step; none of the reviewed state-space methods addresses this problem. The second is temporal and spatio-temporal knowledge graph reasoning, where typed entities and relations must support retrieval of a specific fact at a specific time. As discussed in Section~\ref{sec}, this is a retrieval-intensive setting in which the rank bound may constrain fixed-width state representations, suggesting that hybrid architectures with a limited attention component are a natural direction for evaluation. The architectural rationale is therefore stronger than the current empirical coverage, which remains concentrated on one task class.

\subsection{Beyond sequence data: structured and tabular prediction}

The final domain is included because the current evidence provides a comparatively limited basis for a distinct architectural advantage. Recognizing such boundaries is important to a balanced assessment of where the approach is most, and least, well supported. State-space backbones have been applied to structured tabular prediction, where none of
the three properties named at the head of this section is available: a tabular record has
no sequence to scale over, no streaming decode, and no temporal salience for selectivity
to exploit. MambaTab is the clearest external example. It applies a Mamba block to the
feature vector of a record and reports results on eight public benchmarks against
fourteen baselines spanning gradient-boosted trees, multilayer perceptrons and the
tabular Transformer family, including TabNet, FT-Transformer and TransTab
\citep{ahamed2024mambatab}. Its authors' own framing is instructive: the reported
advantage is attributed to parameter efficiency, with comparable or better accuracy than
TransTab at a small fraction of the learnable parameters, rather than to any sequence
modeling capability.

In transportation safety, state-space classifiers have been evaluated for crash-severity
prediction for motorcyclists and for child bicyclists \citep{Somvanshi2025YM,
Somvanshi2025}. These are small studies, and their own results are mixed: in the
motorcyclist study the state-space model is competitive with, but does not exceed, the
strongest tabular baseline. Read beside MambaTab, they fit the same pattern, in which the
architecture is applied outside sequence data and the reported benefit, where there is
one, is a parameter-count argument rather than a sequence-modeling one. They are
reported here as an early indication that the architecture can be applied outside
sequence data, not as evidence that it should be, and the reader who wants a
demonstrated advantage on tabular data should not look to this family for it.

\section{Limitations, failure modes and deployment}
\label{sec:limitations}

\subsection{Capability failure modes}

The failure modes of this family are structural rather than incidental, which is what
makes them worth stating precisely, and what distinguishes them from the ordinary
shortfalls that scale eventually removes. Three of the four below follow from a single
fact established in Section~\ref{sec:ssd}: everything a state-space layer passes from an
earlier position to a later one goes through a state of fixed width $N$, which the
semiseparable characterization states as a rank bound. A larger state raises the
ceiling and no finite state removes it, which is why the hybrid designs of
Section~\ref{sec:hybrids} exist. The fourth, order sensitivity, has a different origin:
it arises when a sequential recurrence is applied to data that have no canonical order,
so the scan itself becomes an unjustified modeling assumption.
Table~\ref{tab:failuremodes} sets out each failure with its cause, what mitigates it,
and what remains after mitigation, since it is the residual rather than the failure that
determines when the family is the wrong choice.

\begingroup
\setlength{\tabcolsep}{4pt}
\renewcommand{\arraystretch}{1.2}
\begin{longtable}{@{}>{\footnotesize\raggedright\arraybackslash}p{0.1745\linewidth} >{\footnotesize\raggedright\arraybackslash}p{0.2617\linewidth} >{\footnotesize\raggedright\arraybackslash}p{0.2399\linewidth} >{\footnotesize\raggedright\arraybackslash}p{0.2726\linewidth}@{}}
\caption{Capability failure modes of the state-space family.}
\label{tab:failuremodes} \\
\toprule
\textbf{Failure mode} & \textbf{Why it happens} & \textbf{What mitigates it} & \textbf{Residual after mitigation} \\
\midrule
\endfirsthead
\ltconthead{4}
\toprule
\textbf{Failure mode} & \textbf{Why it happens} & \textbf{What mitigates it} & \textbf{Residual after mitigation} \\
\midrule
\endhead
\midrule
\ltcontfoot{4}
\endfoot
\bottomrule
\endlastfoot

Exact retrieval and associative recall &
A state of fixed width $N$ is a hard capacity bound on what can be carried forward, stated as a rank bound by the semiseparable characterization &
Larger state size: recall improves monotonically as $N$ grows from 16 to 256 in Mamba-2. A small fraction of attention layers supplies unbounded exact lookup &
The bound is raised, never removed. A pure state-space model of any finite state size still fails at sufficiently long exact recall \\
\addlinespace

Copying and precise span manipulation &
The same bound. Attention addresses positions directly; a recurrent state must have anticipated that the span would be needed &
Hybrid designs, or a windowed attention layer when the spans of interest are local &
Windowed attention gives no exact copying beyond the window; global attention restores a cache that grows with context \\
\addlinespace

In-context learning over discrete tokens &
Related to the same bound, but confounded with training data and scale, so not cleanly attributable to architecture alone &
Hybridization; larger state; more training data &
This review treats the gap as real but not well isolated, and the isolating experiment has not been published \\
\addlinespace

Order sensitivity in non-sequential data &
A sequential recurrence applied to images, graphs or tabular records requires choosing a scan, and the architecture does not justify the choice &
Multi-directional and learned scan orders; multi-dimensional kernels such as S4ND that avoid serializing at all &
Scanning strategy remains the dominant design variable in vision, which is a symptom of the problem rather than a solution to it \\

\end{longtable}
\endgroup

The distinction between the first three rows and the fourth matters for how the field
should read its own progress. Increasing state size, adding attention layers and
improving scan design are all real improvements, but the first two move a bound and the
third removes an assumption. Only the third is the kind of problem that a better
architecture could close entirely.

\subsection{Evidence, training and reproducibility limits}

The limits in this subsection belong to the evidence base rather than to the models, and
they are the reason this review reports configurations rather than headline numbers. Three of them compound: an architecture that is hard to train reliably is also hard to evaluate fairly, and an evidence base assembled by copying reported results between papers preserves whatever errors the first copy introduced.

The first is sensitivity to initialization. State-space layers are more sensitive to it than the architectures they compete with. This is why HiPPO exists, why
S4's parameterization is intricate, and why the diagonal successors needed a derived
initialization rather than a random one. Stable reparameterization of the state
transition is an active theoretical concern in its own right \citep{wang2023stablessm},
and the theory of what selective state-space models can represent is still developing. The practical consequence is that the optimization
recipes, learning-rate schedules and failure diagnostics that practitioners have
accumulated for Transformers do not transfer directly, so a failed training run is
harder to diagnose here than it would be for an equivalent Transformer.

The second is benchmark dependence. Section~7 sets out the specific hazards in
the Long Range Arena results, and they generalize. Averages computed under different
conventions circulate under the same name, protocol differences hide beneath
identical-looking rows, and the benchmark that defined this family's early evaluation was
quietly abandoned by the models that succeeded at long-context language modeling. No
benchmark has replaced it, which leaves long-context claims resting on perplexity and on
retrieval probes that the models were partly designed against.

The third, and the most consequential for a reader assembling evidence from several papers, is reproducibility. Headline efficiency numbers are frequently
single-paper, single-device, kernel-level measurements, and they do not transfer to other
hardware or to end-to-end settings. More seriously, a widely quoted result for one model
in this family originates in a different paper about a different model, as
Section~7 documents. That is not an isolated slip. It suggests that
cross-paper table stitching is common enough to be a systematic risk in this literature,
and it is the direct reason every quantitative cell in Table~\ref{tab:empirical} names
the paper it came from.

\subsection{Deployment and practical adoption}

The case for state-space models is strongest in deployment, and the obstacles are also
concentrated there. This subsection addresses them directly, since it is the aspect least
covered by existing surveys. The four that follow are ordered from the one where the
architecture transfers least well to the one where it transfers best, and then to the
obstacle that has nothing to do with the architecture at all.

Quantization is where the architecture's advantage is least transferable. In a Transformer the dominant quantization targets are weights and the
key-value cache, both well understood. In a state-space model the recurrent state is
carried across every timestep, so quantization error in the state does not stay local:
it is fed back through the recurrence and can accumulate over thousands of steps. The
state is also the model's entire memory, so it has less redundancy to absorb error than
a weight matrix does. Reduced-precision inference for these models therefore needs
treatment that separates weights from the propagated state, and the published evidence
is thinner than for Transformer quantization. This review flags it as an open
engineering problem rather than reporting a settled result.

Distributed training and inference run the other way, because here the duality of Section~\ref{sec:ssd}
changes the outlook materially. Once the state-space layer is expressed as structured
masked attention, the tensor and sequence parallelism strategies developed for
Transformers apply to it, which removes what would otherwise be a serious obstacle to
training at scale. Serving is a different matter: the constant decoding state makes
per-request memory predictable, which simplifies batching and capacity planning, but
the sequential dependence of the recurrence constrains how decoding can be split across
devices.

Edge and memory-constrained deployment is the clearest practical case for the family. A decoding state that does not grow with context turns long-context inference on a fixed memory budget from an impossibility into a fixed cost, and the reported cache results, 9~GB against 80~GB at a 256K context in one hybrid and an elevenfold reduction in another, are large enough to change what fits on a device. Two caveats apply. The
constant state is not a small state, since a large state size is what makes selective
models effective, and the advantage only materializes at context lengths where the
Transformer's cache is genuinely the constraint.

The most underestimated obstacle, finally, is not architectural at all.
Transformer deployment rests on a decade of accumulated tooling: fused attention
kernels, serving frameworks, quantization toolchains, parameter-efficient fine-tuning
methods, and a large body of practitioner knowledge about what to do when training
diverges. State-space models have a fraction of this, though the gap has narrowed and it is worth
being precise about where it remains. Mainstream library support now exists: the
Hugging Face Transformers library ships model classes for Mamba, Mamba-2, Jamba and
several derived hybrids, so loading and fine-tuning a released checkpoint is no longer the
obstacle it was. What remains uneven is everything downstream of loading. Competitive
inference depends on custom fused kernels that must be maintained against changing
hardware, and the relationship between those kernels and the portable fallback is not the
simple one the literature implies: for Mamba-2 the library documents the compiled PyTorch
path as several times faster than the original CUDA kernels in ordinary use, because the
kernels carry a warm-up cost at prefill. A practitioner therefore cannot assume that the
published kernel speedups are what they will observe. Parameter-efficient fine-tuning
methods designed around attention projections also do not map cleanly onto a selective
recurrence, since there is no query or key matrix to adapt, and quantization toolchains
target the weights and the key-value cache rather than a propagated state. Distillation offers a partial route around this, since a pretrained
Transformer can be converted into a state-space or hybrid student for a small fraction
of pretraining cost \citep{bick2024transformers}, which lowers the entry cost without
addressing the tooling gap itself.

\subsection{Guidance: choosing between attention, state spaces and hybrids}

Taken together, the preceding analysis provides a basis for informed model selection. Table~\ref{tab:guidance} translates these considerations into workload-oriented guidance, reflecting the practical decisions users face when selecting among architectures.

\begingroup
\setlength{\tabcolsep}{4pt}
\renewcommand{\arraystretch}{1.2}
\begin{longtable}{@{}>{\footnotesize\raggedright\arraybackslash}p{0.1636\linewidth} >{\footnotesize\raggedright\arraybackslash}p{0.3381\linewidth} >{\footnotesize\raggedright\arraybackslash}p{0.2181\linewidth} >{\footnotesize\raggedright\arraybackslash}p{0.2290\linewidth}@{}}
\caption{Choosing between attention, state-space and hybrid architectures.}
\label{tab:guidance} \\
\toprule
\textbf{Choose} & \textbf{When the workload looks like this} & \textbf{Because} & \textbf{Main cost of the choice} \\
\midrule
\endfirsthead
\ltconthead{4}
\toprule
\textbf{Choose} & \textbf{When the workload looks like this} & \textbf{Because} & \textbf{Main cost of the choice} \\
\midrule
\endhead
\midrule
\ltcontfoot{4}
\endfoot
\bottomrule
\endlastfoot

\textbf{Attention} &
Context short enough that the quadratic term is not binding; the task is dominated by exact retrieval, verbatim copying or in-context learning over discrete tokens; or delivery schedule makes mature tooling decisive &
Exact addressing of any earlier position is the mechanism the task relies on, and no fixed-width state provides it &
Quadratic training cost, and a key-value cache that grows linearly with context and eventually exhausts the device \\
\addlinespace

\textbf{State space} &
Sequences are long; decoding memory rather than arithmetic is the binding constraint; inputs are continuous or naturally smooth, as in audio, sensor and biological signals; deployment is memory-limited and the workload is streaming &
Decoding cost and memory are constant in the context length, which is an architectural property and transfers across hardware &
Bounded exact recall; a narrower tooling ecosystem; custom kernels needed for competitive speed \\
\addlinespace

\textbf{Hybrid} &
The workload has both characteristics, which describes most general-purpose language modeling &
A small attention fraction, on the order of ten percent of layers, restores retrieval capability while keeping the cache bounded &
Two architectures to tune, and the attention fraction becomes an additional hyperparameter \\

\end{longtable}
\endgroup

Two aspects of the table require clarification. The hybrid row reflects the dominant design pattern in general-purpose language modeling. Evidence reviewed in Section~\ref{sec:hybrids} indicates that a small attention fraction, approximately ten percent of layers, can restore retrieval capability while preserving a bounded cache. The tooling criterion in the first row likewise represents a practical engineering consideration: under deployment constraints, ecosystem maturity may outweigh asymptotic advantages and should therefore be included in model-selection guidance.

\section{Conclusions}

State-space sequence models have evolved through a series of architectural developments, each addressing specific limitations of earlier formulations. HiPPO introduced a principled initialization; S4 made the resulting operator computationally tractable; diagonal formulations showed that much of the original machinery could be simplified once the state matrix was diagonalized; S5 restored parallel execution through scan-based computation; selectivity introduced content dependence while removing time invariance and the associated convolutional formulation; and structured state-space duality recovered efficient hardware utilization under the resulting computational constraints. Selectivity marks the principal qualitative transition in this progression, while structured state-space duality provides its unifying analytical result.

The cross-cutting evidence supports three broader conclusions. First, the defining deployment advantage of this model family lies in its fixed-width decoding state rather than in operation counts alone. Comparisons framed only as quadratic versus linear complexity obscure the more consequential distinction: a Transformer's key-value cache grows with context length, whereas the recurrent state of a state-space model does not. Reported cache reductions, including approximately ninefold at a 256K context in one hybrid architecture and elevenfold relative to a small Transformer in another, follow from this architectural property and are therefore more transferable across hardware than implementation-specific throughput multipliers. For this reason, the efficiency results in this review are reported with their baseline, device and measurement scope.

Second, state-space models are not general substitutes for attention. The limitation is structural: all information transmitted from an earlier position to a later one must pass through a fixed-width state, and the semiseparable representation expresses this restriction as a rank bound. Tasks requiring exact retrieval of a specific earlier element therefore remain a characteristic weakness. Increasing the state dimension raises the effective ceiling but does not remove the bound. In areas where the empirical record remains limited, including in-context learning and morphologically rich or lower-resource languages, the review does not extend conclusions beyond the available evidence.

Third, hybridization represents the most consistent architectural convergence in the literature. Structured state-space duality explains this pattern by showing that state-space and attention mechanisms belong to the same broader operator class but differ in the rank constraints imposed on information transfer. Attention layers provide the exact retrieval capability that a rank-bounded recurrent state lacks. This interpretation is consistent with the repeated observation that architectures containing a small but nonzero fraction of attention layers often outperform both purely recurrent and purely attention-based alternatives. The resulting research direction is therefore not the complete removal of attention, but the principled placement, frequency and sparsity of attention within recurrent backbones.

The remaining research agenda spans capability, evidence and engineering. Capability questions include whether more effective state allocation can raise the recall ceiling without uniformly increasing state size, and whether selectivity and depth-dependent gating address the same representational limitation or distinct ones. Evidence questions remain equally important: the benchmark that shaped the early development of this literature no longer clearly distinguishes among leading models, yet no broadly accepted replacement has emerged. As a result, many long-context claims continue to rely on perplexity measures and retrieval probes that overlap with the capabilities targeted during model design. Engineering challenges include quantization, for which recurrent error accumulation presents a distinctive difficulty because state perturbations are propagated through subsequent updates and the recurrent state may provide less redundancy than a large weight matrix.

At present, the difference between the demonstrated capabilities of state-space models and their practical adoption is driven increasingly by ecosystem maturity rather than by architectural feasibility. Progress will therefore depend not only on new model formulations, but also on advances in kernels, serving systems, quantization, fine-tuning methods and evaluation protocols that make the architectural advantages measurable and usable in practice.

\section*{Acknowledgments}
We extend our sincere gratitude to Biplov Pandey for his assistance in preparing the images for this paper. We also thank our colleagues at Texas State University for their valuable guidance throughout this work.


\bibliographystyle{plainnat}

\bibliography{cas-refs}

\begin{thebibliography}{100}
\providecommand{\natexlab}[1]{#1}
\providecommand{\url}[1]{\texttt{#1}}
\expandafter\ifx\csname urlstyle\endcsname\relax
  \providecommand{\doi}[1]{doi: #1}\else
  \providecommand{\doi}{doi: \begingroup \urlstyle{rm}\Url}\fi

\bibitem[Agrawal and Mittal(2020)]{agrawal2020using}
Abhinav Agrawal and Namita Mittal.
\newblock Using cnn for facial expression recognition: A study of the effects
  of kernel size and number of filters on accuracy.
\newblock \emph{The Visual Computer}, 36\penalty0 (2):\penalty0 405--412, 2020.

\bibitem[Ahamed and Cheng(2024)]{ahamed2024mambatab}
Md~Atik Ahamed and Qiang Cheng.
\newblock Mambatab: A plug-and-play model for learning tabular data.
\newblock In \emph{2024 IEEE 7th International Conference on Multimedia
  Information Processing and Retrieval (MIPR)}, pages 369--375. IEEE, 2024.
\newblock \doi{10.1109/MIPR62202.2024.00065}.

\bibitem[Bain et~al.(2021)Bain, Nagrani, Varol, and Zisserman]{bain2021frozen}
Max Bain, Arsha Nagrani, G{\"u}l Varol, and Andrew Zisserman.
\newblock Frozen in time: A joint video and image encoder for end-to-end
  retrieval.
\newblock In \emph{Proceedings of the IEEE/CVF International Conference on
  Computer Vision}, pages 1728--1738, 2021.

\bibitem[Bansal et~al.(2024)Bansal, Sreeharish, Madhava~Prasath, Manikandan,
  Madisetty, Rehman, Raghaw, Duggal, and Kumar]{bansal2024comprehensive}
Shubhi Bansal, A~Sreeharish, J~Madhava~Prasath, S~Manikandan, Sreekanth
  Madisetty, Mohammad Zia~Ur Rehman, Chandravardhan~Singh Raghaw, Gaurav
  Duggal, and Nagendra Kumar.
\newblock A comprehensive survey of mamba architectures for medical image
  analysis: Classification, segmentation, restoration and beyond, 2024.

\bibitem[Bao et~al.(2025)Bao, Lyu, Xu, Zhou, Ren, Xiang, Li, and
  Cheng]{bao2025visionmambaremotesensing}
Muyi Bao, Shuchang Lyu, Zhaoyang Xu, Huiyu Zhou, Jinchang Ren, Shiming Xiang,
  Xiangtai Li, and Guangliang Cheng.
\newblock Vision mamba in remote sensing: A comprehensive survey of techniques,
  applications and outlook.
\newblock \emph{arXiv preprint arXiv:2505.00630}, 2025.
\newblock \doi{10.48550/arXiv.2505.00630}.

\bibitem[Becker et~al.(2024)Becker, Freymuth, and Neumann]{becker2024kalmamba}
Philipp Becker, Niklas Freymuth, and Gerhard Neumann.
\newblock Kalmamba: Towards efficient probabilistic state space models for rl
  under uncertainty, 2024.

\bibitem[Bengio et~al.(1994)Bengio, Simard, and Frasconi]{bengio1994learning}
Yoshua Bengio, Patrice Simard, and Paolo Frasconi.
\newblock Learning long-term dependencies with gradient descent is difficult.
\newblock \emph{IEEE Transactions on Neural Networks}, 5\penalty0 (2):\penalty0
  157--166, 1994.

\bibitem[Bhat(2024)]{bhat2024mathematical}
Siddhanth Bhat.
\newblock Mathematical formalism for memory compression in selective state
  space models, 2024.

\bibitem[Bick et~al.(2024)Bick, Li, Xing, Kolter, and Gu]{bick2024transformers}
A.~Bick, K.~Y. Li, E.~P. Xing, J.~Z. Kolter, and A.~Gu.
\newblock Transformers to {SSMs}: Distilling quadratic knowledge to
  subquadratic models.
\newblock In \emph{Advances in Neural Information Processing Systems 37
  (NeurIPS 2024)}, 2024.
\newblock URL
  \url{https://proceedings.neurips.cc/paper_files/paper/2024/hash/3848fef259495bfd04d60cdc5c1b4db7-Abstract-Conference.html}.

\bibitem[Botev et~al.(2024)Botev, De, Smith, Fernando, Muraru,
  et~al.]{botev2024recurrentgemma}
A.~Botev, S.~De, S.~L. Smith, A.~Fernando, G.-C. Muraru, et~al.
\newblock Recurrentgemma: Moving past transformers for efficient open language
  models.
\newblock \emph{arXiv preprint arXiv:2404.07839}, 2024.
\newblock \doi{10.48550/arXiv.2404.07839}.
\newblock URL \url{https://arxiv.org/abs/2404.07839}.

\bibitem[Brown et~al.(2020)Brown, Mann, Ryder, Subbiah, Kaplan, Dhariwal,
  Neelakantan, Shyam, Sastry, Askell, et~al.]{brown2020language}
Tom Brown, Benjamin Mann, Nick Ryder, Melanie Subbiah, Jared~D. Kaplan,
  Prafulla Dhariwal, Arvind Neelakantan, Pranav Shyam, Girish Sastry, Amanda
  Askell, et~al.
\newblock Language models are few-shot learners.
\newblock In \emph{Advances in Neural Information Processing Systems},
  volume~33, pages 1877--1901, 2020.

\bibitem[Cahuantzi et~al.(2023)Cahuantzi, Chen, and
  G{\"u}ttel]{cahuantzi2023comparison}
Roberto Cahuantzi, Xinye Chen, and Stefan G{\"u}ttel.
\newblock A comparison of lstm and gru networks for learning symbolic
  sequences.
\newblock In \emph{Science and Information Conference}, pages 771--785.
  Springer, 2023.

\bibitem[Cai et~al.(2024)Cai, Zhu, Wang, and Yao]{cai2024mambats}
Xiuding Cai, Yaoyao Zhu, Xueyao Wang, and Yu~Yao.
\newblock Mambats: Improved selective state space models for long-term time
  series forecasting, 2024.

\bibitem[Chen et~al.(2018)Chen, Rubanova, Bettencourt, and
  Duvenaud]{chen2018neural}
Ricky T.~Q. Chen, Yulia Rubanova, Jesse Bettencourt, and David~K. Duvenaud.
\newblock Neural ordinary differential equations.
\newblock In \emph{Advances in Neural Information Processing Systems},
  volume~31, 2018.

\bibitem[Chung et~al.(2014)Chung, Gulcehre, Cho, and
  Bengio]{chung2014empirical}
Junyoung Chung, Caglar Gulcehre, KyungHyun Cho, and Yoshua Bengio.
\newblock Empirical evaluation of gated recurrent neural networks on sequence
  modeling.
\newblock \emph{arXiv preprint arXiv:1412.3555}, 2014.
\newblock Presented at the NIPS 2014 Deep Learning and Representation Learning
  Workshop.

\bibitem[Cirone et~al.(2024)Cirone, Orvieto, Walker, Salvi, and
  Lyons]{muca2024theoretical}
Nicola~Muca Cirone, Antonio Orvieto, Benjamin Walker, Cristopher Salvi, and
  Terry Lyons.
\newblock Theoretical foundations of deep selective state-space models.
\newblock \emph{Advances in Neural Information Processing Systems},
  37:\penalty0 127226--127272, 2024.
\newblock \doi{10.52202/079017-4041}.

\bibitem[Dai et~al.(2019)Dai, Yang, Yang, Carbonell, Le, and
  Salakhutdinov]{dai2019transformer}
Zihang Dai, Zhilin Yang, Yiming Yang, Jaime Carbonell, Quoc Le, and Ruslan
  Salakhutdinov.
\newblock Transformer-{XL}: Attentive language models beyond a fixed-length
  context.
\newblock In \emph{Proceedings of the 57th Annual Meeting of the Association
  for Computational Linguistics}, pages 2978--2988, Florence, Italy, July 2019.
  Association for Computational Linguistics.
\newblock \doi{10.18653/v1/P19-1285}.
\newblock URL \url{https://aclanthology.org/P19-1285/}.

\bibitem[Dao and Gu(2024)]{dao2024transformers}
Tri Dao and Albert Gu.
\newblock Transformers are {SSM}s: Generalized models and efficient algorithms
  through structured state space duality.
\newblock In \emph{Proceedings of the 41st International Conference on Machine
  Learning}, volume 235 of \emph{Proceedings of Machine Learning Research},
  pages 10041--10071. PMLR, 2024.
\newblock URL \url{https://proceedings.mlr.press/v235/dao24a.html}.

\bibitem[De et~al.(2024)De, Smith, Fernando, Botev, Muraru, Gu, Haroun,
  Berrada, Chen, Srinivasan, Desjardins, Doucet, Budden, Teh, Pascanu,
  De~Freitas, and Gulcehre]{de2024griffin}
S.~De, S.~L. Smith, A.~Fernando, A.~Botev, G.-C. Muraru, A.~Gu, R.~Haroun,
  L.~Berrada, Y.~Chen, S.~Srinivasan, G.~Desjardins, A.~Doucet, D.~Budden,
  Y.~W. Teh, R.~Pascanu, N.~De~Freitas, and C.~Gulcehre.
\newblock Griffin: Mixing gated linear recurrences with local attention for
  efficient language models.
\newblock \emph{arXiv preprint arXiv:2402.19427}, 2024.
\newblock \doi{10.48550/arXiv.2402.19427}.
\newblock URL \url{https://arxiv.org/abs/2402.19427}.

\bibitem[Ding et~al.(2025)Ding, Li, He, Norelli, Wu, Tresp, Bronstein, and
  Ma]{ding2024dygmamba}
Zifeng Ding, Yifeng Li, Yuan He, Antonio Norelli, Jingcheng Wu, Volker Tresp,
  Michael~M. Bronstein, and Yunpu Ma.
\newblock Dygmamba: Efficiently modeling long-term temporal dependency on
  continuous-time dynamic graphs with state space models.
\newblock \emph{Transactions on Machine Learning Research}, 2025, 2025.
\newblock URL \url{https://openreview.net/forum?id=sq5AJvVuha}.
\newblock arXiv:2408.04713.

\bibitem[Dong et~al.(2025)Dong, Fu, Diao, Byeon, Chen, Mahabaleshwarkar, Liu,
  Van~Keirsbilck, Chen, Suhara, Lin, Kautz, and Molchanov]{dong2025hymba}
X.~Dong, Y.~Fu, S.~Diao, W.~Byeon, Z.~Chen, A.~Mahabaleshwarkar, S.-Y. Liu,
  M.~Van~Keirsbilck, M.-H. Chen, Y.~Suhara, Y.~C. Lin, J.~Kautz, and
  P.~Molchanov.
\newblock Hymba: A hybrid-head architecture for small language models.
\newblock In \emph{International Conference on Learning Representations
  (ICLR)}, 2025.
\newblock URL \url{https://openreview.net/forum?id=A1ztozypga}.

\bibitem[Elman(1990)]{elman1990finding}
Jeffrey~L. Elman.
\newblock Finding structure in time.
\newblock \emph{Cognitive Science}, 14\penalty0 (2):\penalty0 179--211, 1990.

\bibitem[Fu et~al.(2023)Fu, Dao, Saab, Thomas, Rudra, and R{\'e}]{fu2022hungry}
Daniel~Y. Fu, Tri Dao, Khaled~K. Saab, Armin~W. Thomas, Atri Rudra, and
  Christopher R{\'e}.
\newblock Hungry hungry hippos: Towards language modeling with state space
  models.
\newblock In \emph{International Conference on Learning Representations
  (ICLR)}, 2023.
\newblock URL \url{https://openreview.net/forum?id=COZDy0WYGg}.

\bibitem[Gao et~al.(2023)Gao, Zhu, Zhang, Wang, and Li]{gao2023dgcn}
Chao Gao, Junyou Zhu, Fan Zhang, Zhen Wang, and Xuelong Li.
\newblock A novel representation learning for dynamic graphs based on graph
  convolutional networks.
\newblock \emph{IEEE Transactions on Cybernetics}, 53\penalty0 (6):\penalty0
  3599--3612, 2023.
\newblock \doi{10.1109/TCYB.2022.3159661}.

\bibitem[Glorioso et~al.(2024)Glorioso, Anthony, Tokpanov, Whittington,
  Pilault, Ibrahim, and Millidge]{glorioso2024zamba}
P.~Glorioso, Q.~Anthony, Y.~Tokpanov, J.~Whittington, J.~Pilault, A.~Ibrahim,
  and B.~Millidge.
\newblock Zamba: A compact 7b {SSM} hybrid model.
\newblock \emph{arXiv preprint arXiv:2405.16712}, 2024.
\newblock \doi{10.48550/arXiv.2405.16712}.
\newblock URL \url{https://arxiv.org/abs/2405.16712}.

\bibitem[Gogianu et~al.(2021)Gogianu, Berariu, Rosca, Clopath, Busoniu, and
  Pascanu]{gogianu2021spectral}
Florin Gogianu, Tudor Berariu, Mihaela~C. Rosca, Claudia Clopath, Lucian
  Busoniu, and Razvan Pascanu.
\newblock Spectral normalisation for deep reinforcement learning: An
  optimisation perspective.
\newblock In \emph{International Conference on Machine Learning}, pages
  3734--3744. PMLR, 2021.

\bibitem[Graves et~al.(2013)Graves, Mohamed, and Hinton]{graves2013speech}
Alex Graves, Abdel-rahman Mohamed, and Geoffrey Hinton.
\newblock Speech recognition with deep recurrent neural networks.
\newblock In \emph{2013 IEEE International Conference on Acoustics, Speech and
  Signal Processing}, pages 6645--6649. IEEE, 2013.
\newblock \doi{10.1109/ICASSP.2013.6638947}.

\bibitem[Gu and Dao(2024)]{gu2023mamba}
Albert Gu and Tri Dao.
\newblock Mamba: Linear-time sequence modeling with selective state spaces.
\newblock In \emph{First Conference on Language Modeling (COLM)}, 2024.
\newblock URL \url{https://openreview.net/forum?id=tEYskw1VY2}.

\bibitem[Gu et~al.(2020)Gu, Dao, Ermon, Rudra, and R{\'e}]{gu2020hippo}
Albert Gu, Tri Dao, Stefano Ermon, Atri Rudra, and Christopher R{\'e}.
\newblock {HiPPO}: Recurrent memory with optimal polynomial projections.
\newblock In \emph{Advances in Neural Information Processing Systems 33
  (NeurIPS 2020)}, volume~33, 2020.
\newblock URL
  \url{https://proceedings.neurips.cc/paper/2020/hash/102f0bb6efb3a6128a3c750dd16729be-Abstract.html}.

\bibitem[Gu et~al.(2021)Gu, Johnson, Goel, Saab, Dao, Rudra, and
  R{\'e}]{gu2021combining}
Albert Gu, Isys Johnson, Karan Goel, Khaled Saab, Tri Dao, Atri Rudra, and
  Christopher R{\'e}.
\newblock Combining recurrent, convolutional, and continuous-time models with
  linear state-space layers.
\newblock In \emph{Advances in Neural Information Processing Systems 34
  (NeurIPS 2021)}, volume~34, pages 572--585, 2021.
\newblock URL
  \url{https://proceedings.neurips.cc/paper/2021/hash/05546b0e38ab9175cd905eebcc6ebb76-Abstract.html}.

\bibitem[Gu et~al.(2022{\natexlab{a}})Gu, Goel, Gupta, and
  R{\'e}]{gu2022parameterization}
Albert Gu, Karan Goel, Ankit Gupta, and Christopher R{\'e}.
\newblock On the parameterization and initialization of diagonal state space
  models.
\newblock In \emph{Advances in Neural Information Processing Systems 35
  (NeurIPS 2022)}, volume~35, 2022{\natexlab{a}}.
\newblock URL
  \url{https://papers.nips.cc/paper_files/paper/2022/hash/e9a32fade47b906de908431991440f7c-Abstract-Conference.html}.

\bibitem[Gu et~al.(2022{\natexlab{b}})Gu, Goel, and R{\'e}]{gu2021efficiently}
Albert Gu, Karan Goel, and Christopher R{\'e}.
\newblock Efficiently modeling long sequences with structured state spaces.
\newblock In \emph{International Conference on Learning Representations
  (ICLR)}, 2022{\natexlab{b}}.
\newblock URL \url{https://openreview.net/forum?id=uYLFoz1vlAC}.

\bibitem[Gu et~al.(2023)Gu, Johnson, Timalsina, Rudra, and R{\'e}]{gu2022train}
Albert Gu, Isys Johnson, Aman Timalsina, Atri Rudra, and Christopher R{\'e}.
\newblock How to train your {HiPPO}: State space models with generalized
  orthogonal basis projections.
\newblock In \emph{International Conference on Learning Representations
  (ICLR)}, 2023.
\newblock URL \url{https://openreview.net/forum?id=klK17OQ3KB}.

\bibitem[Gupta et~al.(2022)Gupta, Gu, and Berant]{gupta2022diagonal}
Ankit Gupta, Albert Gu, and Jonathan Berant.
\newblock Diagonal state spaces are as effective as structured state spaces.
\newblock In \emph{Advances in Neural Information Processing Systems 35
  (NeurIPS 2022)}, volume~35, 2022.
\newblock URL
  \url{https://papers.nips.cc/paper_files/paper/2022/hash/9156b0f6dfa9bbd18c79cc459ef5d61c-Abstract-Conference.html}.

\bibitem[Hamilton(1994)]{hamilton2020time}
James~D. Hamilton.
\newblock \emph{Time Series Analysis}.
\newblock Princeton University Press, 1994.
\newblock ISBN 9780691042893.

\bibitem[Hangos et~al.(2006)Hangos, Bokor, and
  Szederk{\'e}nyi]{hangos2006analysis}
Katalin~M. Hangos, J{\'o}zsef Bokor, and G{\'a}bor Szederk{\'e}nyi.
\newblock \emph{Analysis and control of nonlinear process systems}.
\newblock Springer Science \& Business Media, 2006.

\bibitem[Hasani et~al.(2023)Hasani, Lechner, Wang, Chahine, Amini, and
  Rus]{hasani2022liquid}
Ramin Hasani, Mathias Lechner, Tsun-Hsuan Wang, Makram Chahine, Alexander
  Amini, and Daniela Rus.
\newblock Liquid structural state-space models.
\newblock In \emph{International Conference on Learning Representations
  (ICLR)}, 2023.
\newblock URL \url{https://openreview.net/forum?id=g4OTKRKfS7R}.

\bibitem[He et~al.(2024)He, Bai, Zhang, He, Chen, Gan, Wang, Li, Tian, and
  Xie]{he2024mambaad}
Haoyang He, Yuhu Bai, Jiangning Zhang, Qingdong He, Hongxu Chen, Zhenye Gan,
  Chengjie Wang, Xiangtai Li, Guanzhong Tian, and Lei Xie.
\newblock {MambaAD}: Exploring state space models for multi-class unsupervised
  anomaly detection.
\newblock In \emph{Advances in Neural Information Processing Systems},
  volume~37, pages 71162--71187, 2024.

\bibitem[Heidari et~al.(2024)Heidari, Ghorbani~Kolahi, Karimijafarbigloo, Azad,
  Bozorgpour, Hatami, Azad, Diba, Bagci, Merhof, and
  Hacihaliloglu]{heidari2024computation}
Moein Heidari, Sina Ghorbani~Kolahi, Sanaz Karimijafarbigloo, Bobby Azad,
  Afshin Bozorgpour, Soheila Hatami, Reza Azad, Ali Diba, Ulas Bagci, Dorit
  Merhof, and Ilker Hacihaliloglu.
\newblock Computation-efficient era: A comprehensive survey of state space
  models in medical image analysis, 2024.

\bibitem[Hochreiter and Schmidhuber(1997)]{hochreiter1997long}
Sepp Hochreiter and J{\"u}rgen Schmidhuber.
\newblock Long short-term memory.
\newblock \emph{Neural Computation}, 9\penalty0 (8):\penalty0 1735--1780, 1997.

\bibitem[Huang et~al.(2019)Huang, Cheng, Bapna, Firat, Chen, Chen, Lee, Ngiam,
  Le, Wu, et~al.]{huang2019gpipe}
Yanping Huang, Youlong Cheng, Ankur Bapna, Orhan Firat, Dehao Chen, Mia Chen,
  HyoukJoong Lee, Jiquan Ngiam, Quoc~V. Le, Yonghui Wu, et~al.
\newblock Gpipe: Efficient training of giant neural networks using pipeline
  parallelism.
\newblock In \emph{Advances in Neural Information Processing Systems},
  volume~32, 2019.

\bibitem[{Jamba Team}(2024)]{team2024jamba}
{Jamba Team}.
\newblock Jamba-1.5: Hybrid transformer-mamba models at scale.
\newblock \emph{arXiv preprint arXiv:2408.12570}, 2024.

\bibitem[Jordan(1997)]{jordan1997serial}
Michael~I. Jordan.
\newblock Serial order: A parallel distributed processing approach.
\newblock In \emph{Advances in Psychology}, volume 121, pages 471--495.
  Elsevier, 1997.

\bibitem[Jumper et~al.(2021)Jumper, Evans, Pritzel, Green, Figurnov,
  Ronneberger, Tunyasuvunakool, Bates, {\v{Z}}{\'\i}dek, Potapenko,
  et~al.]{jumper2021highly}
John Jumper, Richard Evans, Alexander Pritzel, Tim Green, Michael Figurnov,
  Olaf Ronneberger, Kathryn Tunyasuvunakool, Russ Bates, Augustin
  {\v{Z}}{\'\i}dek, Anna Potapenko, et~al.
\newblock Highly accurate protein structure prediction with alphafold.
\newblock \emph{Nature}, 596\penalty0 (7873):\penalty0 583--589, 2021.

\bibitem[Kailath(1980)]{kailath1980linear}
Thomas Kailath.
\newblock \emph{Linear Systems}, volume 156.
\newblock Prentice-Hall, Englewood Cliffs, NJ, 1980.

\bibitem[Kalman(1960)]{kalman1960new}
Rudolph~Emil Kalman.
\newblock A new approach to linear filtering and prediction problems.
\newblock \emph{Transactions of the ASME--Journal of Basic Engineering},
  82\penalty0 (1):\penalty0 35--45, 1960.

\bibitem[Katharopoulos et~al.(2020)Katharopoulos, Vyas, Pappas, and
  Fleuret]{katharopoulos2020transformers}
Angelos Katharopoulos, Apoorv Vyas, Nikolaos Pappas, and Fran{\c{c}}ois
  Fleuret.
\newblock Transformers are rnns: Fast autoregressive transformers with linear
  attention.
\newblock In \emph{International Conference on Machine Learning}, pages
  5156--5165. PMLR, 2020.

\bibitem[LeCun et~al.(1989)LeCun, Boser, Denker, Henderson, Howard, Hubbard,
  and Jackel]{lecun1989backpropagation}
Yann LeCun, Bernhard Boser, John~S. Denker, Donnie Henderson, Richard~E.
  Howard, Wayne Hubbard, and Lawrence~D. Jackel.
\newblock Backpropagation applied to handwritten zip code recognition.
\newblock \emph{Neural Computation}, 1\penalty0 (4):\penalty0 541--551, 1989.

\bibitem[LeCun et~al.(1998)LeCun, Bottou, Bengio, and
  Haffner]{lecun1998gradient}
Yann LeCun, L{\'e}on Bottou, Yoshua Bengio, and Patrick Haffner.
\newblock Gradient-based learning applied to document recognition.
\newblock \emph{Proceedings of the IEEE}, 86\penalty0 (11):\penalty0
  2278--2324, 1998.

\bibitem[Lenz et~al.(2025)Lenz, Lieber, Arazi, Bergman, Manevich, Peleg,
  Aviram, Almagor, Fridman, Padnos, Gissin, Jannai, Muhlgay, Zimberg, Gerber,
  Dolev, Krakovsky, Safahi, Schwartz, Cohen, Shachaf, Rozenblum, Bata, Blass,
  Magar, Dalmedigos, Osin, Fadlon, Rozman, Danos, Gokhman, Zusman, Gidron,
  Ratner, Gat, Rozen, Fried, Leshno, Antverg, Abend, Dagan, Cohavi, Alon,
  Belson, Cohen, Gilad, Glozman, Lev, Shalev-Shwartz, Meirom, Delbari, Ness,
  Asida, Gal, Braude, Pumerantz, Cohen, Belinkov, Globerson, Levy, and
  Shoham]{lieber2024jamba}
Barak Lenz, Opher Lieber, Alan Arazi, Amir Bergman, Avshalom Manevich, Barak
  Peleg, Ben Aviram, Chen Almagor, Clara Fridman, Dan Padnos, Daniel Gissin,
  Daniel Jannai, Dor Muhlgay, Dor Zimberg, Edden~M. Gerber, Elad Dolev, Eran
  Krakovsky, Erez Safahi, Erez Schwartz, Gal Cohen, Gal Shachaf, Haim
  Rozenblum, Hofit Bata, Ido Blass, Inbal Magar, Itay Dalmedigos, Jhonathan
  Osin, Julie Fadlon, Maria Rozman, Matan Danos, Michael Gokhman, Mor Zusman,
  Naama Gidron, Nir Ratner, Noam Gat, Noam Rozen, Oded Fried, Ohad Leshno, Omer
  Antverg, Omri Abend, Or~Dagan, Orit Cohavi, Raz Alon, Ro'i Belson, Roi Cohen,
  Rom Gilad, Roman Glozman, Shahar Lev, Shai Shalev-Shwartz, Shaked~Haim
  Meirom, Tal Delbari, Tal Ness, Tomer Asida, Tom~Ben Gal, Tom Braude, Uriya
  Pumerantz, Josh Cohen, Yonatan Belinkov, Yuval Globerson, Yuval~Peleg Levy,
  and Yoav Shoham.
\newblock Jamba: Hybrid transformer-mamba language models.
\newblock In \emph{The Thirteenth International Conference on Learning
  Representations (ICLR)}, 2025.
\newblock URL \url{https://openreview.net/forum?id=JFPaD7lpBD}.
\newblock Preprint: arXiv:2403.19887.

\bibitem[Li et~al.(2024)Li, Zhou, Yu, Song, and Yang]{li2024coupled}
Wenbing Li, Hang Zhou, Junqing Yu, Zikai Song, and Wei Yang.
\newblock Coupled mamba: Enhanced multimodal fusion with coupled state space
  model.
\newblock In \emph{Advances in Neural Information Processing Systems 37
  (NeurIPS 2024)}, pages 59808--59832, 2024.
\newblock \doi{10.52202/079017-1910}.
\newblock arXiv:2405.18014.

\bibitem[Li et~al.(2025{\natexlab{a}})Li, Qiao, Yin, Wu, Gao, Wang, and
  Li]{li2025mfndsurvey}
Xianghua Li, Jiao Qiao, Shu Yin, Lianwei Wu, Chao Gao, Zhen Wang, and Xuelong
  Li.
\newblock A survey of multimodal fake news detection: A cross-modal interaction
  perspective.
\newblock \emph{IEEE Transactions on Emerging Topics in Computational
  Intelligence}, 9\penalty0 (4):\penalty0 2658--2675, 2025{\natexlab{a}}.
\newblock \doi{10.1109/TETCI.2025.3543389}.

\bibitem[Li et~al.(2025{\natexlab{b}})Li, Xia, Chang, and Wu]{li2024survey}
Zhiyuan Li, Tingyu Xia, Yi~Chang, and Yuan Wu.
\newblock A survey of {RWKV}.
\newblock \emph{Neurocomputing}, 649:\penalty0 130711, 2025{\natexlab{b}}.
\newblock ISSN 0925-2312.
\newblock \doi{10.1016/j.neucom.2025.130711}.

\bibitem[Lim et~al.(2021)Lim, Ar{\i}k, Loeff, and Pfister]{lim2021temporal}
Bryan Lim, Sercan~{\"O} Ar{\i}k, Nicolas Loeff, and Tomas Pfister.
\newblock Temporal fusion transformers for interpretable multi-horizon time
  series forecasting.
\newblock \emph{International Journal of Forecasting}, 37\penalty0
  (4):\penalty0 1748--1764, 2021.

\bibitem[Ljung(1987)]{ljung1987theory}
Lennart Ljung.
\newblock \emph{System Identification: Theory for the User}.
\newblock Prentice-Hall, 1987.

\bibitem[Long et~al.(2024)Long, Zhou, Li, Lu, Ying, Luo, Ma, and
  Yan]{long2024dgmamba}
Shaocong Long, Qianyu Zhou, Xiangtai Li, Xuequan Lu, Chenhao Ying, Yuan Luo,
  Lizhuang Ma, and Shuicheng Yan.
\newblock {DGMamba}: Domain generalization via generalized state space model.
\newblock In \emph{Proceedings of the 32nd ACM International Conference on
  Multimedia}, pages 3607--3616. Association for Computing Machinery, 2024.
\newblock \doi{10.1145/3664647.3681247}.

\bibitem[Ma et~al.(2024)Ma, Li, and Wang]{ma2024umamba}
Jun Ma, Feifei Li, and Bo~Wang.
\newblock U-mamba: Enhancing long-range dependency for biomedical image
  segmentation.
\newblock \emph{arXiv preprint arXiv:2401.04722}, 2024.

\bibitem[Ma et~al.(2023)Ma, Zhou, Kong, He, Gui, Neubig, May, and
  Zettlemoyer]{ma2022mega}
Xuezhe Ma, Chunting Zhou, Xiang Kong, Junxian He, Liangke Gui, Graham Neubig,
  Jonathan May, and Luke Zettlemoyer.
\newblock {Mega}: Moving average equipped gated attention.
\newblock In \emph{International Conference on Learning Representations
  (ICLR)}, 2023.
\newblock URL \url{https://openreview.net/forum?id=qNLe3iq2El}.

\bibitem[Mehta et~al.(2023)Mehta, Gupta, Cutkosky, and
  Neyshabur]{mehta2022long}
Harsh Mehta, Ankit Gupta, Ashok Cutkosky, and Behnam Neyshabur.
\newblock Long range language modeling via gated state spaces.
\newblock In \emph{International Conference on Learning Representations
  (ICLR)}, 2023.
\newblock URL \url{https://openreview.net/forum?id=5MkYIYCbva}.

\bibitem[Miyato et~al.(2018)Miyato, Kataoka, Koyama, and
  Yoshida]{miyato2018spectral}
Takeru Miyato, Toshiki Kataoka, Masanori Koyama, and Yuichi Yoshida.
\newblock Spectral normalization for generative adversarial networks.
\newblock In \emph{International Conference on Learning Representations}, 2018.
\newblock URL \url{https://openreview.net/forum?id=B1QRgziT-}.
\newblock arXiv:1802.05957.

\bibitem[Miyazaki et~al.(2024)Miyazaki, Masuyama, and
  Murata]{miyazaki2024exploring}
Koichi Miyazaki, Yoshiki Masuyama, and Masato Murata.
\newblock Exploring the capability of mamba in speech applications.
\newblock In \emph{Interspeech 2024}, pages 237--241, 2024.
\newblock \doi{10.21437/Interspeech.2024-994}.
\newblock arXiv:2406.16808.

\bibitem[Nguyen et~al.(2022)Nguyen, Goel, Gu, Downs, Shah, Dao, Baccus, and
  R{\'e}]{nguyen2022s4nd}
Eric Nguyen, Karan Goel, Albert Gu, Gordon~W. Downs, Preey Shah, Tri Dao,
  Stephen Baccus, and Christopher R{\'e}.
\newblock {S4ND}: Modeling images and videos as multidimensional signals with
  state spaces.
\newblock In \emph{Advances in Neural Information Processing Systems 35
  (NeurIPS 2022)}, volume~35, 2022.
\newblock URL
  \url{https://papers.nips.cc/paper_files/paper/2022/hash/13388efc819c09564c66ab2dc8463809-Abstract-Conference.html}.

\bibitem[Orvieto et~al.(2023)Orvieto, Smith, Gu, Fernando, Gulcehre, Pascanu,
  and De]{orvieto2023resurrecting}
Antonio Orvieto, Samuel~L. Smith, Albert Gu, Anushan Fernando, Caglar Gulcehre,
  Razvan Pascanu, and Soham De.
\newblock Resurrecting recurrent neural networks for long sequences.
\newblock In \emph{Proceedings of the 40th International Conference on Machine
  Learning (ICML)}, volume 202 of \emph{Proceedings of Machine Learning
  Research}, pages 26670--26698. PMLR, 2023.
\newblock URL \url{https://proceedings.mlr.press/v202/orvieto23a.html}.

\bibitem[Patro and Agneeswaran(2024)]{patro2024simba}
Badri~N. Patro and Vijay~S. Agneeswaran.
\newblock Simba: Simplified mamba-based architecture for vision and
  multivariate time series, 2024.

\bibitem[Patro and Agneeswaran(2025)]{patro2024mamba}
Badri~Narayana Patro and Vijay~Srinivas Agneeswaran.
\newblock Mamba-360: Survey of state space models as transformer alternative
  for long sequence modelling: Methods, applications, and challenges.
\newblock \emph{Engineering Applications of Artificial Intelligence},
  159:\penalty0 111279, 2025.
\newblock ISSN 0952-1976.
\newblock \doi{10.1016/j.engappai.2025.111279}.

\bibitem[Peng et~al.(2023)Peng, Alcaide, Anthony, Albalak, Arcadinho, Biderman,
  Cao, Cheng, Chung, Derczynski, Du, Grella, Gv, He, Hou, Kazienko, Kocon,
  Kong, Koptyra, Lau, Lin, Mantri, Mom, Saito, Song, Tang, Wind, Wo{\'z}niak,
  Zhang, Zhou, Zhu, and Zhu]{peng2023rwkv}
Bo~Peng, Eric Alcaide, Quentin Anthony, Alon Albalak, Samuel Arcadinho, Stella
  Biderman, Huanqi Cao, Xin Cheng, Michael Chung, Leon Derczynski, Xingjian Du,
  Matteo Grella, Kranthi Gv, Xuzheng He, Haowen Hou, Przemyslaw Kazienko, Jan
  Kocon, Jiaming Kong, Bart{\l}omiej Koptyra, Hayden Lau, Jiaju Lin, Krishna
  Sri~Ipsit Mantri, Ferdinand Mom, Atsushi Saito, Guangyu Song, Xiangru Tang,
  Johan Wind, Stanis{\l}aw Wo{\'z}niak, Zhenyuan Zhang, Qinghua Zhou, Jian Zhu,
  and Rui-Jie Zhu.
\newblock {RWKV}: Reinventing {RNN}s for the transformer era.
\newblock In \emph{Findings of the Association for Computational Linguistics:
  EMNLP 2023}, pages 14048--14077, Singapore, December 2023. Association for
  Computational Linguistics.
\newblock \doi{10.18653/v1/2023.findings-emnlp.936}.
\newblock URL \url{https://aclanthology.org/2023.findings-emnlp.936/}.

\bibitem[Poli et~al.(2023)Poli, Massaroli, Nguyen, Fu, Dao, Baccus, Bengio,
  Ermon, and R{\'e}]{poli2023hyena}
Michael Poli, Stefano Massaroli, Eric Nguyen, Daniel~Y. Fu, Tri Dao, Stephen
  Baccus, Yoshua Bengio, Stefano Ermon, and Christopher R{\'e}.
\newblock Hyena hierarchy: Towards larger convolutional language models.
\newblock In \emph{Proceedings of the 40th International Conference on Machine
  Learning (ICML)}, volume 202 of \emph{Proceedings of Machine Learning
  Research}, pages 28043--28078. PMLR, 2023.
\newblock URL \url{https://proceedings.mlr.press/v202/poli23a.html}.

\bibitem[Qiao et~al.(2024)Qiao, Yu, Zhao, Chen, Sun, Guo, Wu, and
  Liu]{qiao2024vl}
Yanyuan Qiao, Zheng Yu, Zijia Zhao, Sihan Chen, Mingzhen Sun, Longteng Guo,
  Qi~Wu, and Jing Liu.
\newblock {VL-Mamba}: Exploring state space models for multimodal learning.
\newblock In \emph{Proceedings of The 4th NeurIPS Efficient Natural Language
  and Speech Processing Workshop}, volume 262 of \emph{Proceedings of Machine
  Learning Research}, pages 102--113. PMLR, 2024.
\newblock URL \url{https://proceedings.mlr.press/v262/qiao24a.html}.
\newblock arXiv:2403.13600.

\bibitem[Qin et~al.(2023{\natexlab{a}})Qin, Han, Sun, He, Li, Li, Dai, Kong,
  and Zhong]{qin2023toeplitz}
Zhen Qin, Xiaodong Han, Weixuan Sun, Bowen He, Dong Li, Dongxu Li, Yuchao Dai,
  Lingpeng Kong, and Yiran Zhong.
\newblock Toeplitz neural network for sequence modeling.
\newblock In \emph{International Conference on Learning Representations
  (ICLR)}, 2023{\natexlab{a}}.
\newblock URL \url{https://openreview.net/forum?id=IxmWsm4xrua}.

\bibitem[Qin et~al.(2023{\natexlab{b}})Qin, Yang, and Zhong]{qin2023hgrn}
Zhen Qin, Songlin Yang, and Yiran Zhong.
\newblock Hierarchically gated recurrent neural network for sequence modeling.
\newblock In \emph{Advances in Neural Information Processing Systems 36
  (NeurIPS 2023)}, volume~36, 2023{\natexlab{b}}.
\newblock URL
  \url{https://papers.nips.cc/paper_files/paper/2023/hash/694be3548697e9cc8999d45e8d16fe1e-Abstract-Conference.html}.
\newblock Spotlight.

\bibitem[Qin et~al.(2024)Qin, Yang, Sun, Shen, Li, Sun, and
  Zhong]{qin2024hgrn2}
Zhen Qin, Songlin Yang, Weixuan Sun, Xuyang Shen, Dong Li, Weigao Sun, and
  Yiran Zhong.
\newblock {HGRN2}: Gated linear {RNNs} with state expansion.
\newblock In \emph{Proceedings of the Conference on Language Modeling (COLM)},
  2024.
\newblock URL \url{https://arxiv.org/abs/2404.07904}.

\bibitem[Qu et~al.(2026)Qu, Ning, An, Fan, Derr, Liu, Xu, and Li]{qu2024survey}
Haohao Qu, Liangbo Ning, Rui An, Wenqi Fan, Tyler Derr, Hui Liu, Xin Xu, and
  Qing Li.
\newblock A survey of mamba.
\newblock \emph{ACM Transactions on Intelligent Systems and Technology}, 2026.
\newblock \doi{10.1145/3819582}.
\newblock arXiv:2408.01129.

\bibitem[Radford et~al.(2023)Radford, Kim, Xu, Brockman, McLeavey, and
  Sutskever]{radford2023robust}
Alec Radford, Jong~Wook Kim, Tao Xu, Greg Brockman, Christine McLeavey, and
  Ilya Sutskever.
\newblock Robust speech recognition via large-scale weak supervision.
\newblock In \emph{International Conference on Machine Learning}, pages
  28492--28518. PMLR, 2023.

\bibitem[Raffel et~al.(2020)Raffel, Shazeer, Roberts, Lee, Narang, Matena,
  Zhou, Li, and Liu]{raffel2020exploring}
Colin Raffel, Noam Shazeer, Adam Roberts, Katherine Lee, Sharan Narang, Michael
  Matena, Yanqi Zhou, Wei Li, and Peter~J. Liu.
\newblock Exploring the limits of transfer learning with a unified text-to-text
  transformer.
\newblock \emph{Journal of Machine Learning Research}, 21\penalty0
  (140):\penalty0 1--67, 2020.

\bibitem[Ren et~al.(2025)Ren, Liu, Lu, Shen, Liang, and Chen]{ren2025samba}
L.~Ren, Y.~Liu, Y.~Lu, Y.~Shen, C.~Liang, and W.~Chen.
\newblock Samba: Simple hybrid state space models for efficient unlimited
  context language modeling.
\newblock In \emph{International Conference on Learning Representations
  (ICLR)}, 2025.
\newblock URL \url{https://openreview.net/forum?id=bIlnpVM4bc}.

\bibitem[Salam et~al.(2025)Salam, Mahmud, Islam, Mukta, and
  Shatabda]{salam2025comprehensivemamba}
Abdus Salam, Rasel Mahmud, Tohedul Islam, Saddam Mukta, and Swakkhar Shatabda.
\newblock A comprehensive survey on {Mamba}: Architectures, challenges, and
  opportunities.
\newblock \emph{Computer}, 58\penalty0 (8):\penalty0 64--76, 2025.
\newblock \doi{10.1109/MC.2025.3571322}.
\newblock No arXiv preprint of this article exists.

\bibitem[Shoeybi et~al.(2019)Shoeybi, Patwary, Puri, LeGresley, Casper, and
  Catanzaro]{shoeybi2019megatron}
Mohammad Shoeybi, Mostofa Patwary, Raul Puri, Patrick LeGresley, Jared Casper,
  and Bryan Catanzaro.
\newblock Megatron-lm: Training multi-billion parameter language models using
  model parallelism.
\newblock \emph{arXiv preprint arXiv:1909.08053}, 2019.

\bibitem[Smith et~al.(2023{\natexlab{a}})Smith, Mello, Kautz, Linderman, and
  Byeon]{smith2023convolutional}
Jimmy Smith, Shalini~De Mello, Jan Kautz, Scott Linderman, and Wonmin Byeon.
\newblock Convolutional state space models for long-range spatiotemporal
  modeling.
\newblock \emph{Advances in Neural Information Processing Systems},
  36:\penalty0 80690--80729, 2023{\natexlab{a}}.

\bibitem[Smith et~al.(2023{\natexlab{b}})Smith, Warrington, and
  Linderman]{smith2022simplified}
Jimmy T.~H. Smith, Andrew Warrington, and Scott~W. Linderman.
\newblock Simplified state space layers for sequence modeling.
\newblock In \emph{International Conference on Learning Representations
  (ICLR)}, 2023{\natexlab{b}}.
\newblock URL \url{https://openreview.net/forum?id=Ai8Hw3AXqks}.

\bibitem[Somvanshi et~al.(2025{\natexlab{a}})Somvanshi, Chakraborty, Das, and
  Dutta]{Somvanshi2025}
Shriyank Somvanshi, Rohit Chakraborty, Subasish Das, and Anandi~K. Dutta.
\newblock Crash severity analysis of child bicyclists using {Arm-Net} and
  {MambaNet}.
\newblock In \emph{2025 IEEE Conference on Artificial Intelligence (CAI)},
  pages 821--824. IEEE, 2025{\natexlab{a}}.
\newblock \doi{10.1109/CAI64502.2025.00146}.

\bibitem[Somvanshi et~al.(2025{\natexlab{b}})Somvanshi, Tusti, Chakraborty, and
  Das]{Somvanshi2025YM}
Shriyank Somvanshi, Anannya~Ghosh Tusti, Rohit Chakraborty, and Subasish Das.
\newblock Applying tabular deep learning models to estimate crash injury types
  of young motorcyclists.
\newblock In \emph{2025 IEEE Conference on Artificial Intelligence (CAI)},
  pages 1010--1015. IEEE, 2025{\natexlab{b}}.
\newblock \doi{10.1109/CAI64502.2025.00177}.

\bibitem[Sun et~al.(2023)Sun, Dong, Huang, Ma, Xia, Xue, Wang, and
  Wei]{sun2023retnet}
Yutao Sun, Li~Dong, Shaohan Huang, Shuming Ma, Yuqing Xia, Jilong Xue, Jianyong
  Wang, and Furu Wei.
\newblock Retentive network: A successor to transformer for large language
  models, 2023.

\bibitem[Tang et~al.(2023)Tang, Dunnmon, Liangqiong, Saab, Baykaner,
  Lee-Messer, and Rubin]{tang2023modeling}
Siyi Tang, Jared~A. Dunnmon, Qu~Liangqiong, Khaled~K. Saab, Tina Baykaner,
  Christopher Lee-Messer, and Daniel~L. Rubin.
\newblock Modeling multivariate biosignals with graph neural networks and
  structured state space models.
\newblock In \emph{Conference on Health, Inference, and Learning}, pages
  50--71. PMLR, 2023.

\bibitem[Tay et~al.(2021)Tay, Dehghani, Abnar, Shen, Bahri, Pham, Rao, Yang,
  Ruder, and Metzler]{tay2020long}
Yi~Tay, Mostafa Dehghani, Samira Abnar, Yikang Shen, Dara Bahri, Philip Pham,
  Jinfeng Rao, Liu Yang, Sebastian Ruder, and Donald Metzler.
\newblock Long range arena: A benchmark for efficient transformers.
\newblock In \emph{International Conference on Learning Representations
  (ICLR)}, 2021.
\newblock URL \url{https://openreview.net/forum?id=qVyeW-grC2k}.

\bibitem[Tay et~al.(2022)Tay, Dehghani, Bahri, and Metzler]{tay2022efficient}
Yi~Tay, Mostafa Dehghani, Dara Bahri, and Donald Metzler.
\newblock Efficient transformers: A survey.
\newblock \emph{ACM Computing Surveys}, 55\penalty0 (6):\penalty0 1--28, 2022.

\bibitem[Teng et~al.(2026)Teng, Gao, Li, Wang, Fan, and
  Nekorkin]{teng2026multiscale}
Min Teng, Chao Gao, Xianghua Li, Zhen Wang, Kefeng Fan, and Vladimir Nekorkin.
\newblock Multi-scale graph contrastive learning for community detection in
  dynamic graphs.
\newblock \emph{Information Processing \& Management}, 63\penalty0
  (2):\penalty0 104410, 2026.
\newblock \doi{10.1016/j.ipm.2025.104410}.

\bibitem[Vaswani et~al.(2017)Vaswani, Shazeer, Parmar, Uszkoreit, Jones, Gomez,
  Kaiser, and Polosukhin]{vaswani2017attention}
Ashish Vaswani, Noam Shazeer, Niki Parmar, Jakob Uszkoreit, Llion Jones,
  Aidan~N. Gomez, {\L}ukasz Kaiser, and Illia Polosukhin.
\newblock Attention is all you need.
\newblock In \emph{Advances in Neural Information Processing Systems},
  volume~30, 2017.

\bibitem[Vats et~al.(2025)Vats, Raja, Mathur, Chadha, and
  Jain]{vats2025multilingual}
Arpita Vats, Rahul Raja, Mrinal Mathur, Aman Chadha, and Vinija Jain.
\newblock Multilingual state space models for structured question answering in
  {I}ndic languages.
\newblock In \emph{Proceedings of the Eighth Workshop on Technologies for
  Machine Translation of Low-Resource Languages (LoResMT 2025)}, pages
  115--128, Albuquerque, New Mexico, U.S.A., May 2025. Association for
  Computational Linguistics.
\newblock \doi{10.18653/v1/2025.loresmt-1.11}.
\newblock URL \url{https://aclanthology.org/2025.loresmt-1.11/}.
\newblock arXiv:2502.01673.

\bibitem[Wang et~al.(2024{\natexlab{a}})Wang, Tsepa, Ma, and
  Wang]{wang2024graph}
Chloe Wang, Oleksii Tsepa, Jun Ma, and Bo~Wang.
\newblock Graph-mamba: Towards long-range graph sequence modeling with
  selective state spaces.
\newblock \emph{arXiv preprint arXiv:2402.00789}, 2024{\natexlab{a}}.

\bibitem[Wang et~al.(2023)Wang, Yan, Gu, and Rush]{wang2023bigs}
Junxiong Wang, Jing~Nathan Yan, Albert Gu, and Alexander~M. Rush.
\newblock Pretraining without attention.
\newblock In \emph{Findings of the Association for Computational Linguistics:
  EMNLP 2023}, pages 58--69, 2023.
\newblock \doi{10.18653/v1/2023.findings-emnlp.5}.

\bibitem[Wang and Li(2024)]{wang2023stablessm}
Shida Wang and Qianxiao Li.
\newblock {S}table{SSM}: Alleviating the curse of memory in state-space models
  through stable reparameterization.
\newblock In Ruslan Salakhutdinov, Zico Kolter, Katherine Heller, Adrian
  Weller, Nuria Oliver, Jonathan Scarlett, and Felix Berkenkamp, editors,
  \emph{Proceedings of the 41st International Conference on Machine Learning},
  volume 235 of \emph{Proceedings of Machine Learning Research}, pages
  50766--50793. PMLR, 2024.
\newblock URL \url{https://proceedings.mlr.press/v235/wang24ag.html}.
\newblock arXiv:2311.14495.

\bibitem[Wang et~al.(2020)Wang, Li, Khabsa, Fang, and Ma]{wang2020linformer}
Sinong Wang, Belinda~Z. Li, Madian Khabsa, Han Fang, and Hao Ma.
\newblock Linformer: Self-attention with linear complexity, 2020.

\bibitem[Wang et~al.(2024{\natexlab{b}})Wang, Wang, Ding, Li, Wu, Rong, Kong,
  Huang, Li, Yang, Wang, Jiang, Li, Wang, Tian, and Tang]{wang2024state}
Xiao Wang, Shiao Wang, Yuhe Ding, Yuehang Li, Wentao Wu, Yao Rong, Weizhe Kong,
  Ju~Huang, Shihao Li, Haoxiang Yang, Ziwen Wang, Bo~Jiang, Chenglong Li,
  Yaowei Wang, Yonghong Tian, and Jin Tang.
\newblock State space model for new-generation network alternative to
  transformers: A survey.
\newblock \emph{arXiv preprint arXiv:2404.09516}, 2024{\natexlab{b}}.

\bibitem[Xie et~al.(2024)Xie, Cui, Tan, Zheng, and Yu]{xie2024fusionmamba}
Xinyu Xie, Yawen Cui, Tao Tan, Xubin Zheng, and Zitong Yu.
\newblock {FusionMamba}: dynamic feature enhancement for multimodal image
  fusion with {Mamba}.
\newblock \emph{Visual Intelligence}, 2\penalty0 (1):\penalty0 37, 2024.
\newblock \doi{10.1007/s44267-024-00072-9}.

\bibitem[Xu et~al.(2024)Xu, Yang, Wang, Cai, Du, and Chen]{xu2024visualmamba}
Rui Xu, Shu Yang, Yihui Wang, Yu~Cai, Bo~Du, and Hao Chen.
\newblock Visual mamba: A survey and new outlooks.
\newblock \emph{arXiv preprint arXiv:2404.18861}, 2024.
\newblock \doi{10.48550/arXiv.2404.18861}.

\bibitem[Yue et~al.(2024)Yue, Wang, He, Zeng, McAuley, and Wang]{yue2024linear}
Zhenrui Yue, Yueqi Wang, Zhankui He, Huimin Zeng, Julian McAuley, and Dong
  Wang.
\newblock Linear recurrent units for sequential recommendation.
\newblock In \emph{Proceedings of the 17th ACM International Conference on Web
  Search and Data Mining (WSDM)}, pages 930--938. ACM, 2024.
\newblock \doi{10.1145/3616855.3635760}.

\bibitem[Zhang et~al.(2024{\natexlab{a}})Zhang, Zhu, Wang, Zhang, Chen, and
  Ye]{zhang2024surveyvisualmamba}
Hanwei Zhang, Ying Zhu, Dan Wang, Lijun Zhang, Tianxiang Chen, and Zi~Ye.
\newblock A survey on visual mamba.
\newblock \emph{arXiv preprint arXiv:2404.15956}, 2024{\natexlab{a}}.
\newblock \doi{10.48550/arXiv.2404.15956}.

\bibitem[Zhang et~al.(2025)Zhang, Yuan, Qi, Zhang, Zhou, Ji, Yan, and
  Li]{zhang2025pointcloudmamba}
Tao Zhang, Haobo Yuan, Lu~Qi, Jiangning Zhang, Qianyu Zhou, Shunping Ji,
  Shuicheng Yan, and Xiangtai Li.
\newblock Point cloud mamba: Point cloud learning via state space model.
\newblock In \emph{Proceedings of the AAAI Conference on Artificial
  Intelligence}, volume~39, pages 10121--10130, 2025.
\newblock \doi{10.1609/aaai.v39i10.33098}.

\bibitem[Zhang et~al.(2024{\natexlab{b}})Zhang, Sun, Chen, Pfister, Zhang, and
  Arik]{zhang2024chain}
Yusen Zhang, Ruoxi Sun, Yanfei Chen, Tomas Pfister, Rui Zhang, and Sercan Arik.
\newblock Chain of agents: Large language models collaborating on long-context
  tasks.
\newblock \emph{Advances in Neural Information Processing Systems},
  37:\penalty0 132208--132237, 2024{\natexlab{b}}.

\bibitem[Zuo et~al.(2022)Zuo, Liu, Jiao, Charles, Manavoglu, Zhao, and
  Gao]{zuo2022efficient}
Simiao Zuo, Xiaodong Liu, Jian Jiao, Denis Charles, Eren Manavoglu, Tuo Zhao,
  and Jianfeng Gao.
\newblock Efficient long sequence modeling via state space augmented
  transformer, 2022.

\end{thebibliography}


\vskip3pt









\vskip3pt

\bio{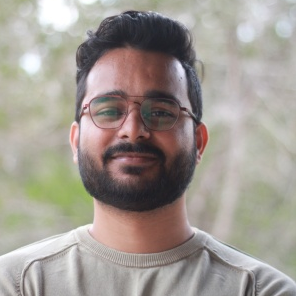}
\textbf{Shriyank Somvanshi} is a doctoral researcher in the Ingram School
of Engineering at Texas State University. He received the M.S. degree in
Civil Engineering from Texas State University and the bachelor's degree
in Civil Engineering from Dr. A.P.J. Abdul Kalam Technical University,
India. His research interests include transportation safety, automated
vehicle crash analysis, interpretable tabular deep learning, explainable
artificial intelligence, structured state-space models, and data-driven
decision-support methods for crash severity and pedestrian risk assessment.
\endbio

\vspace{2em}

\bio{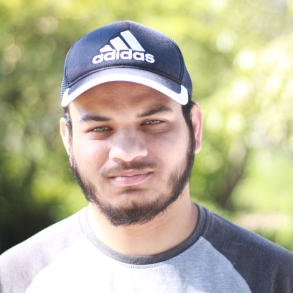}
\textbf{Md Monzurul Islam} is a graduate researcher in Civil Engineering
with a specialization in transportation at Texas State University and a
member of the Artificial Intelligence in Transportation Lab. He received
the M.Eng. degree in Architecture from the University of Tokyo and the
B.Sc. degree in Civil Engineering from Bangladesh University of Engineering
and Technology. His research interests include transportation safety, crash
risk assessment, pedestrian--vehicle interactions, driver behavior, machine
learning, statistical modeling, and GIS-based spatial analysis.
\endbio

\vspace{2em}

\bio{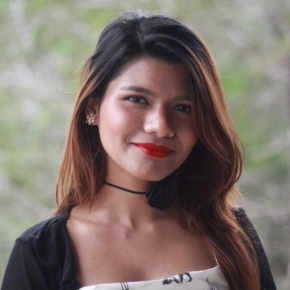}
\textbf{Mahmuda Sultana Mimi} received the M.S. degree in Civil Engineering
from Texas State University and the bachelor's degree in Urban and Regional
Planning from Bangladesh University of Engineering and Technology. She
conducted research in the Artificial Intelligence in Transportation Lab at
Texas State University. Her research interests include vulnerable road-user
safety, driver and travel behavior, micromobility, travel-demand modeling,
machine learning, econometric modeling, GIS-based spatial analysis, and
transportation data mining.
\endbio

\vspace{2em}
\newpage

\bio{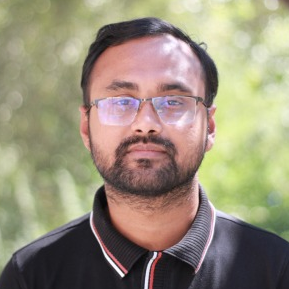}
\textbf{Sazzad Bin Bashar Polock} is a graduate researcher in the Ingram
School of Engineering and a member of the Artificial Intelligence in
Transportation Lab at Texas State University. He received the bachelor's
degree in Civil Engineering from Ahsanullah University of Science and
Technology in 2023. His research interests include traffic safety, crash-data
analysis, driver behavior, connected and automated vehicles, micromobility,
evacuation behavior, and machine-learning methods for crash and injury
severity analysis.
\endbio

\vspace{2em}

\bio{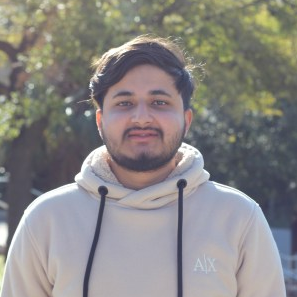}
\textbf{Gaurab Chhetri} is an undergraduate student in Computer Science and
an undergraduate research assistant in the Artificial Intelligence in
Transportation Lab at Texas State University. His research interests include
artificial intelligence, machine learning, natural-language processing,
efficient deep-learning architectures, computational social science, and
intelligent transportation systems. He is also interested in full-stack web
development, open-source research tools, interactive data visualization, and
the development of accessible data-driven systems for transportation research
and public communication.
\endbio
\vspace{2em}

\bio{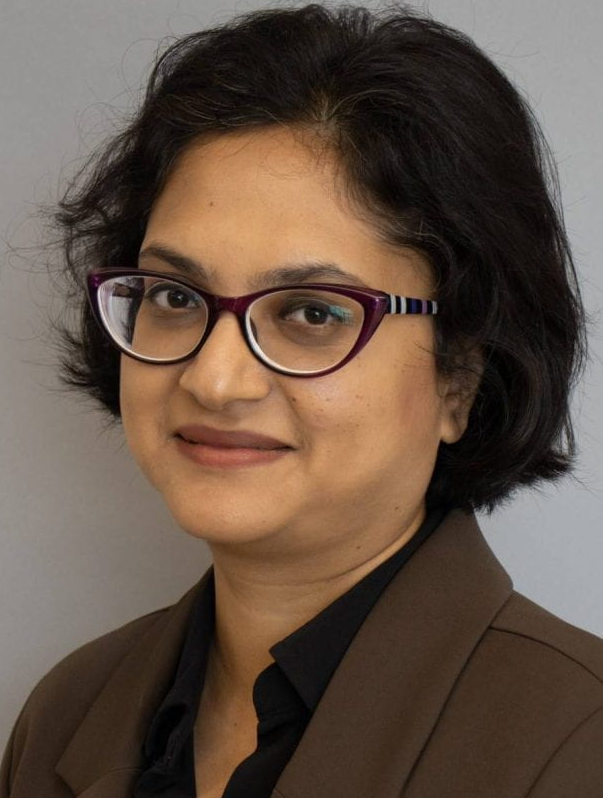}
\textbf{Anandi Dutta} is an Assistant Professor in the Ingram School of
Engineering at Texas State University. She received the Ph.D. degree in
Computer Science and Engineering from the University of Louisiana at
Lafayette and the M.S. degree in Electrical Engineering from Louisiana
State University. Her research interests include artificial intelligence
for engineering applications, edge and agentic AI, AI in healthcare,
natural-language processing, machine learning in transportation, and
energy-efficient computing architectures.
\endbio
\vspace{5em}

\bio{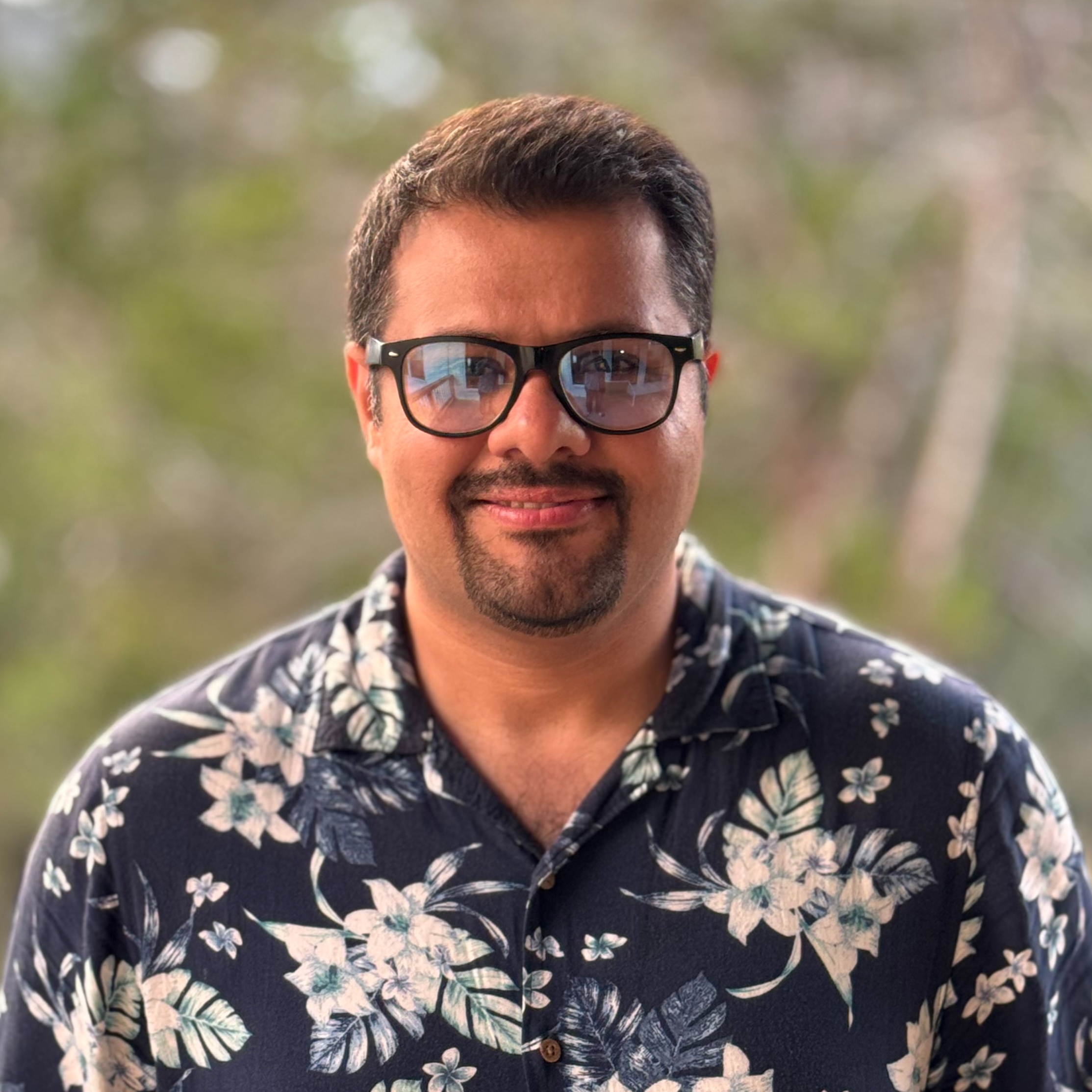}
\textbf{Amir Rafe} is a postdoctoral researcher in the Artificial Intelligence
in Transportation Lab at Texas State University. He received the Ph.D. degree
in Civil and Environmental Engineering from Utah State University. His
research interests include causal artificial intelligence, computational
epistemology, complexity science, transportation and pedestrian safety,
crowd and evacuation modeling, probabilistic reasoning, and disability-inclusive
decision support. His work focuses on explainable and causally informed
intelligent systems for complex, safety-critical environments.
\endbio
\vspace{2em}

\bio{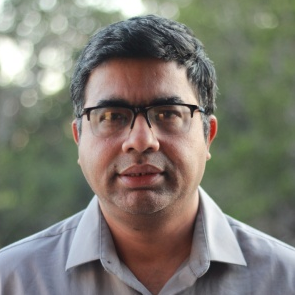}
\textbf{Subasish Das} is an Associate Professor of Civil Engineering and
Director of the Artificial Intelligence in Transportation (AIT) Lab at Texas
State University. He received the Ph.D. degree in Systems Engineering, with
a focus on Civil Engineering, from the University of Louisiana at Lafayette.
His research interests include transportation safety and operations,
connected and automated vehicles, causal and responsible AI, foundation
models, natural-language processing, digital twins, GIS, smart infrastructure,
and data-driven decision-support systems. 
\endbio

\end{document}